\documentclass[runningheads]{llncs}
\usepackage{amssymb}
\usepackage{graphbox}
\usepackage{gensymb}
\usepackage{pifont}% http://ctan.org/pkg/pifont
\newcommand{\cmark}{\ding{51}}%
\newcommand{\xmark}{\ding{55}}%

\usepackage[utf8]{inputenc}
\usepackage[T1]{fontenc}
\usepackage{textcomp}
\usepackage{gensymb}
\usepackage{wrapfig}

\usepackage{comment}
\usepackage{tabularx}
\usepackage{arydshln}
\usepackage{multirow}

\usepackage[mobile]{eccv}

\usepackage{eccvabbrv}

\usepackage{graphicx}
\usepackage{booktabs}

\usepackage[accsupp]{axessibility}  % Improves PDF readability for those with disabilities.

\usepackage{hyperref}

\usepackage{orcidlink}

\begin{document}

\def\negativespace{-3pt}

% ---------------------------------------------------------------
% TODO REVIEW: Replace with your title
\title{Towards Image Ambient Lighting Normalization} 

% TODO REVIEW: If the paper title is too long for the running head, you can set
% an abbreviated paper title here. If not, comment out.
\titlerunning{Ambient Lighting Normalization}

% TODO FINAL: Replace with your author list. 
% Include the authors' OCRID for the camera-ready version, if at all possible.
\author{Florin-Alexandru Vasluianu\inst{1} \and
Tim Seizinger\inst{1}  \and Zongwei Wu\inst{1}\thanks{Corresponding Author}, \\  Rakesh Ranjan\inst{2} \and Radu Timofte\inst{1}}

% TODO FINAL: Replace with an abbreviated list of authors.
\authorrunning{Ambient Lighting Normalization}
% First names are abbreviated in the running head.
% If there are more than two authors, 'et al.' is used.

% TODO FINAL: Replace with your institution list.
\institute{Computer Vision Lab, CAIDAS \& IFI, University of W\"{u}rzburg \and
Meta Reality Labs}

\maketitle

\begin{abstract}
Lighting normalization is a crucial but underexplored restoration task with broad applications. However, existing works often simplify this task within the context of shadow removal, limiting the light sources to one and oversimplifying the scene, thus excluding complex self-shadows and restricting surface classes to smooth ones. Although promising, such simplifications hinder generalizability to more realistic settings encountered in daily use. In this paper, we propose a new challenging task termed Ambient Lighting Normalization (ALN), which enables the study of interactions between shadows, unifying image restoration and shadow removal in a broader context. To address the lack of appropriate datasets for ALN, we introduce the large-scale high-resolution dataset \textbf{Ambient6K}, comprising samples obtained from multiple light sources and including self-shadows resulting from complex geometries, which is the first of its kind. For benchmarking, we select various mainstream methods and rigorously evaluate them on Ambient6K. Additionally, we propose \textbf{IFBlend}, a novel strong baseline that maximizes \textit{I}mage-\textit{F}requency joint entropy to selectively restore local areas under different lighting conditions, without relying on shadow localization priors. Experiments show that IFBlend achieves SOTA scores on Ambient6K and exhibits competitive performance on conventional shadow removal benchmarks compared to shadow-specific models with mask priors. The dataset, benchmark, and code are available at \url{https://github.com/fvasluianu97/IFBlend}.


\end{abstract}
\section{Introduction}
\label{sec:intro}

\begin{figure}[t]
  \centering
  \includegraphics[width=.95\linewidth]{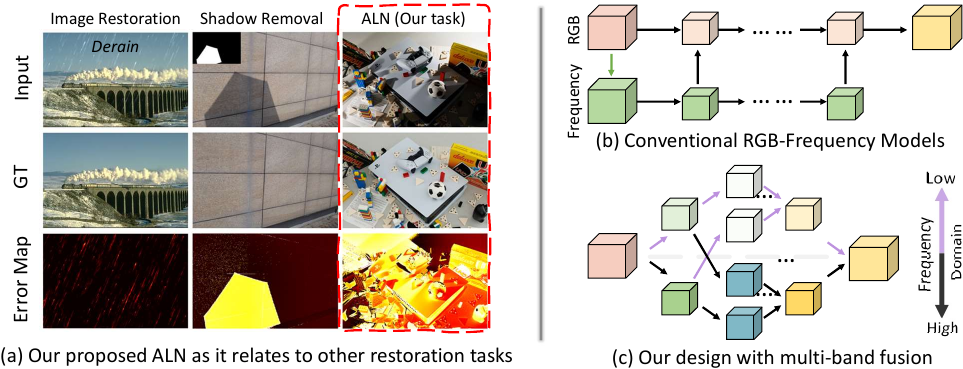}
    \vspace{-3mm}
  \caption{\textbf{Motivation.} (a) Conventional restoration tasks, such as deraining, typically aim to restore information across the entire image, evident from the error map. However, in the context of lighting normalization, which has been traditionally oversimplified as shadow removal with only one lighting source and smooth surface, the objective becomes the recovery of local information within the shadow mask. Nonetheless, this simplified setup lacks realism for everyday scenarios. To address this limitation, we propose the Ambient Lighting Normalization (ALN) task, which explores the complexities of shadow interactions. Model-wise, we introduce a robust baseline by capitalizing on the synergy between image and frequency domains. Unlike conventional approaches that adopt a dual-branch fusion design, as depicted in (b), we propose a shrinkage multi-band fusion technique (c) to maximize their joint entropy.}
  \label{fig:teaser}
  \vspace{-5mm}
\end{figure}

When capturing an image, shadows may manifest when an object obstructs light from reaching the observed surface \cite{stamminger2002perspective}, resulting in darker regions whose characteristics depend on various factors including scene geometry, light sources, and material properties such as color, reflectance, and surface smoothness. These shadowed areas present challenges in various computer vision tasks such as image editing, synthesis, and scene understanding \cite{kristan2015visual,he2017mask,shelhamer2016fully,garcia2018survey}, sparking interest in restoring image information within shadow regions.

Recent advancements in learning-based methods, exemplified by \cite{ISTDwang2018STCGAN, Le_2019_ICCV, jin2021dc, guo2023shadowformer, guo2023shadowdiffusion}, have made significant strides, largely fueled by the availability of benchmark datasets like SRD~\cite{SRDDESHADOW}, ISTD~\cite{ISTDwang2018STCGAN}, USR~\cite{USRhu2019mask}, and WSRD~\cite{Vasluianu_2023_WSRD}. These datasets have enabled solving the shadow removal task as a regression problem. However, the current approach suffers from two primary limitations. On the one hand, dataset-wise, we observe that benchmarking datasets often employ a simplified shadow model where an external occluder casts a shadow on a flat surface under natural light. This setup excludes the most common type of shadow, the self-shadow, and restricts the class of surfaces to smooth ones. Consequently, semantic inconsistencies and noise may arise from self-shadows, impacting the transformation to the shadow-free domain.

On the other hand, from a modeling perspective, existing shadow removal methods often rely on shadow masks, either directly incorporated into the input data or learned as pseudo shadows supervised by ground truth. However, this reliance on mask priors presents several challenges: (i) Obtaining accurate masks in complex scenarios is impractical; (ii) the quality of the shadow mask significantly impacts algorithm performance; (iii) many restoration models, such as \cite{zhang2022spa, jin2021dc}, effectively operate without masks, recovering information across the entire image. Consequently, imposing mask requirements disrupts the restoration process's natural flow, especially in designing a generic restoration model applicable to all tasks, including shadow removal.

In this paper, we aim to unify image restoration and shadow removal within the broader context of high-resolution Ambient Lighting Normalization (ALN). The objective is to restore lost image details in scenarios with uneven lighting, allowing for the study of interactions between shadows from multiple light sources and self-shadows arising from complex geometries. We first conducted a systematic study of established image restoration baselines \cite{liang2021swinir, wang2022uformer, zamir2022restormer} on conventional shadow removal benchmarks. Their competitive performances without requiring additional mask priors validate our initial assumption of generalizing image restoration methods on the shadow removal task. Additionally, to advance ALN research, we introduce the large-scale Ambient6K dataset, featuring complex and realistic lighting conditions, which is the first of its kind. Our dataset contains samples obtained with multiple directional light sources of varying sizes, resulting in more complex and natural shadow formations.

\begin{figure}[t]
    \centering
    \resizebox{.70\linewidth}{!} {
    \setlength{\tabcolsep}{2pt}
    \begin{tabular}{cc}
         \includegraphics[width=0.5\linewidth]{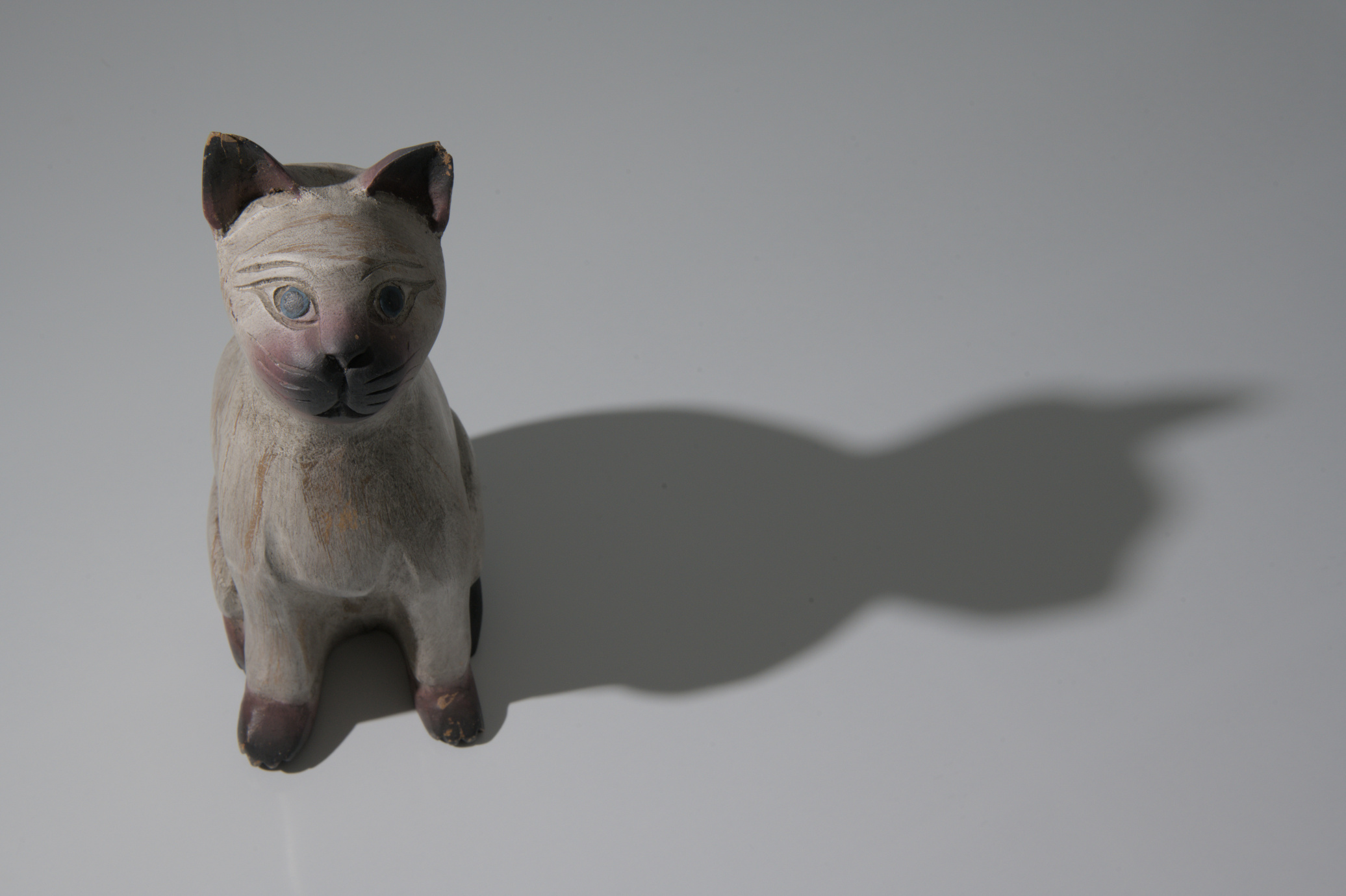}
         &  \includegraphics[width=0.5\linewidth]{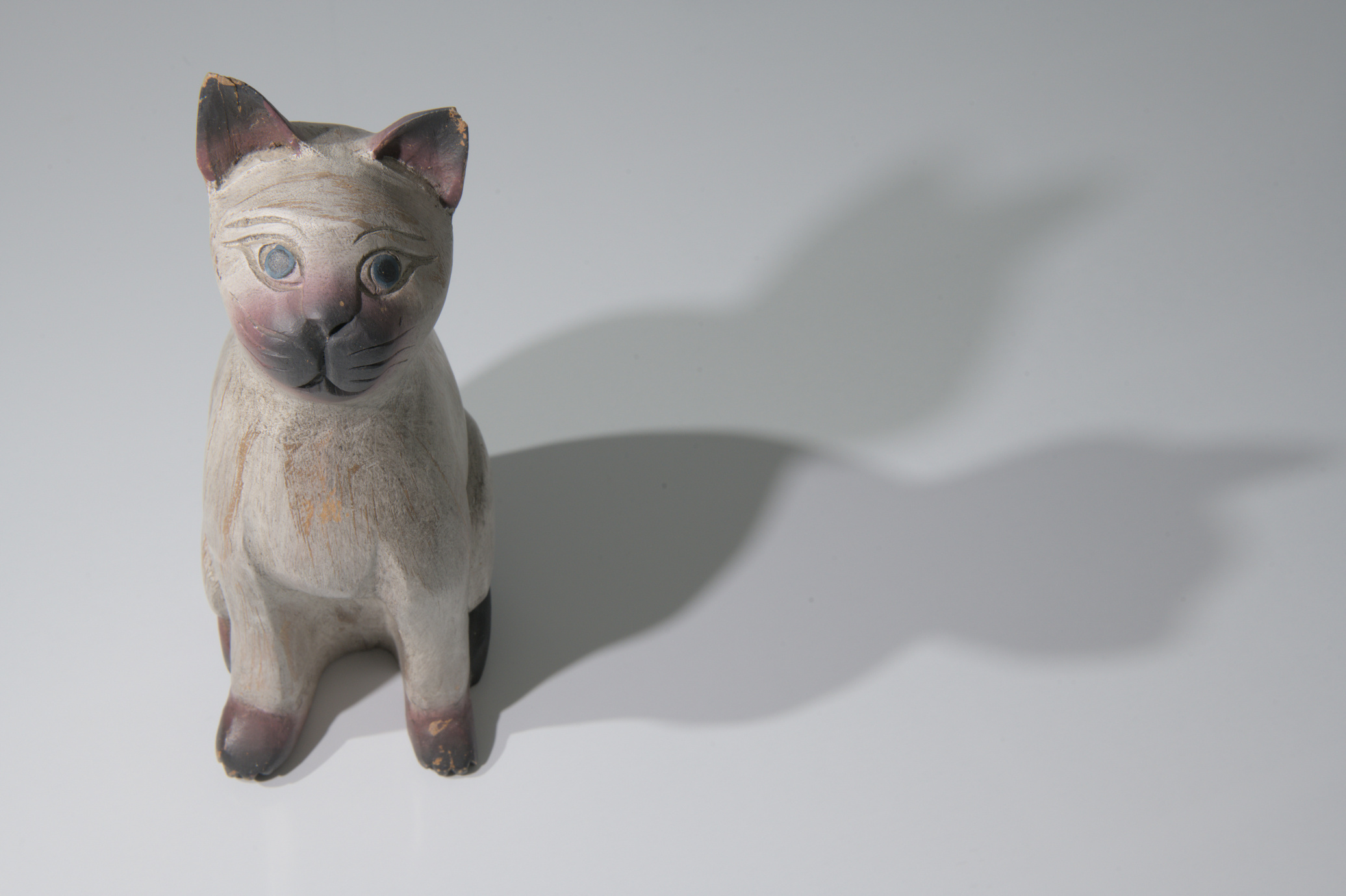} \\
         \textbf{A} & \textbf{B} \\
    \end{tabular}
    }
    \vspace{-3mm}
    \caption{Visualization of shadow configurations. \textbf{A:} Self-casted shadow configuration with a unique directional light. Note the non-linear intensity in the shadow's penumbra towards the edges. \textbf{B:} A second directional light is added to the setup. Note the different shadow intensities in different areas, depending on the intensity of each light source.}
    \label{fig:shadow-explain}
    \vspace{-6mm}
\end{figure}

Based on our Ambient6K dataset, we provide a systematic study on 10 mainstream baselines from tasks such as General Image Restoration \cite{chen2022simple, mehri2021mprnet, cui2022selective, liang2021swinir, wang2022uformer, zamir2022restormer, Chen_2021_CVPR}, or particular subtasks like Image Dehazing \cite{fu2021dw}, or Image Shadow Removal \cite{jin2021dc, zhang2022spa}. We observe that these models excel in globally normalizing lighting across images but struggle with accurately restoring locally shadowed regions. This limitation likely arises from restoration models prioritizing overall information recovery without adequately addressing local semantic inconsistencies. To overcome this challenge, we propose IFBlend to normalize the light through \textbf{I}mage-\textbf{F}reqency \textbf{Blend}ing, leveraging both visual and frequency cues to maximize joint entropy and enhance shadow restoration. While frequency is commonly used in low-level vision tasks \cite{cui2022selective, fu2021dw, 10208804, zhang2022spa}, existing methods \cite{fu2021dw, zhang2022spa, cui2022selective} often lack local awareness, hindering their application in shadow removal or ALN, which demands a nuanced understanding of different local regions. Differently, we introduce a refined mutual-interaction approach, which involves separately combining low-frequency and high-frequency domains before merging them to enhanced feature outputs. Such a decomposition allows us to isolate and address image segments independently based on their illumination characteristics. Meanwhile, decomposing frequency features enables capturing high-frequency details and textures, as well as low-frequency components representing global illumination and image structure. Consequently, our domain-specific fusion enables selective modification of different local regions.

To conclude, our contributions can be summarized as follows:
   \begin{itemize}
    \item We introduce a novel task termed Ambient Lighting Normalization (ALN) to address the challenges associated with non-optimal lighting conditions during image acquisition, unifying image restoration and shadow removal within a broader context.
    \item We present Ambient6K, the first large-scale and high-resolution dataset with more than 6,000 samples captured in a meticulously designed setup refined for optimal light distribution.
    \item To establish benchmarks, we select 10 high-performing approaches from image restoration or shadow removal, with pure images as inputs.
    \item  We propose IFBlend, a novel method that integrates both image and frequency representations for restoration, serving as a baseline for the ALN task. IFBlend outperforms established Image Restoration models and mask-free shadow removal models, showcasing its efficacy.  
\end{itemize}

\section{Related Work}
\label{sec:related_work}

In this study, we aim to explore a more comprehensive understanding of shadow removal under diverse and complex lighting conditions. To achieve this, we propose integrating image restoration and shadow removal within the broader framework of Ambient Lighting Normalization (ALN). This approach allows us to examine interactions between self-shadows and casted shadows, consider multiple light sources with varying intensities, and encompass a diverse range of materials. We anticipate that addressing this challenge will have significant implications for various downstream tasks, such as facial recognition~\cite{heusch2005lighting, xie2006efficient, 5658185}, iris recognition~\cite{takano2007rotation}, and gesture modeling~\cite{lee2005practical, lahiri2021lipsync3d}.

\noindent \textbf{Shadow Removal:} Shadow removal has long posed a significant challenge in research, encompassing various unexplored aspects. Early approaches relied on physical models to segment images into shadow-affected and shadow-free layers~\cite{finlayson2002removing,Finlayson_entropyminimization}. However, these methods, despite utilizing shadow masks for pixel identification, were constrained by simplified shadow models and manual feature engineering~\cite{10.1145/1243980.1243982,Shor:2008:TSM,7893803, guo2013paired, gong2014interactive}.

Advancements in hardware and neural networks facilitated the development of de-shadowing modules trained on large datasets, leading to improved performance~\cite{10.1007/978-3-319-46466-4_49, SRDDESHADOW, ISTDwang2018STCGAN, USRhu2019mask, Vasluianu_2023_WSRD}. Nevertheless, the shadow mask remains crucial, guiding neural networks to address shadow-affected areas explicitly~\cite{SRDDESHADOW, Le_2019_ICCV}. Some approaches leverage the shadow mask for auto-exposure and diffusion modeling, enhancing performance~\cite{fu2021auto, guo2023shadowformer, jin2023des3, ho2020denoising, guo2023shadowdiffusion}. Others incorporate attention mechanisms for improved results~\cite{hu2019direction, 6241432, guo2013paired, liang2021swinir, chen2022simple, wang2022uformer, zamir2022restormer}. However, these methods invariably require paired image-shadow mask inputs~\cite{vasluianu2021shadow, Le_2020_ECCV}, which are labor-intensive and challenging to collect at scale and accurately.

\noindent \textbf{Image Restoration:} Image Restoration aims to recover the information lost due to various degradations suffered by an image during manipulation, or due to under-optimal conditions in image acquisition. It has received significant attention in the Computer Vision field, particularly in tasks such as Image Denoising \cite{buades2011non, 7839189, liang2021swinir, wang2022uformer, zamir2020cycleisp, zamir2022restormer, ho2020denoising}, Image Super-Resolution \cite{10.1007/978-3-319-16817-3_8, zhang2021designing, Cai_2019_ICCV, chen2022simple, lugmayr2020srflow, li2022srdiff, saharia2022image}, Image Deblurring \cite{mehri2021mprnet, chen2022simple, zamir2022restormer, rim2022realistic, liang2022vrt, delbracio2023inversion},  or adverse conditions in image acquisition related problems, like Image Dehazing \cite{ancuti2018haze, ancuti2020ntire, fu2021dw, luo2023refusion}, Deraining \cite{Chen_2021_CVPR, zamir2022restormer, luo2023image, cui2022selective} or  High Dynamic Range (HDR) Imaging \cite{chen2021hdrunet, tel2023alignment}. Similar to many other vision tasks, traditional handcrafted restoration methods \cite{buades2011non,10.1007/978-3-319-16817-3_8} have given way to neural network-based solutions \cite{7839189, Cai_2019_ICCV, mehri2021mprnet, zamir2020cycleisp}. The recent adoption of the attention mechanism has further boosted the performance \cite{vaswani2017attention,liu2021Swin, liang2021swinir, chen2022simple, wang2022uformer, zamir2022restormer} through different task-specific variations.

Despite successful applications in restoration, shadow removal remains underexplored due to differing problem settings. Restoration methods typically focus on global image recovery, while shadow removal targets localized segments. Historically, shadow removal assumed oversimplified scenarios, neglecting self-shadows and heavily relying on image-shadow mask pairs for prediction accuracy, contrary to restoration's mask-free approach. This approach limits the model's efficacy and fails to represent common shadow types, such as self-shadows. To address this, we propose studying shadow removal within the context of ALN, integrating multiple light sources and more complex geometries.

This work unifies image restoration and shadow removal through ALN. We retrain milestone restoration methods on shadow removal benchmarks, showcasing their potential compared to mask-prior models. Additionally, we introduce a new dataset with a complex lighting setup to facilitate ALN research, evaluating mainstream methods meticulously on our benchmark.

\noindent \textbf{Frequency in Low-level Vision:} The inherent duality of spatial and frequency domain image representations is central for restoration, especially by deploying algorithms like Fast-Fourier Transform or Discrete Wavelet Transform for Image Denoising \cite{9917526}, Image Dehazing \cite{fu2021dw, wang2023frequency, zhou2023breaking}, Lens Flare Removal \cite{10208804}, Shadow Removal \cite{zhang2022spa, Li_2023_ICCV}, HDR imaging \cite{xu2022fmnet}, or general restoration models \cite{cui2022selective}.

Many of the aforementioned works \cite{fu2021dw, 10208804, Li_2023_ICCV, xu2022fmnet} simply transform image features into frequency representations and process them in an additional branch. While benefiting from this dual-modality, such designs operate at a global level, lacking awareness of geometry and locality. Directly extending these methods to ALN is non-trivial because the scene's geometry, which determines shadow interactions due to the random orientation of directional lights, plays a crucial role. Therefore, in this work, we enhance frequency modeling to capture fine-grained details, effectively addressing restoration challenges in complex scenes.

\section{Ambient6K Dataset}
%\textbf{Dataset:} 

\begin{table}[t]
\centering
%\tiny
\setlength{\tabcolsep}{2pt}
\caption{\textbf{Dataset Comparison:} Statistics of existing datasets for lighting normalization, oversimplified within the shadow removal context. Our Ambient6K is the first of its kind offering samples captured under more intricate lighting and scene conditions, while boasting the largest sample size and the highest resolution.}
\vspace{-3mm}
\resizebox{\linewidth}{!}
{
\begin{tabular}{lcccccccc}
\toprule
Dataset                         & Samples    & Train/Val./Test       & Resolution                & Scenes    & Objects   & Surfaces & Lights  \\ 
\midrule
ISTD\cite{ISTDwang2018STCGAN} & 1870          & 1330/ --- /\textbf{540}          & 640$\times$480            & 135          & 0            &  ---            & 1          \\
SRD\cite{SRDDESHADOW}          & 3088          & 2680/ --- /480          & 840$\times$640            & 130          & 0            &  ---           & 1          \\
WSRD\cite{Vasluianu_2023_WSRD}  & 1100          & 1000/100/ ---          & 1440$\times$1080 & 150          & 150          & 14          & 1          \\ \midrule
\textbf{Ambient6K \emph{(ours)}}   & \textbf{6000} & \textbf{5000/500/}500 & \textbf{2560$\times$1440} & \textbf{200} & \textbf{250} & \textbf{32} & \textbf{3} \\ 
\bottomrule
\end{tabular}
}
\vspace{-5mm}
\label{tab:dataset-comparison}
\end{table}

\begin{figure}[t]
    \centering
    \includegraphics[width=.8\linewidth]{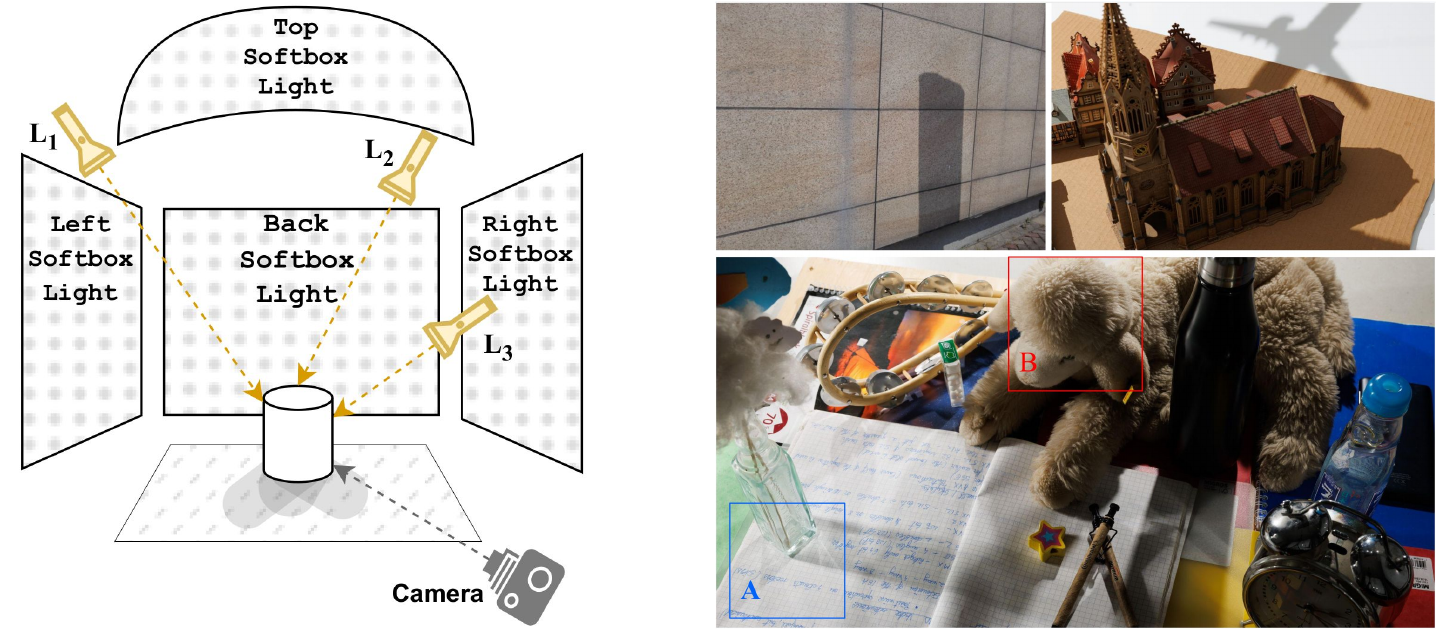}
        \vspace{-2mm}
    \caption{\textbf{Left:} The professional setup used for data acquisition. The directional lights $L_1 - L3$ are used to cast shadows in the scene, while the softbox lights counter the self-shadows casted by rough surfaces. \textbf{Right:} The typical lighting scenario in ISTD \cite{ISTDwang2018STCGAN} (top left), WSRD \cite{Vasluianu_2023_WSRD} (top right), and Ambient6K (bottom). Note the non-uniform lighting distribution due to caustics \textbf{(\textcolor{blue}{A})} or rough surfaces \textbf{(\textcolor{red}{B})}, and the increased complexity scenes represented at data level.}
    \label{fig:setup}
    \vspace{-5mm}
\end{figure}

Our Ambient6K dataset serves as a cornerstone in the field of ALN, encompassing a diverse range of interactions observed within challenging scenes featuring various materials and geometries under a multitude of lighting scenarios. Each scene is meticulously captured, yielding approximately 30 high-resolution images under different lighting conditions, with the ground-truth image acquired under optimal lighting conditions typical of professional setups (see \cref{fig:setup}). The metadata, providing insights into scene objects and properties, will also be provided.

The setup comprises four softbox lights to minimize self-shadows in the captured ground-truth images, along with three directional lights (L1, L2, and L3) whose positions and intensities are adjustable for each image capture. The directional lights are exclusively activated during input image acquisition to cast shadows in the scene, while the softbox lights are utilized solely for capturing the ground-truth image. To introduce further variation in captured shadow phenomena, parabolic reflectors are occasionally employed to alter the diameter of light sources and thereby adjust the relative size of the penumbra region in their cast shadows. All lights feature controllable intensity and temperature settings, aligned to match the ambient lighting corresponding to natural light.

The image samples are captured using a modern high-end DSLM camera, the Canon EOS R6II, coupled with a professional lens-flare-resistant studio zoom lens, the Canon RF 28-70mm F2.0. The camera is mounted on a tripod and operated remotely to ensure perfectly pixel-aligned data. Varying angles and focal lengths for capturing observations add another layer of diversity.

Quantitative evaluations comparing the novel Ambient6K dataset against existing shadow removal benchmarks are presented in \cref{tab:dataset-comparison}, demonstrating its superiority. Notably, the implemented capturing setup enhances the quality of ground-truth acquisition. Unlike previous setups that resulted in soft drop shadows and uneven illumination, our setup utilizes four diffuse softbox lights positioned around the subject scene, ensuring uniform illumination from all angles and eliminating shadow casting. This 360-degree surround illumination minimizes color inconsistencies and semantic changes between input and ground truth images, which were not the case in previous benchmarks like SRD \cite{SRDDESHADOW} and ISTD \cite{ISTDwang2018STCGAN}. Additionally, our setup mitigates pixel misalignment issues present in benchmarks such as WSRD \cite{Vasluianu_2023_WSRD}.

\begin{figure}[t]
    \centering
    \resizebox{\linewidth}{!}{
    \setlength{\tabcolsep}{1pt}
    \def\widthfact{0.192}
    \renewcommand{\arraystretch}{4.8}
    \begin{tabularx}{\linewidth}{cccccc}
         \includegraphics[align=c, width=\widthfact\linewidth]{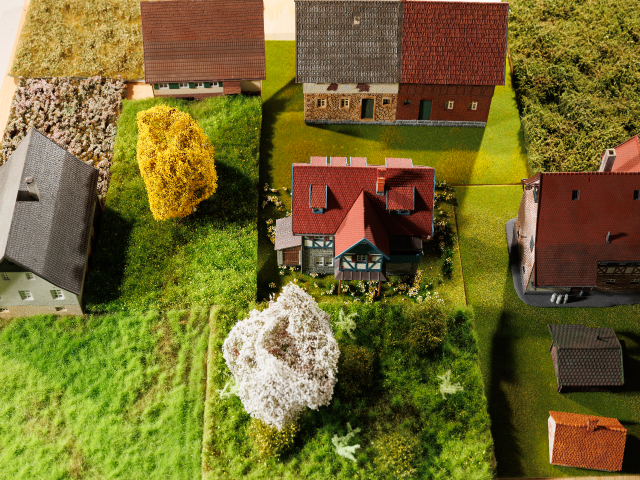}
         &\includegraphics[align=c, width=\widthfact\linewidth]{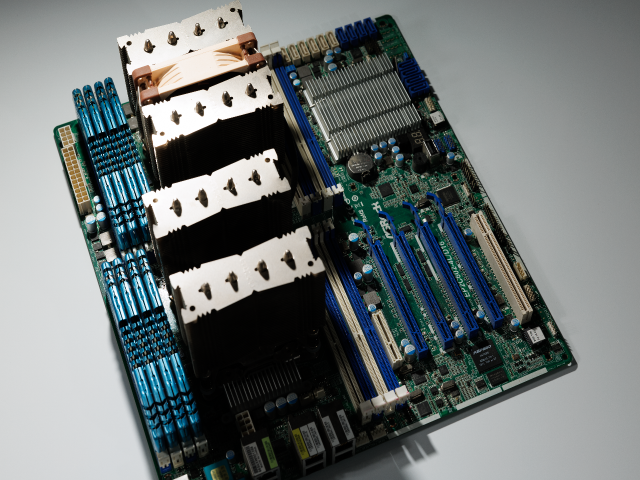}
         &\includegraphics[align=c, width=\widthfact\linewidth]{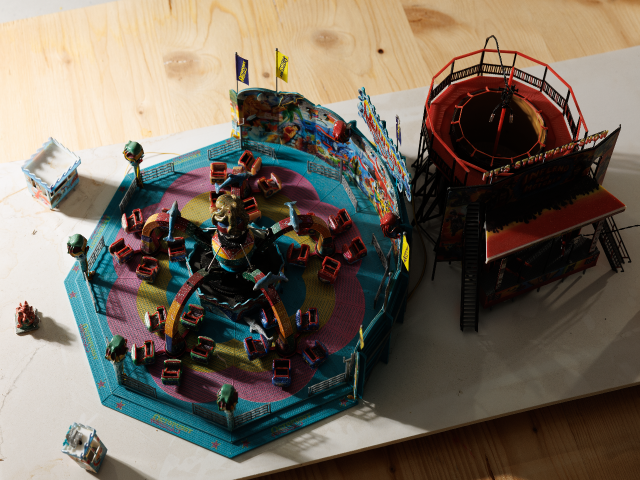}
         &\includegraphics[align=c, width=\widthfact\linewidth]{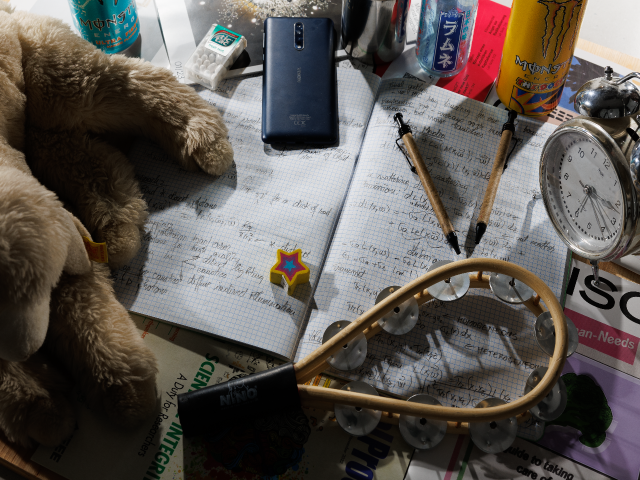}
         &\includegraphics[align=c, width=\widthfact\linewidth]{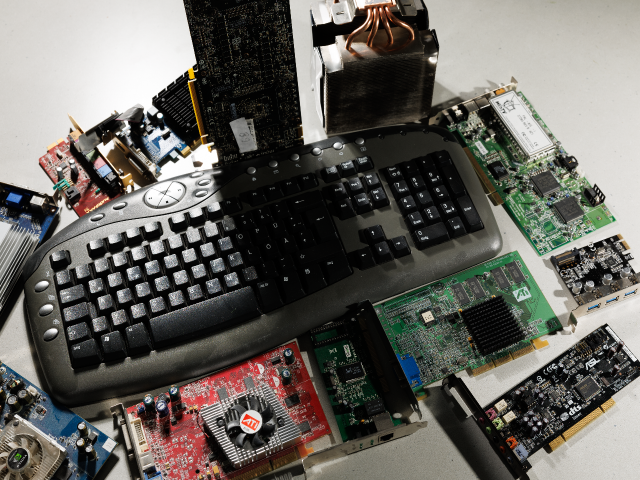} \\
         \includegraphics[align=c, width=\widthfact\linewidth]{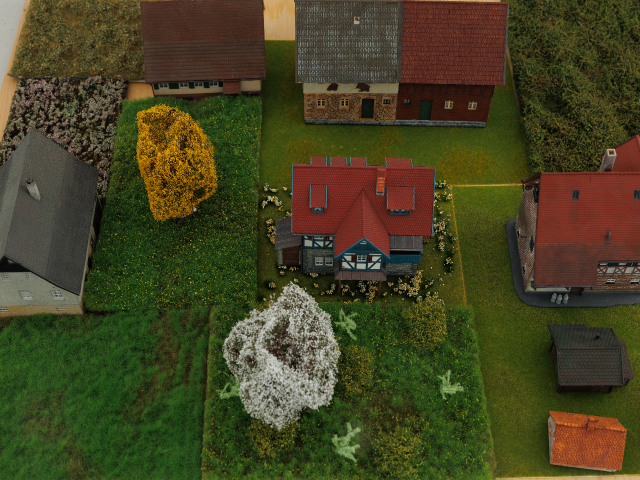} 
         &  \includegraphics[align=c, width=\widthfact\linewidth]{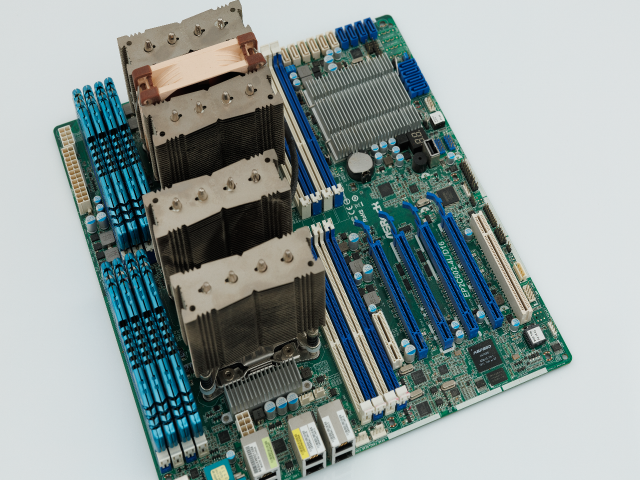}
         &  \includegraphics[align=c, width=\widthfact\linewidth]{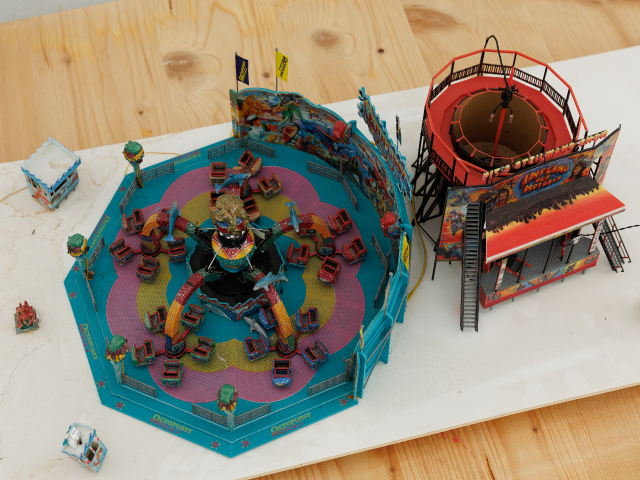}
         &  \includegraphics[align=c, width=\widthfact\linewidth]{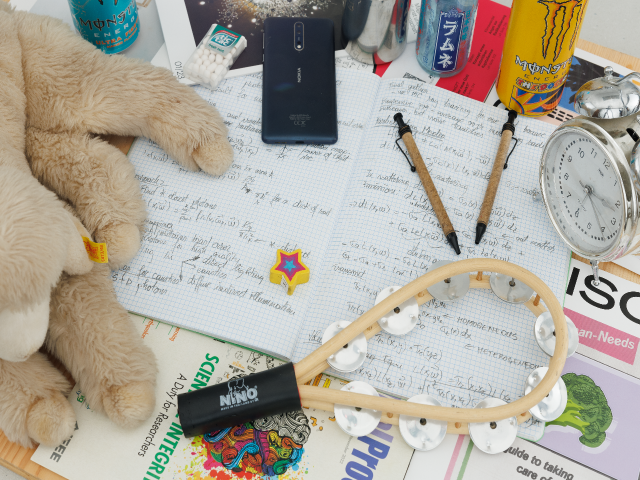}
         &  \includegraphics[align=c, width=\widthfact\linewidth]{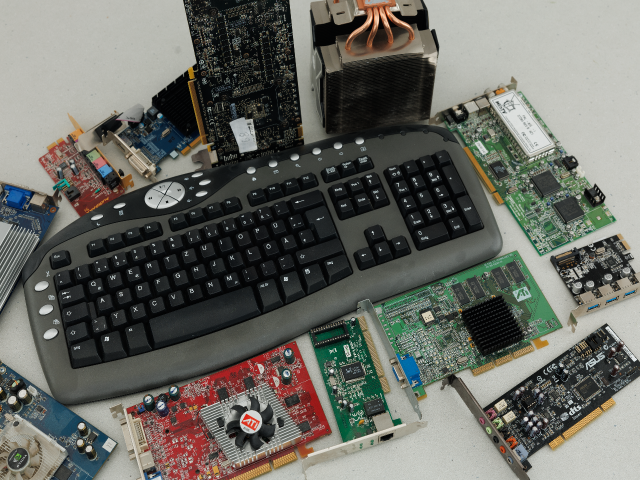}
    \end{tabularx}
    }
    \vspace{-3mm}
    \caption{\textbf{Top:} Input samples from the Ambient6K benchmark. Note the high complexity of the scenes, with various contents and textures. Combined with the complex illumination model, the shadow patterns are characterized by increased variety in terms of shape, intensity and interactions.  \textbf{Bottom:} Equivalent ground-truth images acquired in optimal lightning conditions. }
    \label{fig:ALND-samples}
    \vspace{-4mm}
\end{figure}

\cref{fig:setup} provides a visual comparison of the dataset properties, showcasing the advantages of our setup over existing benchmarks. Our proposed setup facilitates the study of light normalization models under challenging conditions, including complex shadows cast by translucent materials (\textcolor{blue}{\textbf{A}}) and difficult self-shadows on rough surfaces (\textcolor{red}{\textbf{B}}). Further supporting our claims, additional samples are presented in \cref{fig:ALND-samples}, demonstrating the versatility of our data capturing setup across various interaction scenarios.

\section{Method}
\label{sec:method}

\begin{figure}[t]
    \centering
    \includegraphics[width=\linewidth]{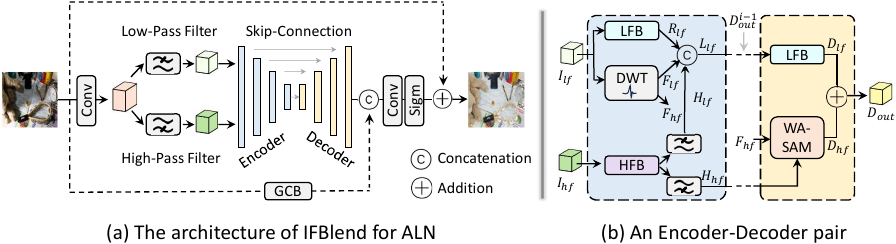}
    \vspace{-5mm}
    \caption{(a) The architecture of our proposed IFBlend for ALN involves splitting the image feature into low- and high-frequency domains, treating them independently. (b) The encoder-decoder pair's inner structure aims to progressively compute, refine, and combine features across various frequency domains. The final output undergoes a shrinkage fusion process to achieve enhanced lighting normalization. The Global Context Branch (GCB) is based on  ConvNext \cite{liu2022convnet}.}
    \label{fig:cln-arch}
    \vspace{-5mm}
\end{figure}

ALN's primary aim is to modify shadowed areas while maintaining image integrity. However, direct scene measurement from visual pixels is challenging due to shadows obscuring details and reducing object clarity. To address these challenges, we propose IFBlend (see Fig.~\ref{fig:cln-arch}) -- a novel model leveraging frequency as a guiding prior for feature modeling and decoding. Different from previous works \cite{fu2021dw, cui2022selective, 10208804} based on global implicit fusion, we introduce locality to guide the fusion process.

Specifically, we introduce a shrinkage fusion design, decomposing input features into low- and high-frequency domains. By separately fusing low- and high-frequency components with their corresponding image counterparts, we selectively modify different local areas.

\noindent \textbf{Encoder:} Technically, we first project the input feature into a latent space at the shallow stage. To decompose this image feature into different frequency domains, we apply low-pass and high-pass filters simultaneously, which are mimicked by the average pooling and maximum pooling operations, respectively. Hence, we obtain the low-frequency image feature $I_{lf}$ and $I_{hf}$. The former $I_{lf}$ is expected to contain information about the global illumination and overall structure of the image, while the latter $I_{hf}$ is expected to contain details and textures that are smaller in scale and have rapid variations.

Then for each domain-specific image feature, we aim to refine it with a secondary level of decomposition. For $I_{lf}$, we adopt a parallel refinement process, with both implicit learning low-frequency block (LFB) and an explicit Haar DWT transformation. The LFB is a series of Conv2D, BatchNorm, LeakyReLU, and Dropout, which aims to have a more nuanced understanding of the low-frequency features, transforming $I_{lf}$ to refined feature $R_{lf}$ thanks to the data prior, while the Haar DWT transformation outputs enhance the model's ability to capture and analyze the frequency characteristics, outputting the low-frequency feature $F_{lf}$ and the high-frequency feature $F_{hf}$. The details of LFB can be found in the supplementary material. Mathematically, we have:

\begin{equation}
    R_{lf} = LFB(I_{lf}); \quad F_{lf}, F_{hf} = H\text{-}DWT(I_{lf}).
\end{equation}

Similarly, we adopt a secondary level of decomposition on $I_{hf}$ with implicit high-frequency learning block (HFB), composed of LayerNorm, Conv2D, DynConv, BatchNorm, LeakyReLU, and Dropout. The details of HFB can be found in the supplementary material. The output feature is further decomposed through low-pass and high-pass filters, yielding $H_{lf}$ and $H_{hf}$. We have:

\begin{equation}
    H_{lf}/ H_{hf} = Avg/MaxPool(HFB(I_{hf})).
\end{equation}

When  $H_{hf}$ directly becomes the encoded high-frequency output and is fed in the next encoder stage,  the $H_{lf}$ is further merged with other low-frequency features $R_{lf}$ and $F_{lf}$, forming the encoded low-frequency output $L_{lf}  = [R_{lf}; F_{lf}; H_{lf}] $, where $[.]$ is the concatenation operation.

\vspace{0.1cm}
\noindent\textbf{Decoder:} For simplicity and clarity, we detail the first/deepest decoding stage as an example, which takes $F_{hf}$, $H_{hf}$, and $L_{lf}$ from the latest encoded stage as inputs. The objective is to decode the domain-specific features separately, and finally merge them together to form the overall decoded feature.

Specifically, for the low-frequency feature $L_{lf}$, we decode it with another LFB block, obtaining $D_{lf}$. For the high-frequency features $H_{hf}$ and $F_{hf}$, we leverage the cross-attention to deeply leverage the mutual benefits between the image and frequency domain. 

\begin{wrapfigure}{r}{0.32\textwidth}
\includegraphics[width=\linewidth,keepaspectratio]{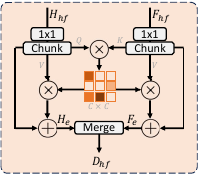}
\caption{A graphical representation of the WA-SAM attention module, where we compute the affinity matrix between image and frequency features, aiming for an efficient fusion.}
    \label{fig:cln-att}
    \vspace{-5mm}
\end{wrapfigure}

As shown in \cref{fig:cln-att}, we first compute the affinity matrix $M$ between $H_{hf}$ and $F_{hf}$, which is further used to improve the image and frequency feature modeling separately, obtaining the enhanced feature $H_{e}$ and $F_e$, respectively. Finally, these enhanced features are merged, through Concat, Conv2D, and ReLU, to form the decoded high-frequency feature $D_{hf}$. Finally, we obtain the decoded output $D_{out}$ by adding $D_{hf}$ with $D_{lf}$.

\vspace{0.1cm}
\noindent\textbf{Remarks:} Our model design is crafted to widen the gap between low-frequency and high-frequency features through a gradual fusion of domain-specific features. This coarse-to-fine fusion process pulls domain-specific features in opposing directions, leading to maximized joint entropy. Consequently, our model efficiently harnesses image and frequency cues, enhancing understanding of light conditions and facilitating ALN. Moreover, our approach strategically allocates computational resources based on feature complexity, enabling the assignment of more learning parameters to intricate high-frequency features, as evidenced by the inclusion of attention mechanisms and sophisticated HFB block.

\vspace{0.3cm}
\noindent{\textbf{Learning Setting: }}
All experiments were performed on two NVIDIA GeForce RTX 4090 GPU modules under the PyTorch framework. We use the Adam~\cite{kingma2014adam} algorithm. The network is trained end-to-end with $\mathcal{L}_1$ loss and the $\mathcal{L}_{SSIM}$~\cite{1284395} loss. Given $I$ the ground-truth image and $I_r$ the reconstruction of our model, we have:
\begin{equation}
    \mathcal{L}(I_r, I) = \mathcal{L}_1(I_r - I) + \lambda \cdot (1 - \mathcal{L}_{SSIM}(I_r, I)),
    \label{eq:loss}
\end{equation}
where $\lambda$ is a proportion hyperparameter.

\section{Results}
\label{sec:Results}

\subsection{Benchmark on Ambient6K}

\noindent \textbf{Metrics:} We report metrics quantifying the restoration fidelity, through PSNR, and the perceived image quality, through SSIM \cite{1284395} and LPIPS \cite{zhang2018perceptual}, characterizing the RGB image domain. 

\noindent \textbf{Compared methods:} For benchmarking, we follow the officially released code and retrain mainstream models on our Ambient6K dataset. We compare the proposed IFBlend model against well-proven solutions in the image restoration field, like NAFNet \cite{chen2022simple}, MPRNet \cite{mehri2021mprnet}, SFNet \cite{cui2022selective}, SwinIR \cite{liang2021swinir}, Uformer \cite{wang2022uformer}, Restormer \cite{zamir2022restormer} or HINet \cite{Chen_2021_CVPR}. Moreover, we include well-established methods in Image Dehazing \cite{fu2021dw} or Image Shadow Removal \cite{jin2021dc, zhang2022spa}. We compare against other methods benefitting from the frequency domain prior, such as SFNet \cite{cui2022selective}, DW-NET \cite{fu2021dw} and SpA-Former \cite{zhang2022spa}.  

All the results are produced after training on the Ambient6K dataset for 200 epochs, under the same conditions and under the training objective provided in \cref{eq:loss}, starting from a random initialization.

\begin{table}[t]
\centering
%\tiny
\setlength{\tabcolsep}{3pt}
\caption{Our IFBlend versus SOTA image restoration solutions, on the proposed Ambient6K test split. \textbf{Best results are in bold.}}
\vspace{-3mm}
\resizebox{\linewidth}{!}
{
\begin{tabular}{@{}X|lcc|ccccc@{}}
    %\begin{tabularx}{\linewidth}{@{}X|ccc|ccccc@{}}
\toprule
Method          &Task            & Type      & Prior  & MACs (G.) & \text{PSNR}$\uparrow$ & \text{SSIM}$\uparrow$ & \text{LPIPS}$\downarrow$ \\ \midrule

unprocessed         & -                & -     & -   & -   & 13.592                & 0.658                 & 0.226   \\
DCShadowNet \cite{jin2021dc}   &Shadow Remov.  & Conv  & RGB  & 13.15     & 17.731                 & 0.711                 & 0.187    \\

SpA-Former \cite{zhang2022spa}   &Shadow Remov.  & Transf.  & RGB + Freq. & 16.82           & 19.594                & 0.806                 & 0.130    \\

NAFNet \cite{chen2022simple}    & Restoration  & Conv   & RGB & 15.92       & 19.750                & 0.798                 & 0.132    \\

DW-NET \cite{fu2021dw}     &   Dehazing  & Conv   & RGB + Freq.  & 7.54        & 19.804                & 0.792                 & 0.136    \\
MPRNet \cite{mehri2021mprnet}  & Restoration & Conv.  & RGB  & 37.21      & 19.967                 & 0.801                 & 0.128   \\

SFNet  \cite{cui2022selective}  & Restoration  & Transf.  & RGB + Freq.  & 31.27       & 20.031                & 0.807                 & 0.132    \\
SwinIR \cite{liang2021swinir}    & Restoration & Transf.   & RGB  & 37.81       & 20.246                & \textbf{0.814}                 & 0.123    \\
Uformer \cite{wang2022uformer}   & Restoration  & Transf.  & RGB  & 19.33       & 20.273                & 0.810                 & 0.125    \\
Restormer \cite{zamir2022restormer} & Restoration & Transf.  & RGB & 35.31     & 20.446                & 0.808                 & 0.124      \\
HINet \cite{Chen_2021_CVPR}     & Restoration  & Conv  & RGB & 42.68       & 20.467                & 0.806                 & 0.127      \\
\hdashline
IFBlend (\emph{ours})          & Restoration      & Conv   & RGB + Freq.  & 26.01        & \textbf{20.714}                & 0.810         & \textbf{0.122} \\

\bottomrule
%\end{tabularx}%
\end{tabular}%
}
\vspace{-3mm}
\label{tab:quanti-alnd-test}
\end{table}

\noindent \textbf{Quantitative Results:}  As shown in \cref{tab:quanti-alnd-test}, we report the benchmarking results on our dataset. Our method further sets new SOTA records.  Compared to mask-free models specifically tailored for shadow removal \cite{jin2021dc, zhang2022spa} in the more complex ALN setting, our approach significantly outperforms them by over +1 dB PSNR. Additionally, when compared to other frequency-based methods \cite{fu2021dw, zhang2022spa, cui2022selective}, our work exhibits a considerable performance margin with over +0.7 dB PSNR. Notably, our convolution-based model achieves superior performance with fewer multiply-accumulate operations (MACs) compared to most transformer-based approaches \cite{wang2022uformer, zamir2022restormer, liang2021swinir}. While the second-best performing model, HINet \cite{Chen_2021_CVPR}, is also convolution-based, it incurs a higher computational cost. Conversely, our model requires approximately 40\% fewer MACs while achieving a +0.3 PSNR improvement over HINet. Finally, as observed from the table, none of the methods achieve a PSNR above 21 dB. This discrepancy is notable when compared to other restoration tasks, where state-of-the-art performances often exceed 30 dB PSNR. This indicates a substantial scope for further research in the field of ALN. We believe that our benchmark and dataset can serve as valuable resources to inspire and facilitate future research endeavors in this domain.

\begin{figure*}[t]
    \centering
    \setlength{\tabcolsep}{1pt}
    \renewcommand{\arraystretch}{0.7}
    \def\widthcomp{0.1639}
    \begin{tabularx}{\textwidth}{cccccc}
         Input                                                                          & SwinIR \cite{liang2021swinir}                                                       & UFormer \cite{wang2022uformer}                                                      & Restormer \cite{zamir2022restormer}                                                   & \emph{Ours}                                                                 & Ground Truth                                                                   \\
\includegraphics[width=\widthcomp\linewidth]{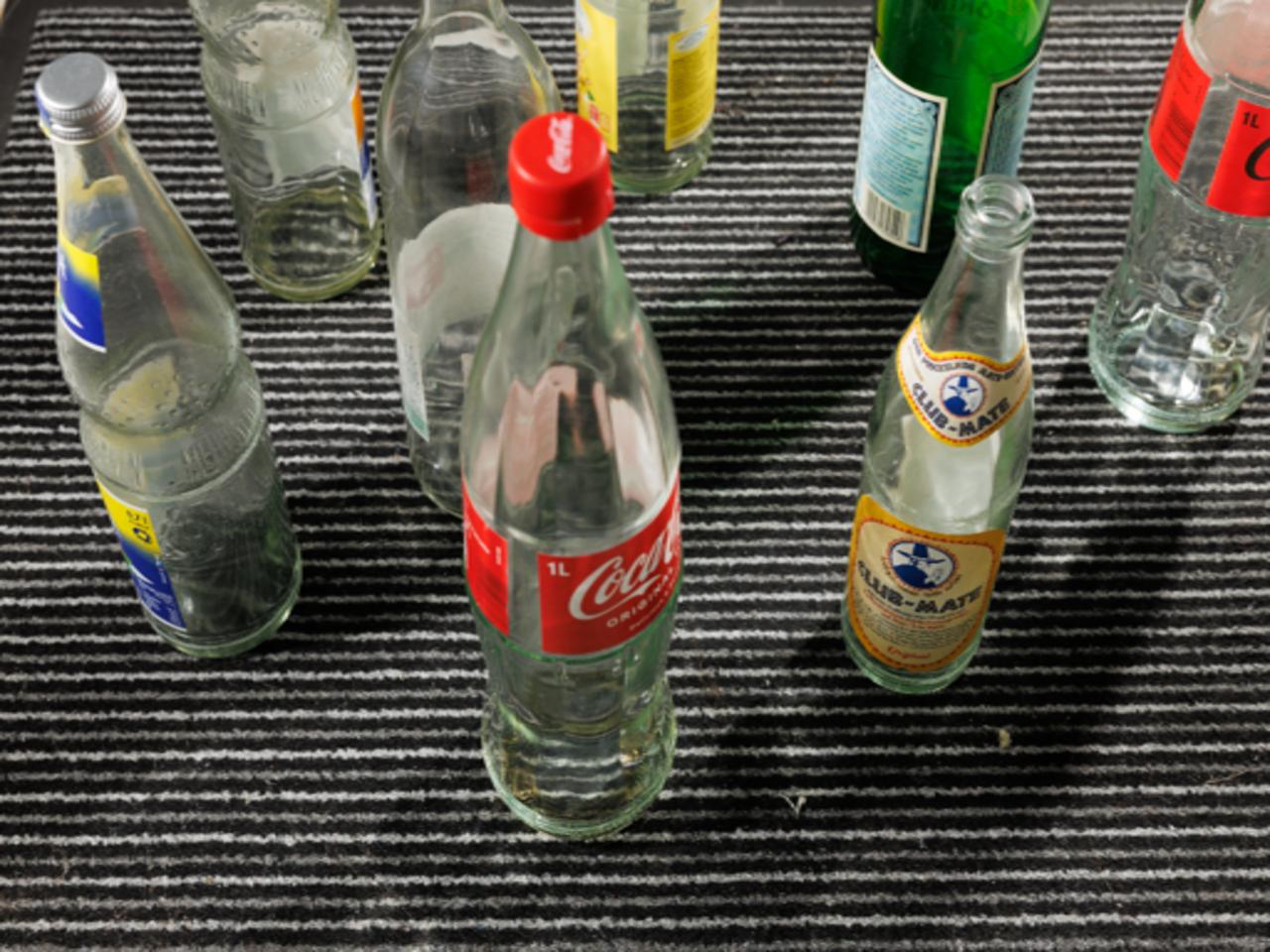} & \includegraphics[width=\widthcomp\linewidth]{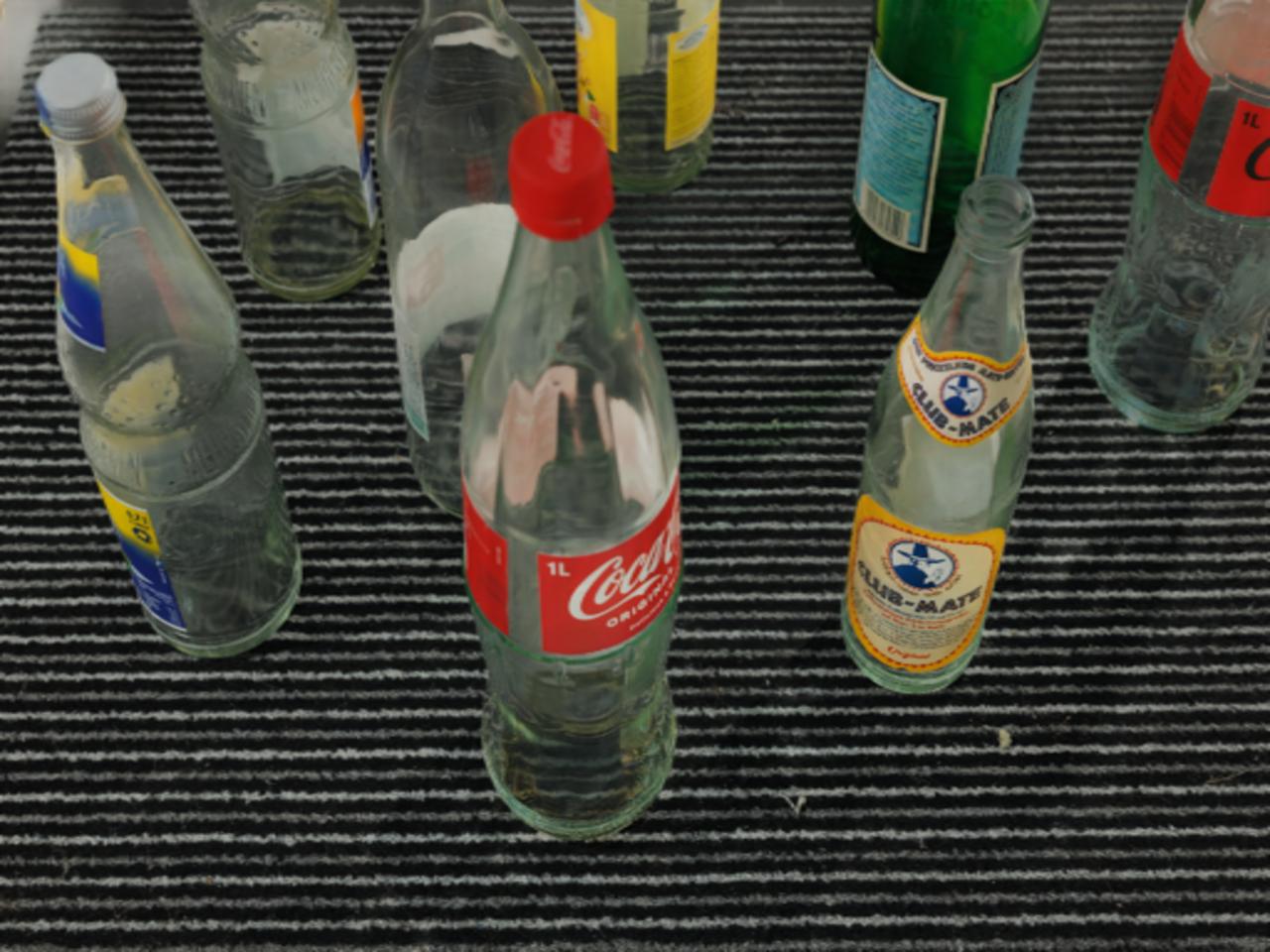} & \includegraphics[width=\widthcomp\linewidth]{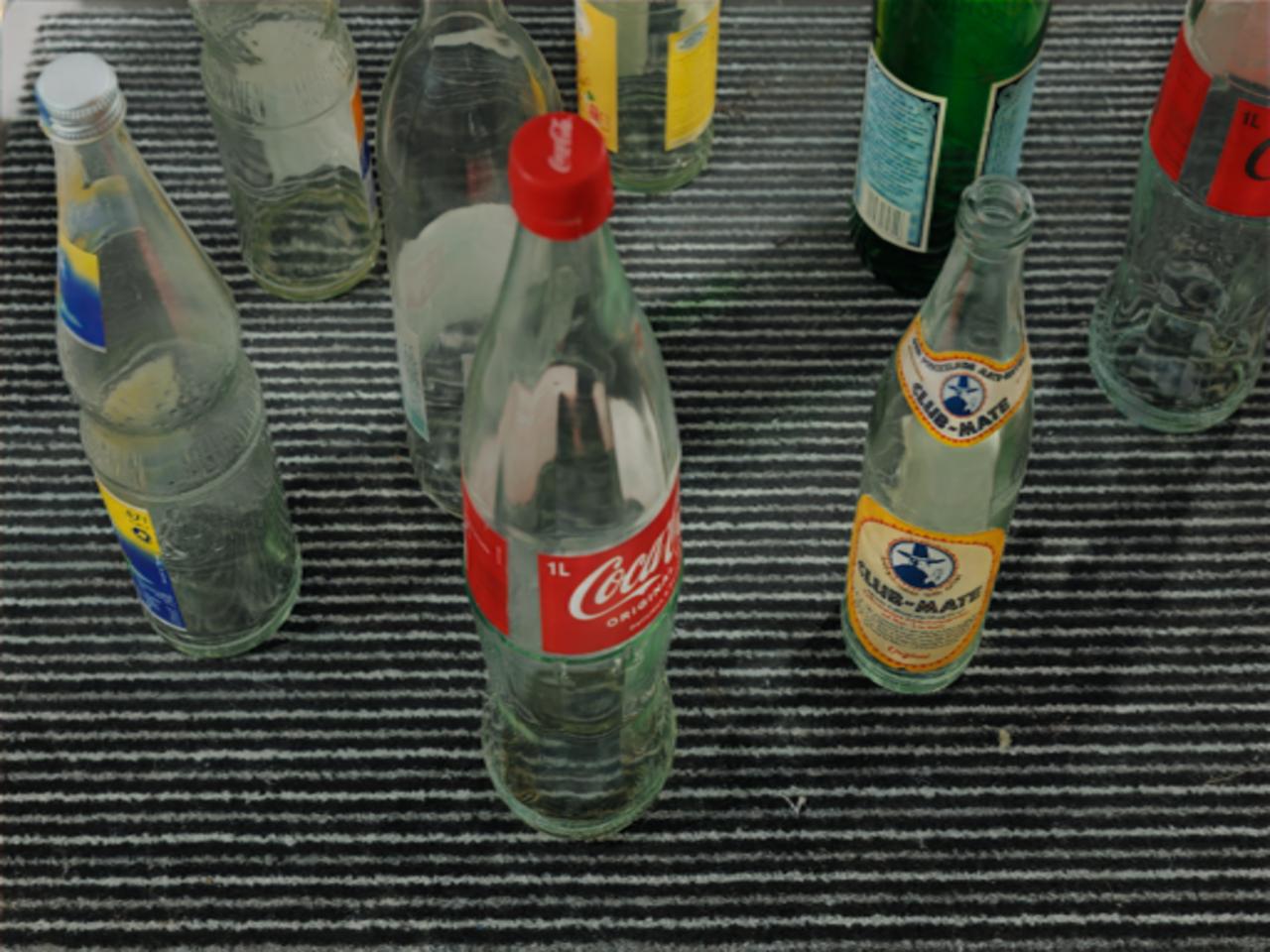} & \includegraphics[width=\widthcomp\linewidth]{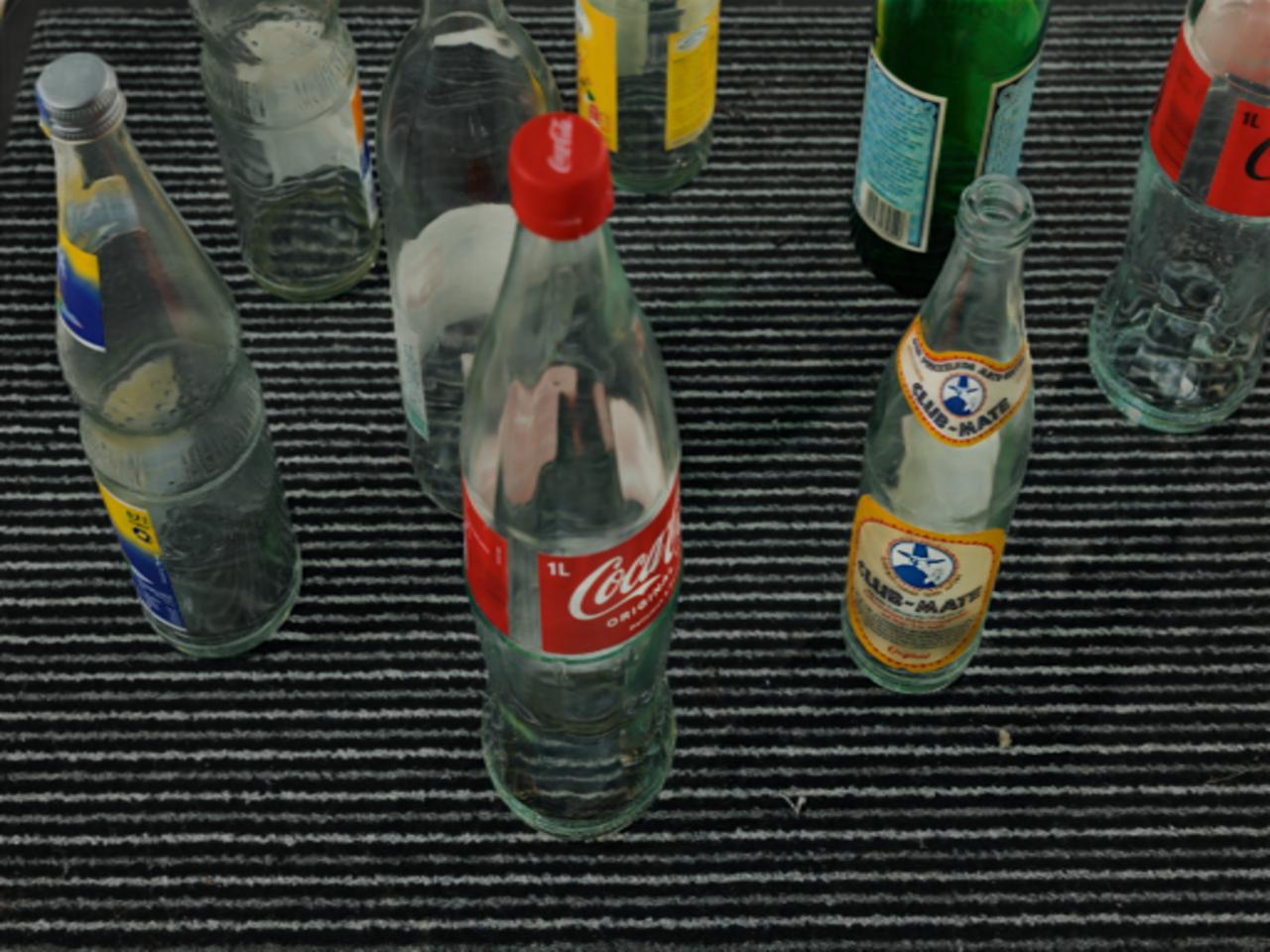} & \includegraphics[width=\widthcomp\linewidth]{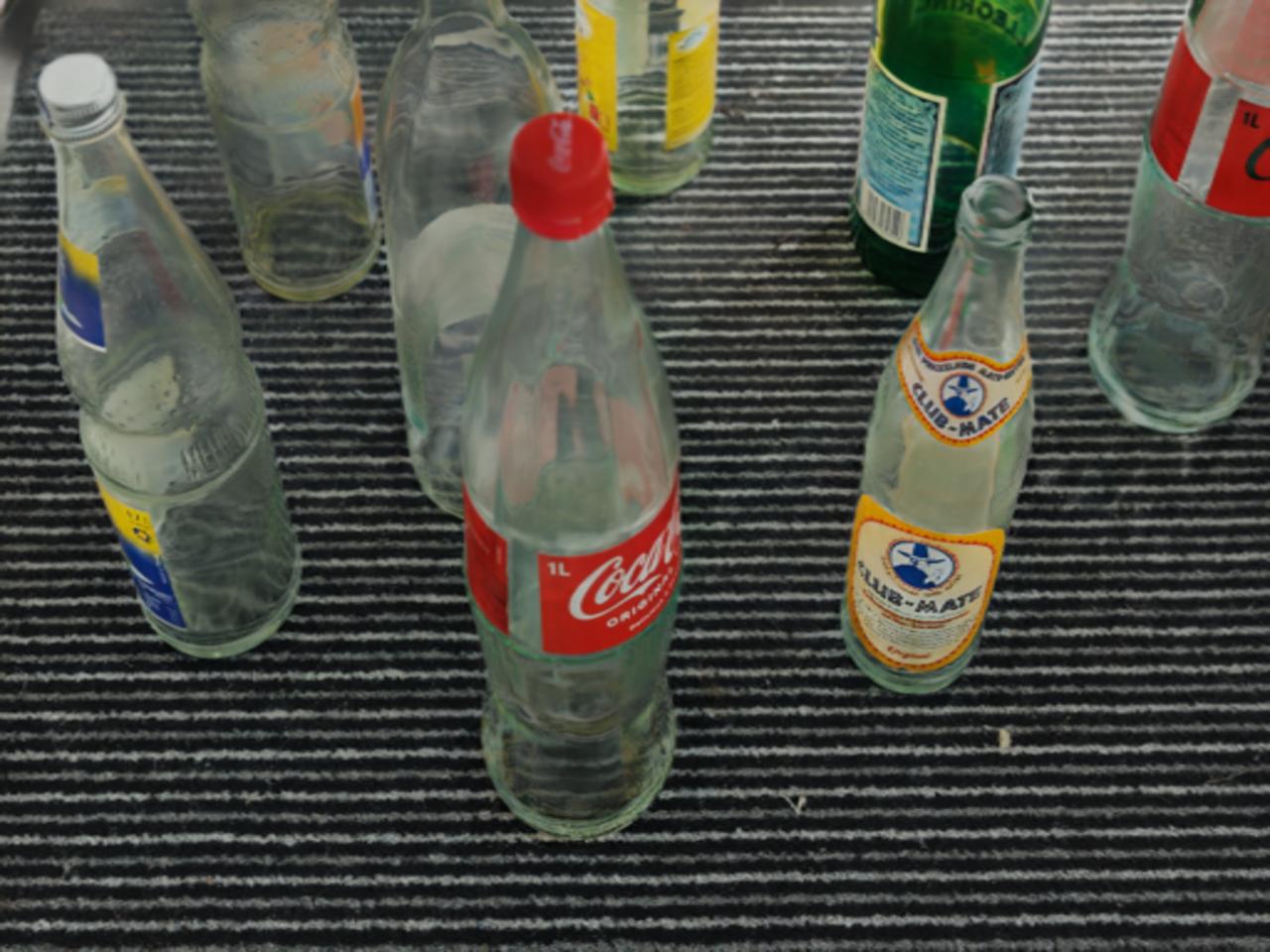} & \includegraphics[width=\widthcomp\linewidth]{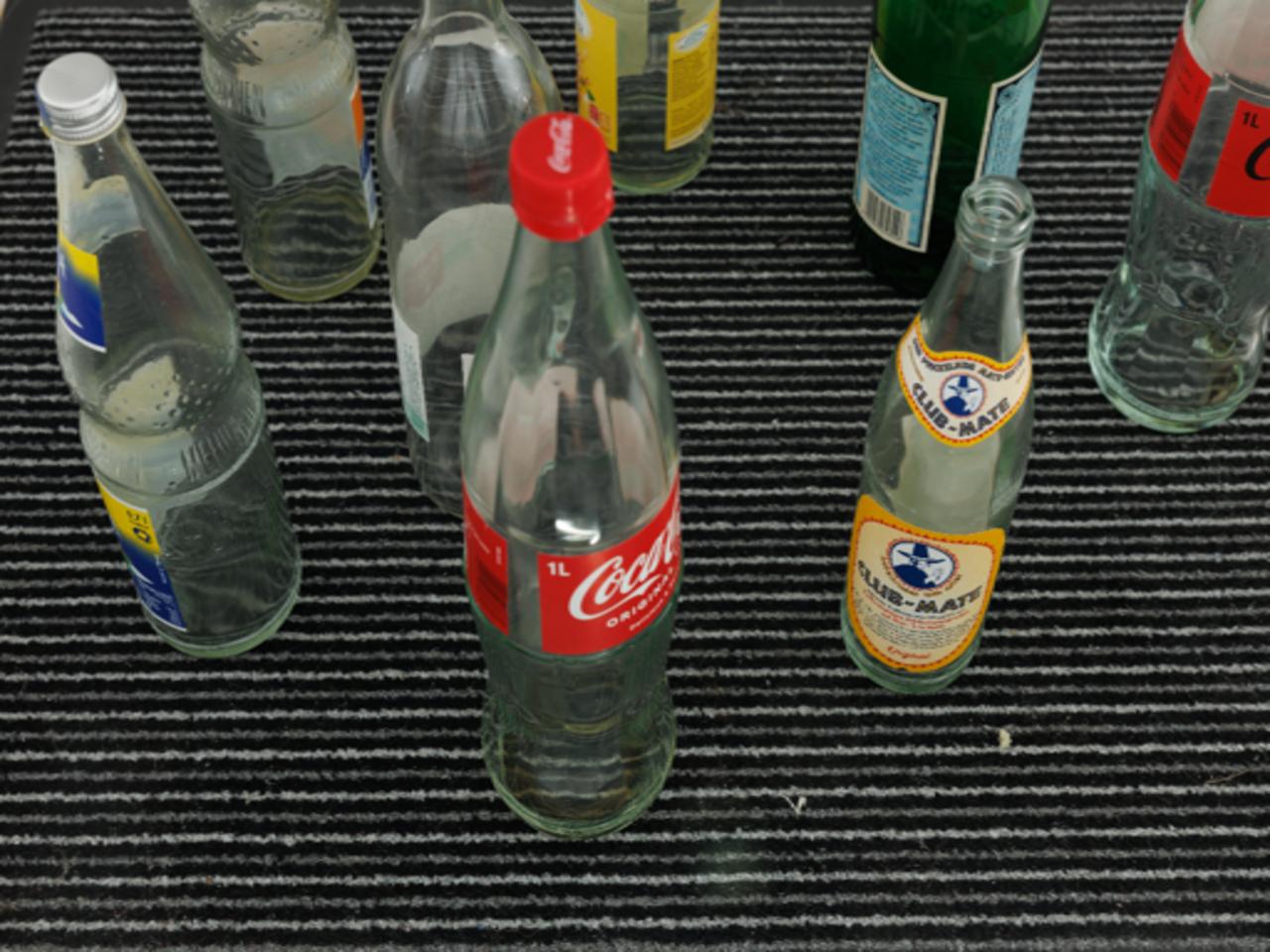} \\
\includegraphics[width=\widthcomp\linewidth]{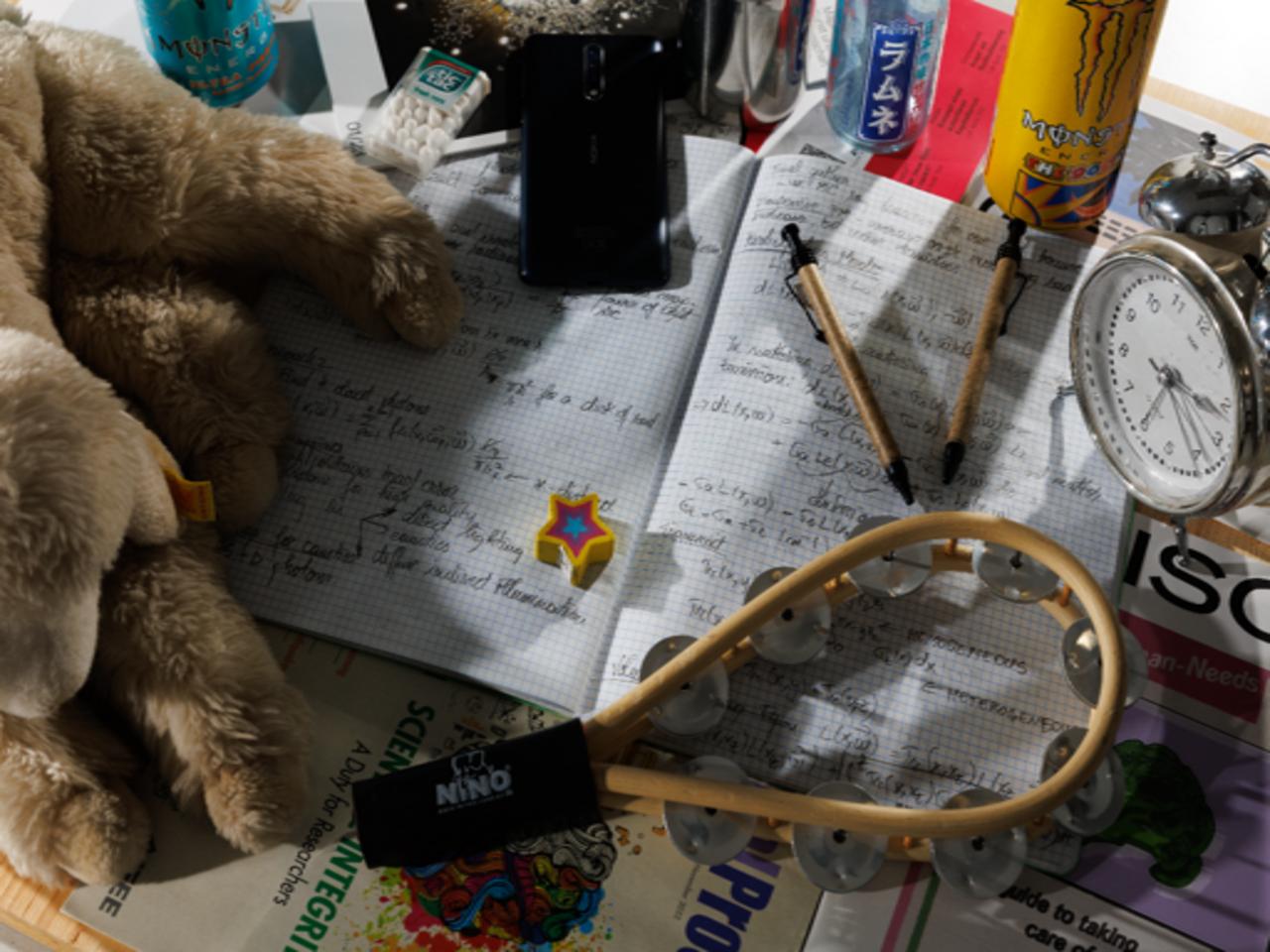} & \includegraphics[width=\widthcomp\linewidth]{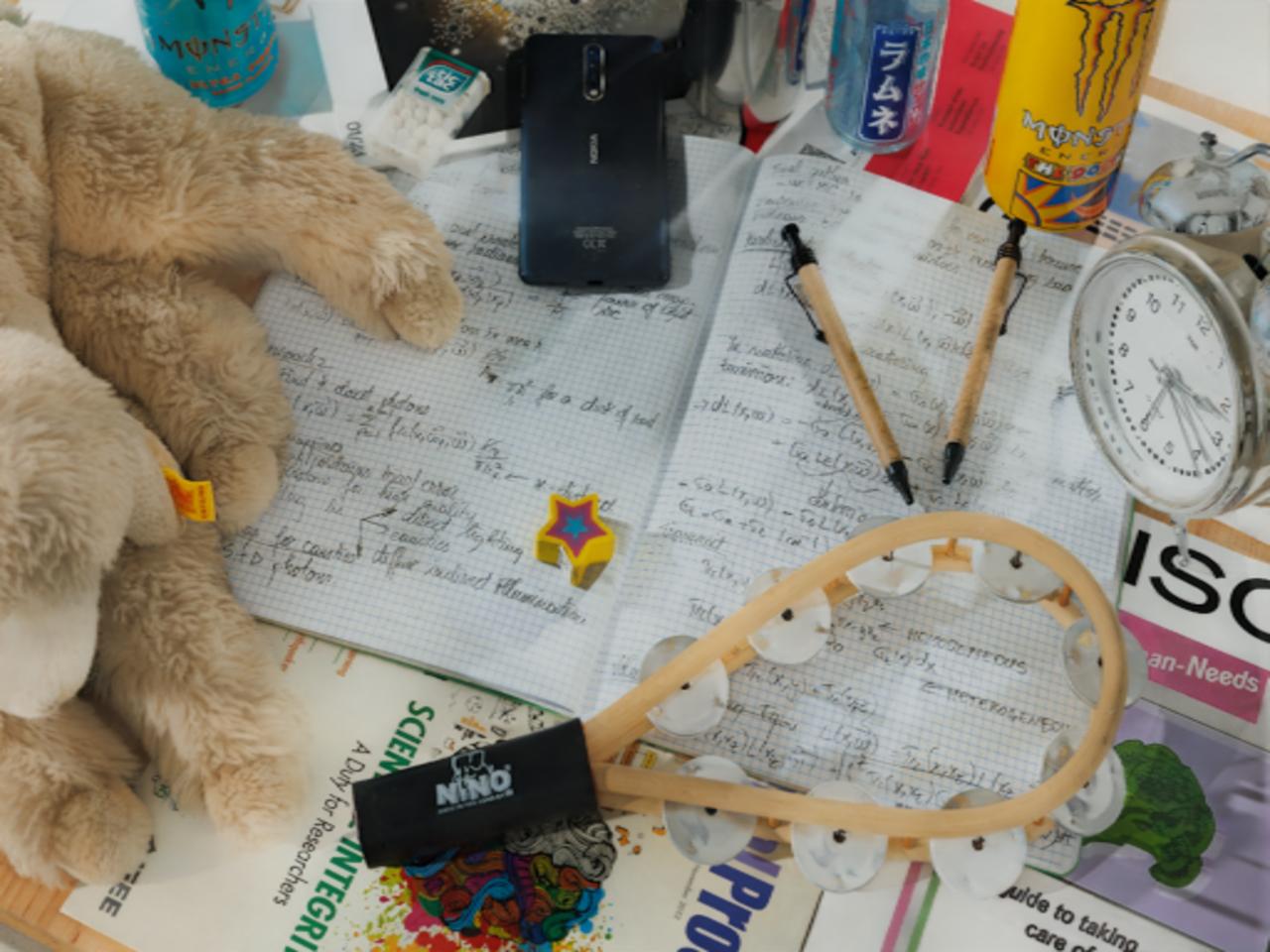} & \includegraphics[width=\widthcomp\linewidth]{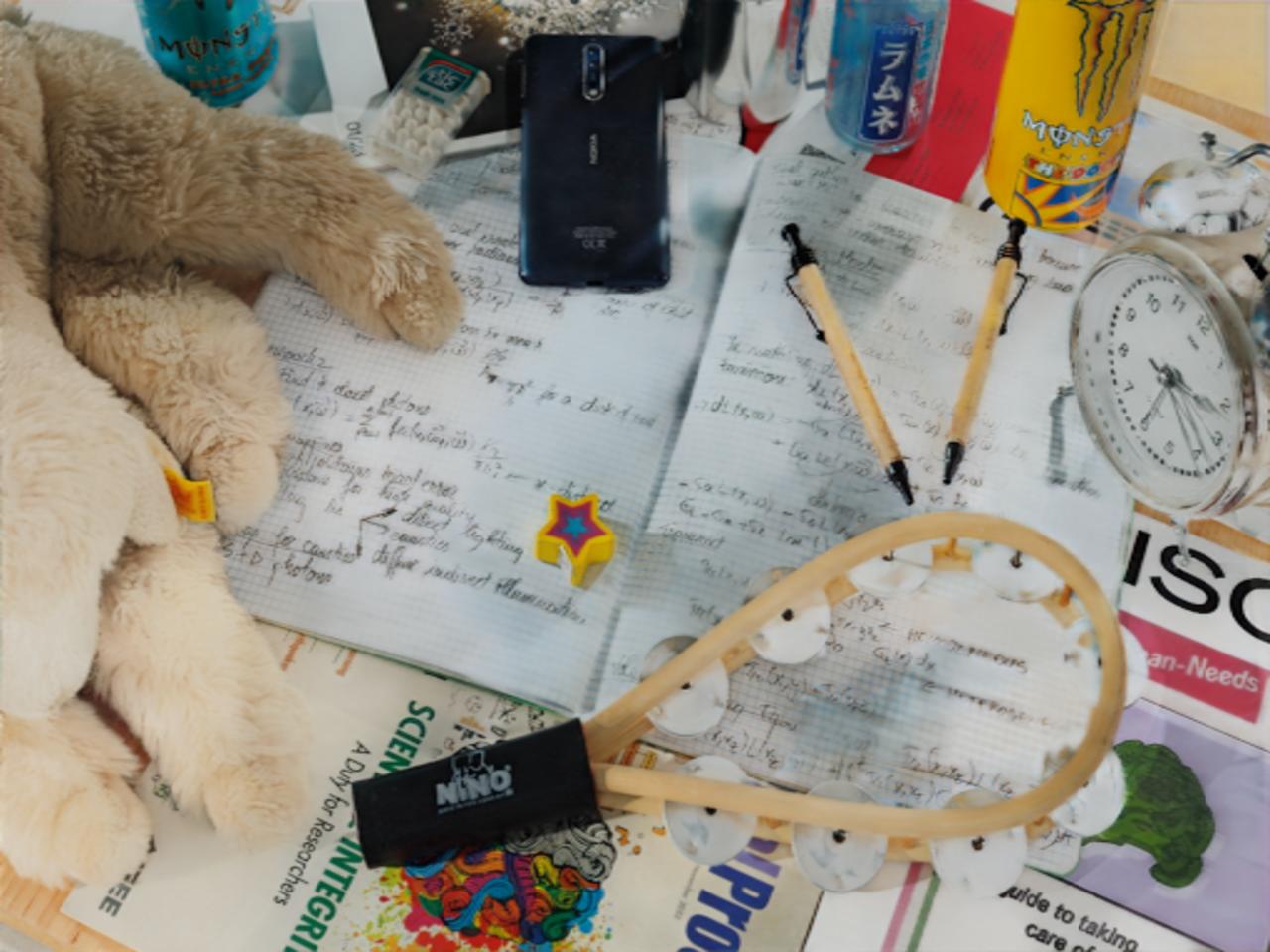} & \includegraphics[width=\widthcomp\linewidth]{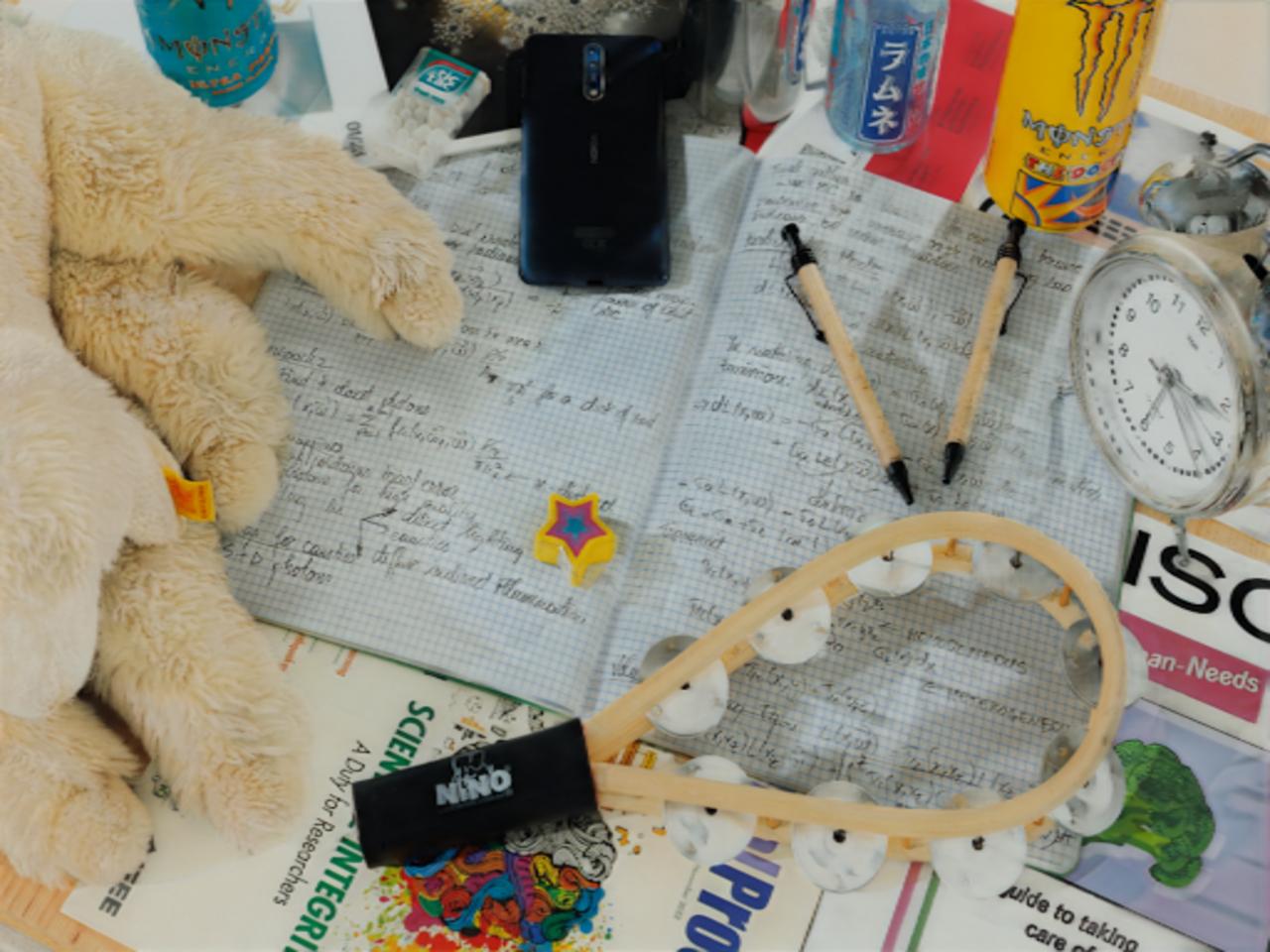} & \includegraphics[width=\widthcomp\linewidth]{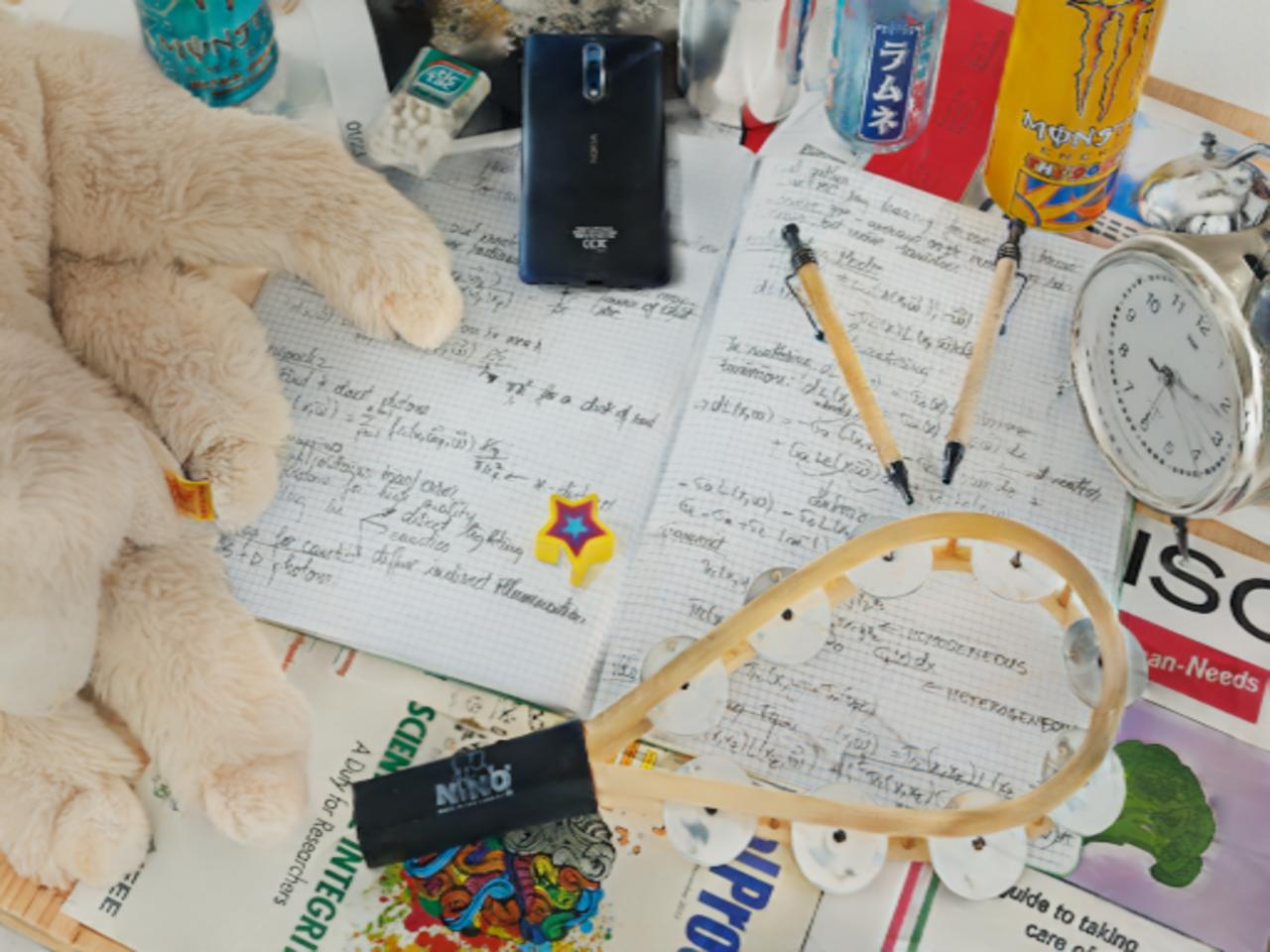} & \includegraphics[width=\widthcomp\linewidth]{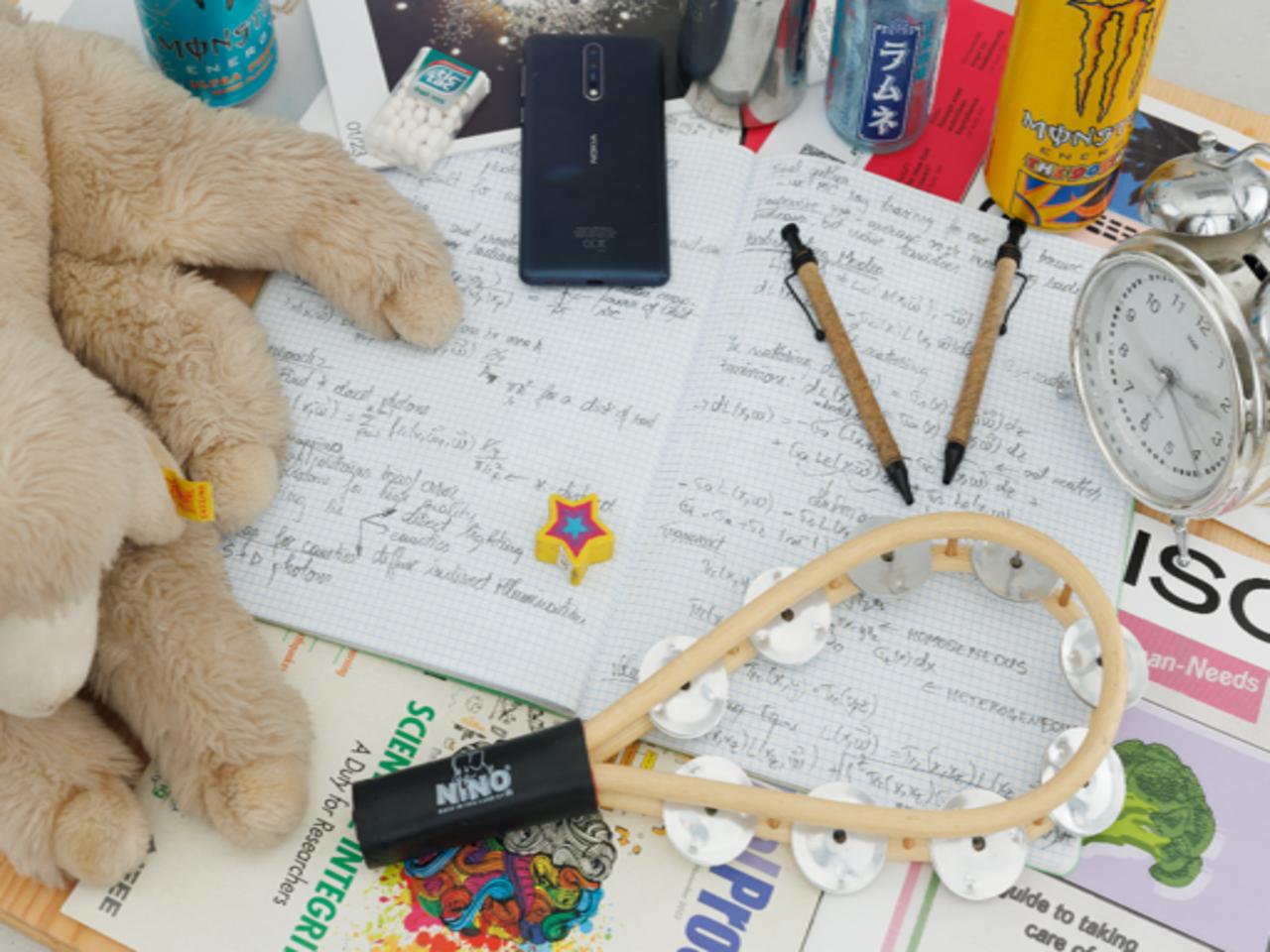} \\
\includegraphics[width=\widthcomp\linewidth]{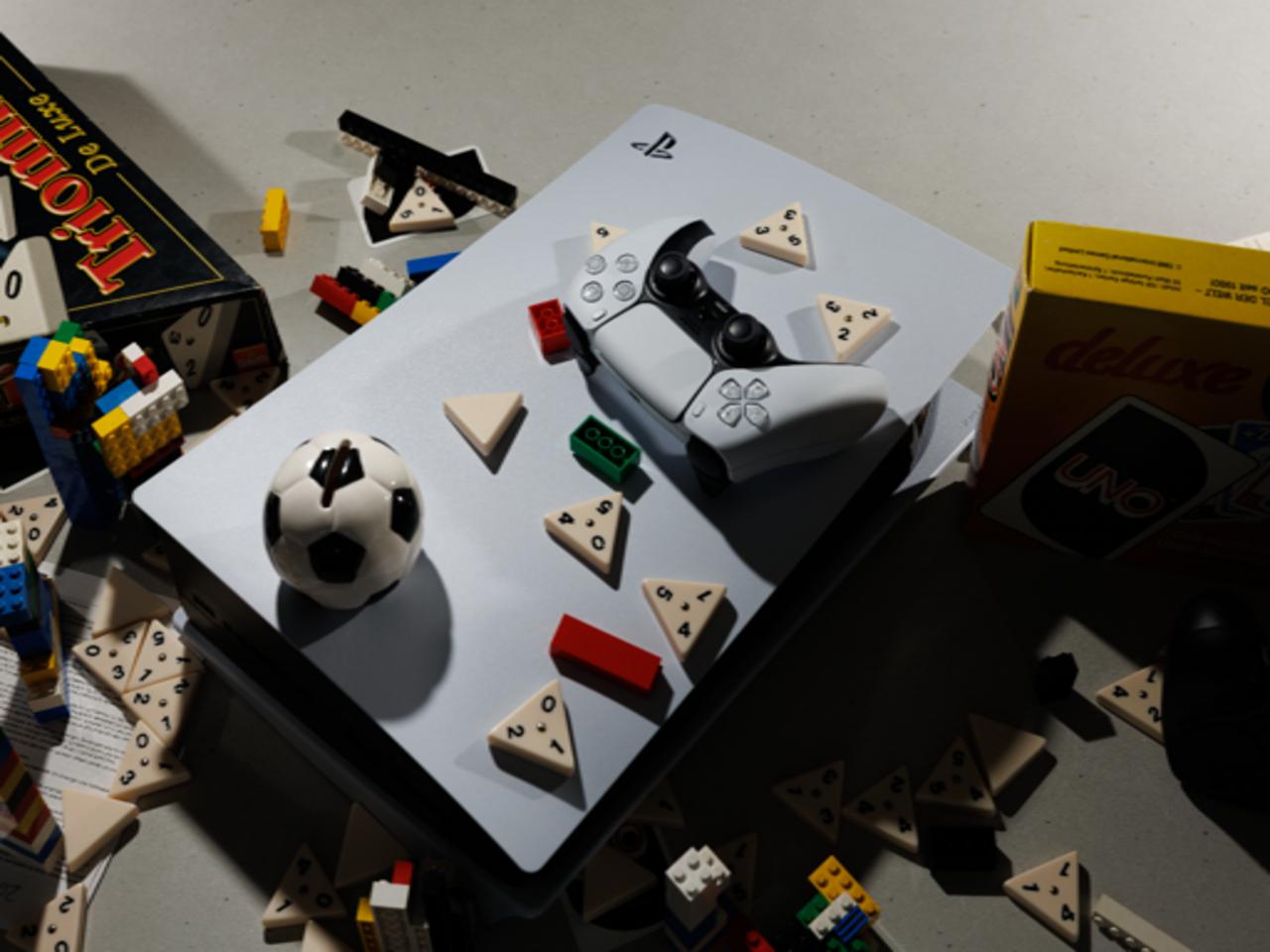} & \includegraphics[width=\widthcomp\linewidth]{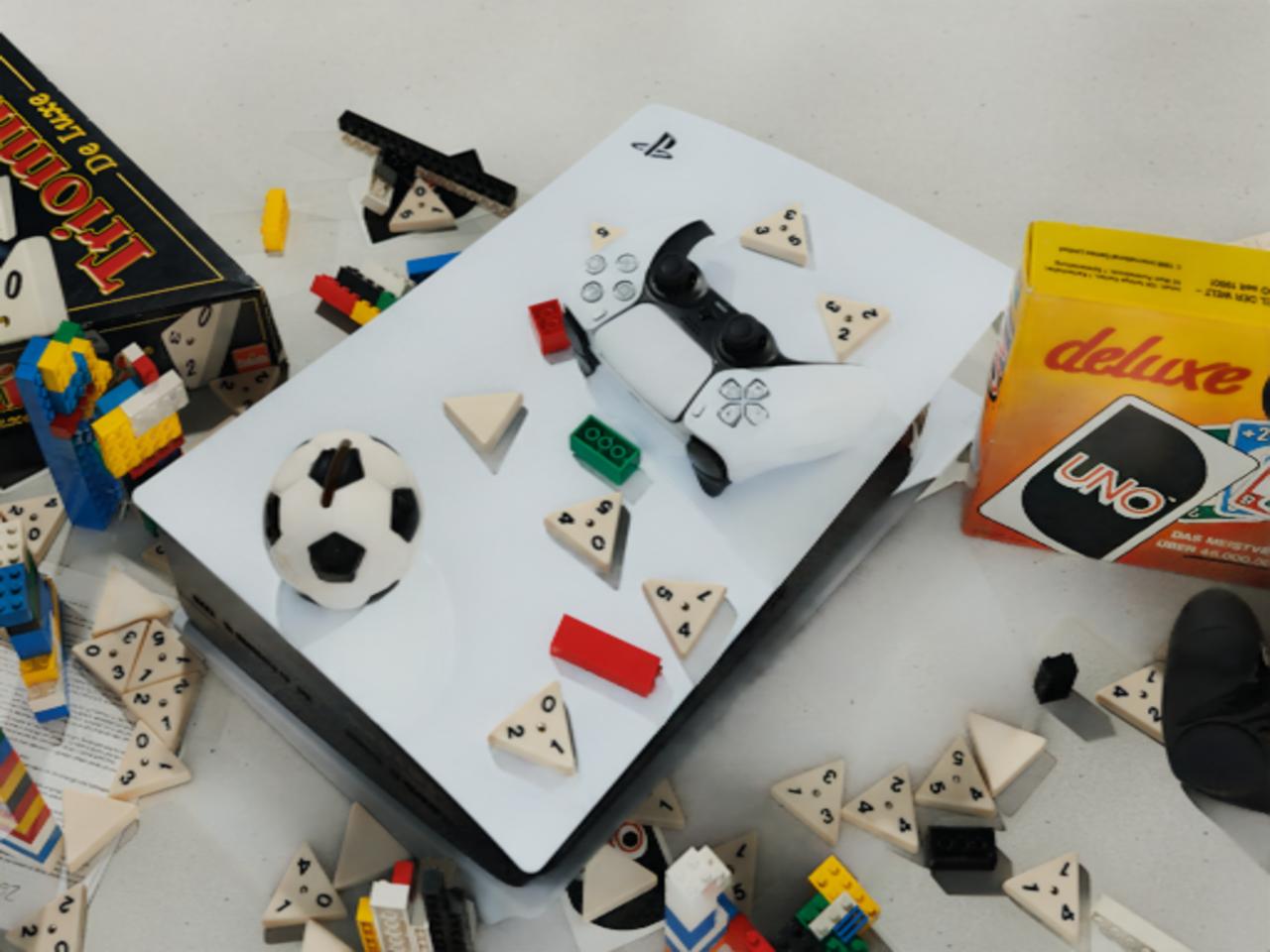} & \includegraphics[width=\widthcomp\linewidth]{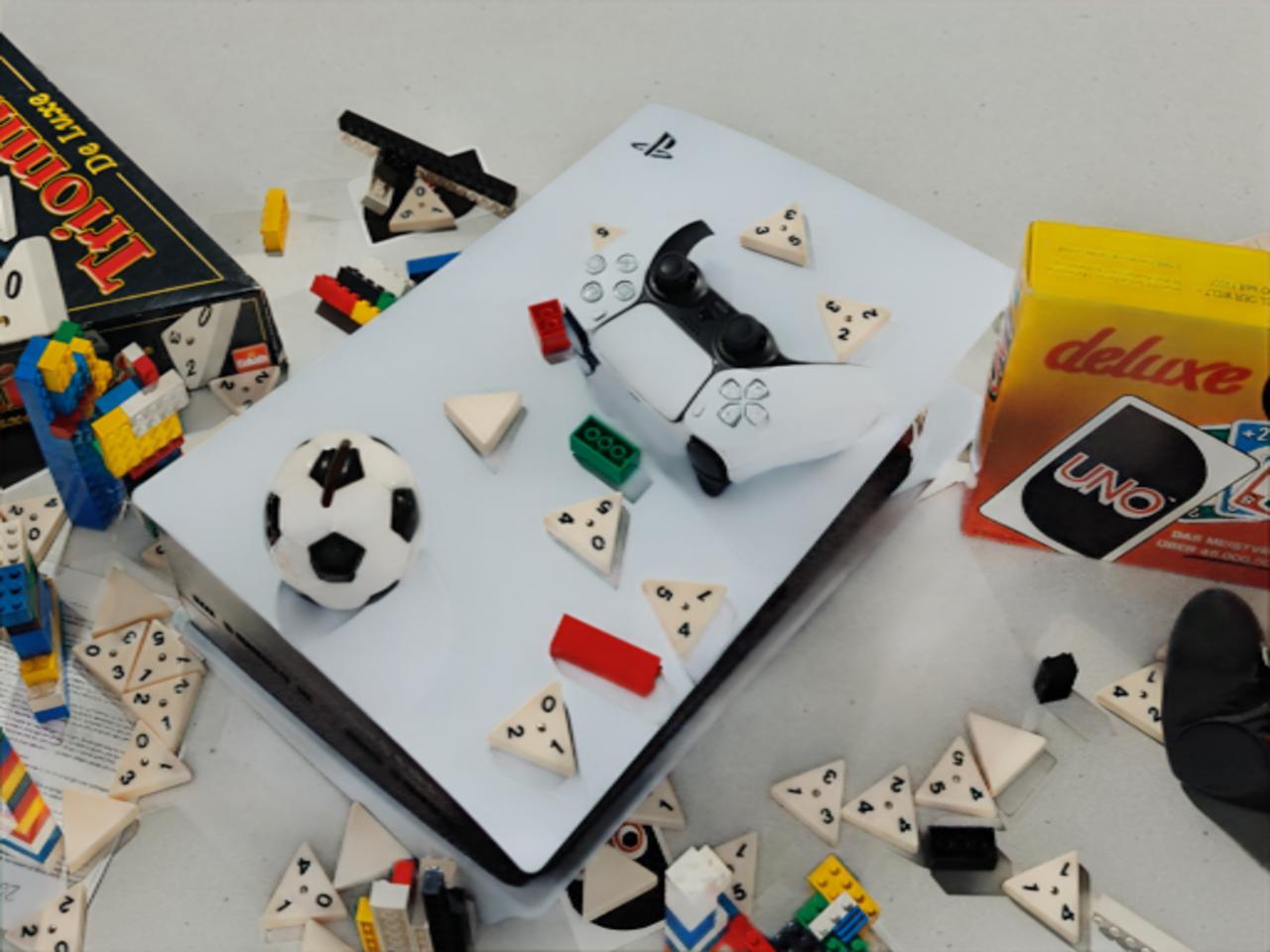} & \includegraphics[width=\widthcomp\linewidth]{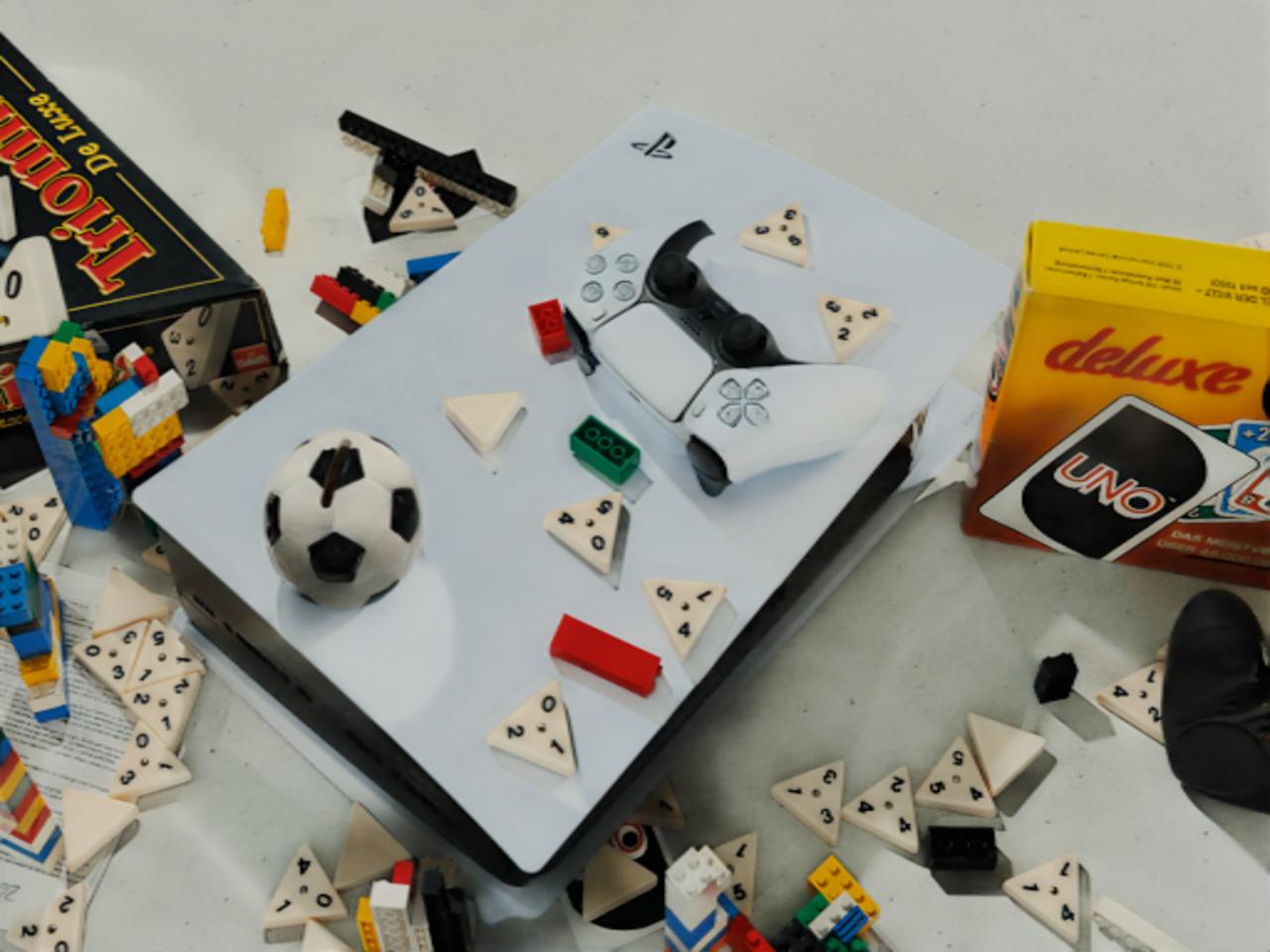} & \includegraphics[width=\widthcomp\linewidth]{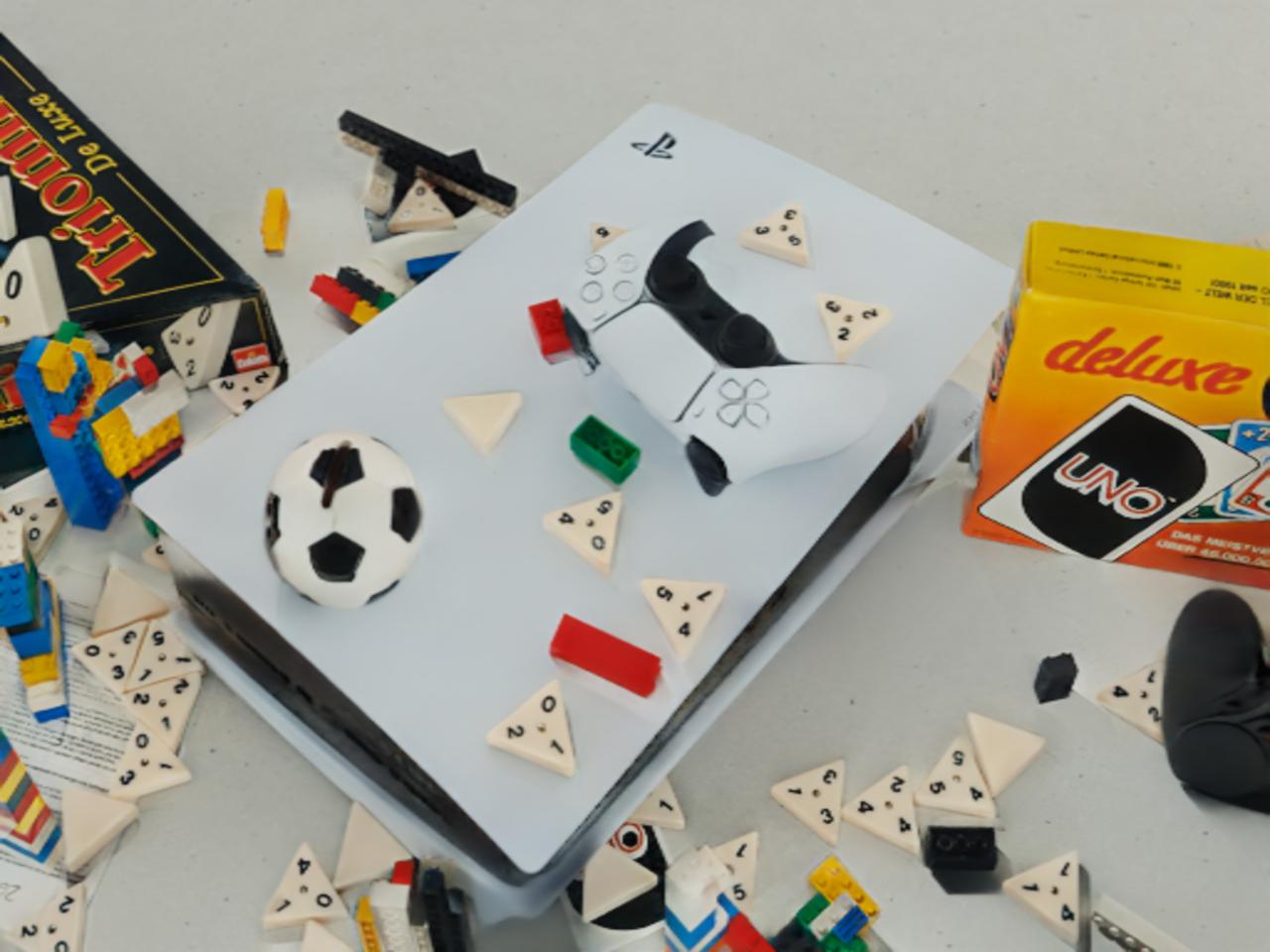} & \includegraphics[width=\widthcomp\linewidth]{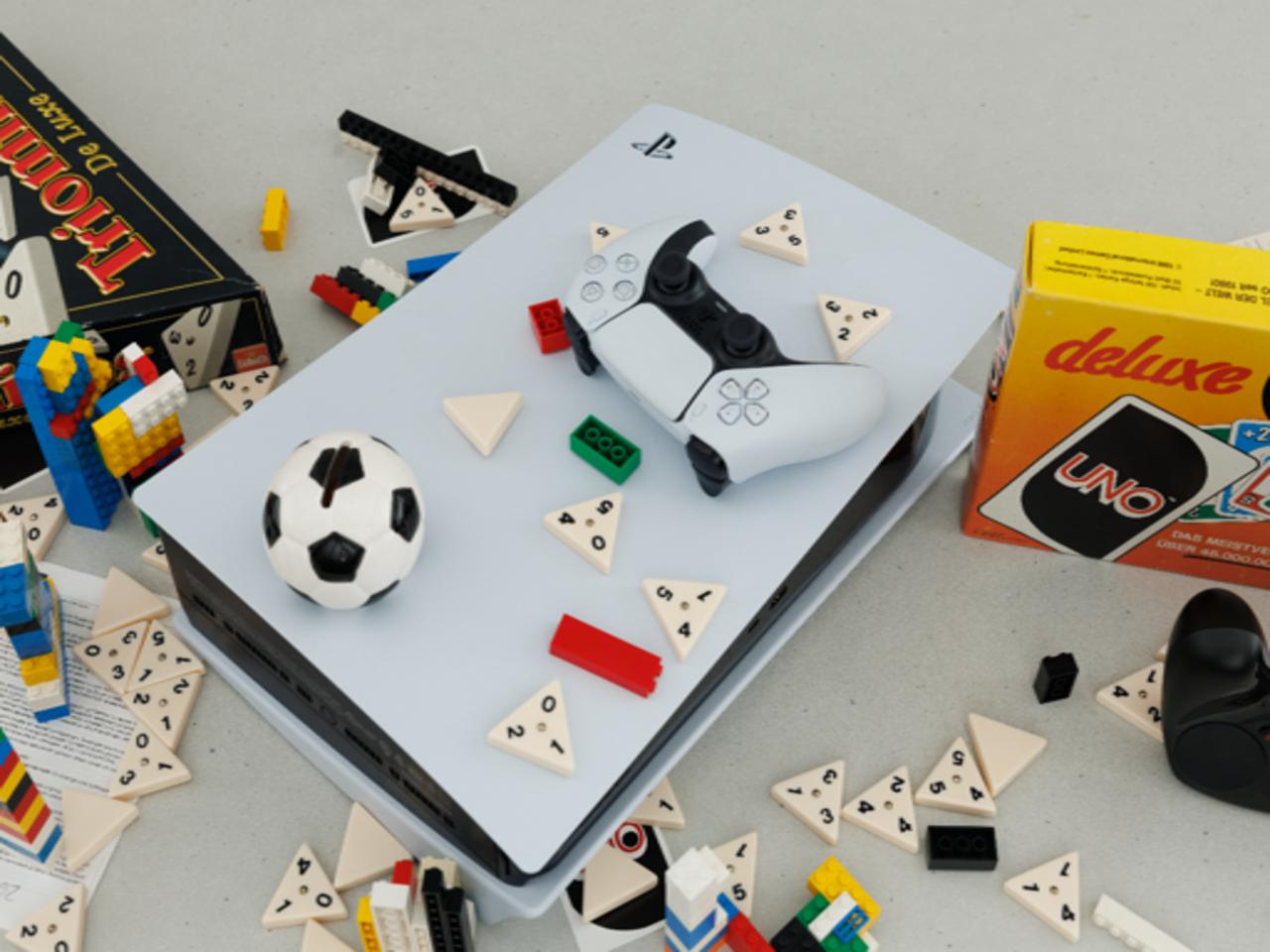} \\
\includegraphics[width=\widthcomp\linewidth]{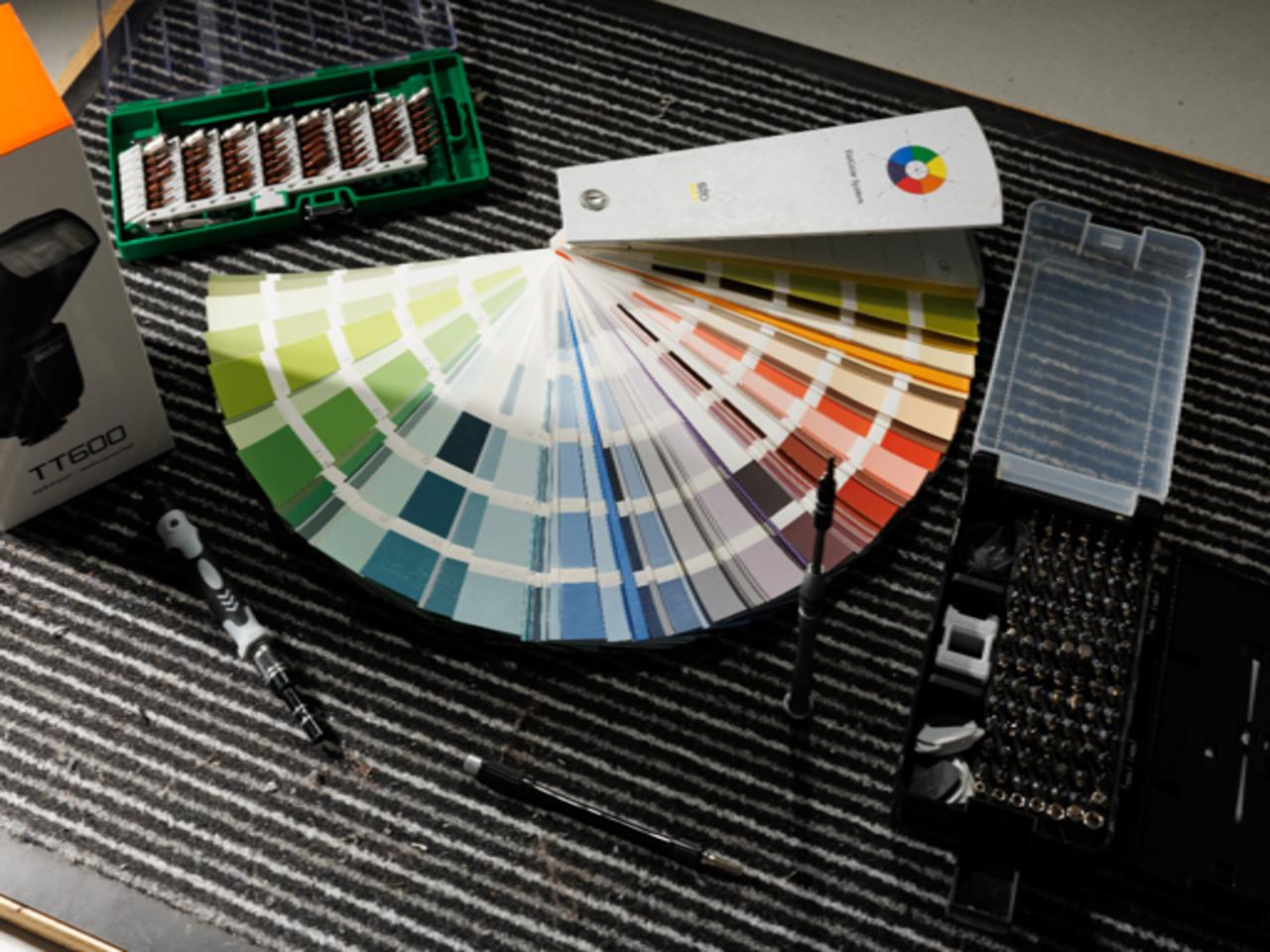} & \includegraphics[width=\widthcomp\linewidth]{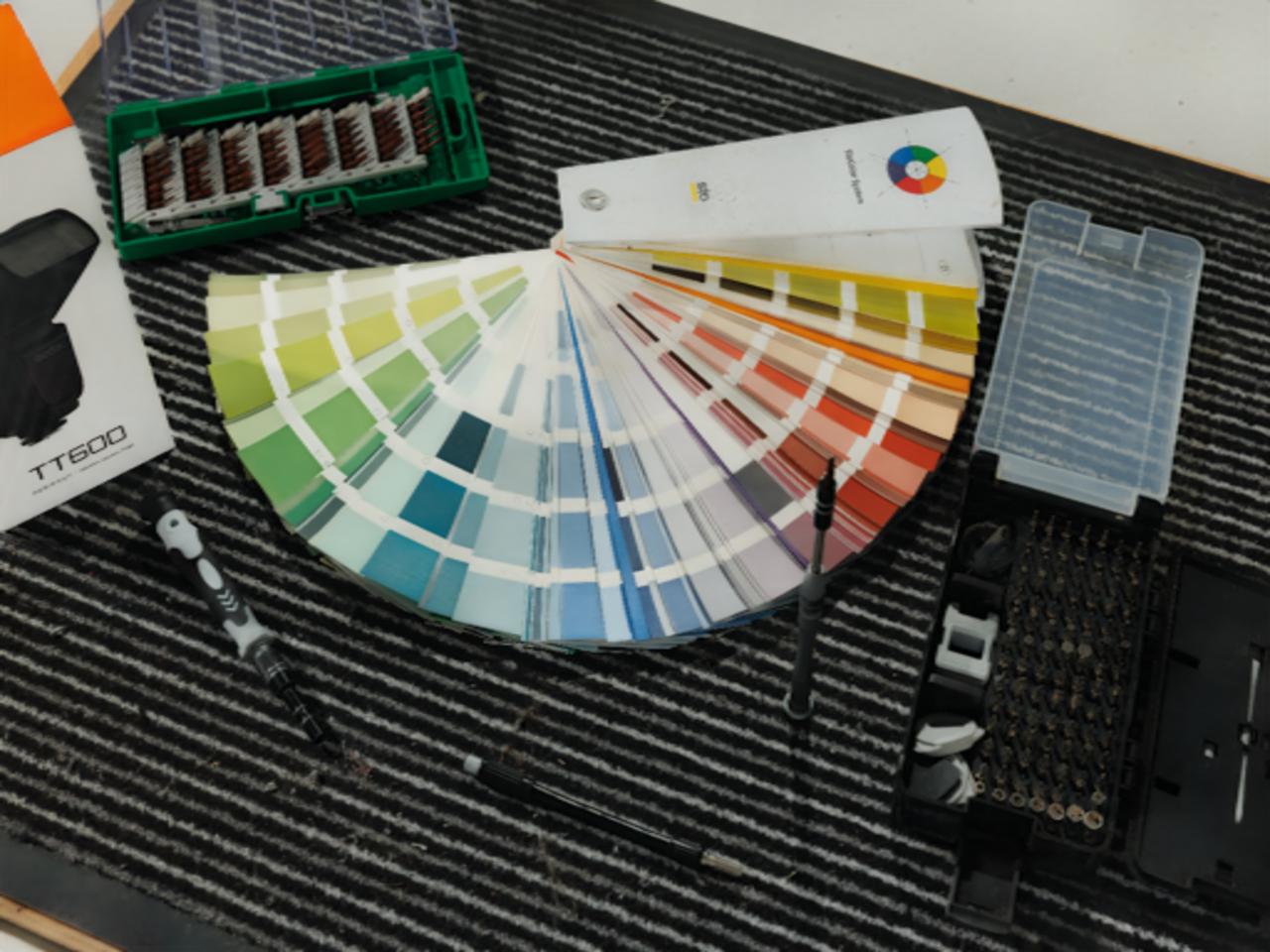} & \includegraphics[width=\widthcomp\linewidth]{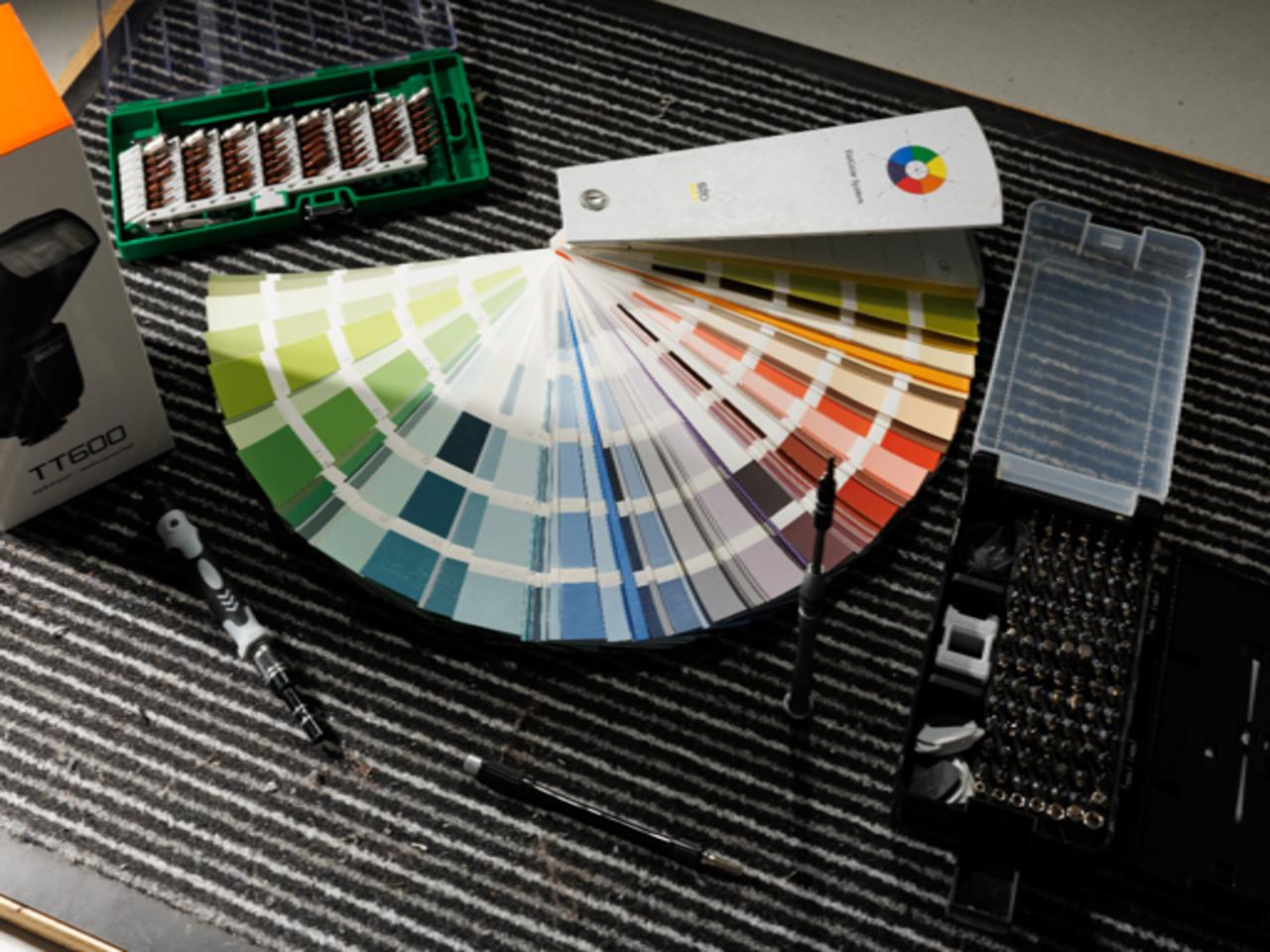} & \includegraphics[width=\widthcomp\linewidth]{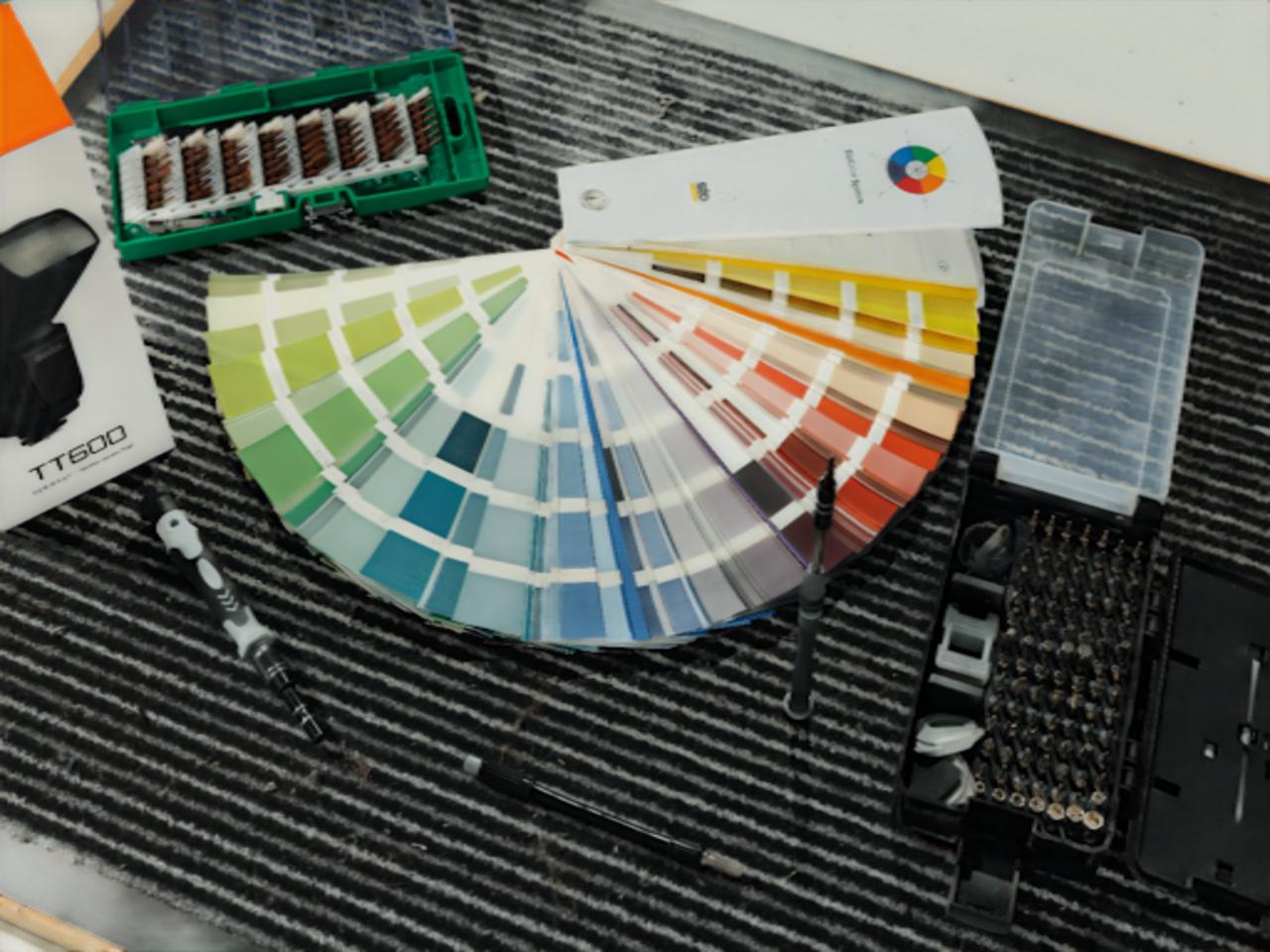} & \includegraphics[width=\widthcomp\linewidth]{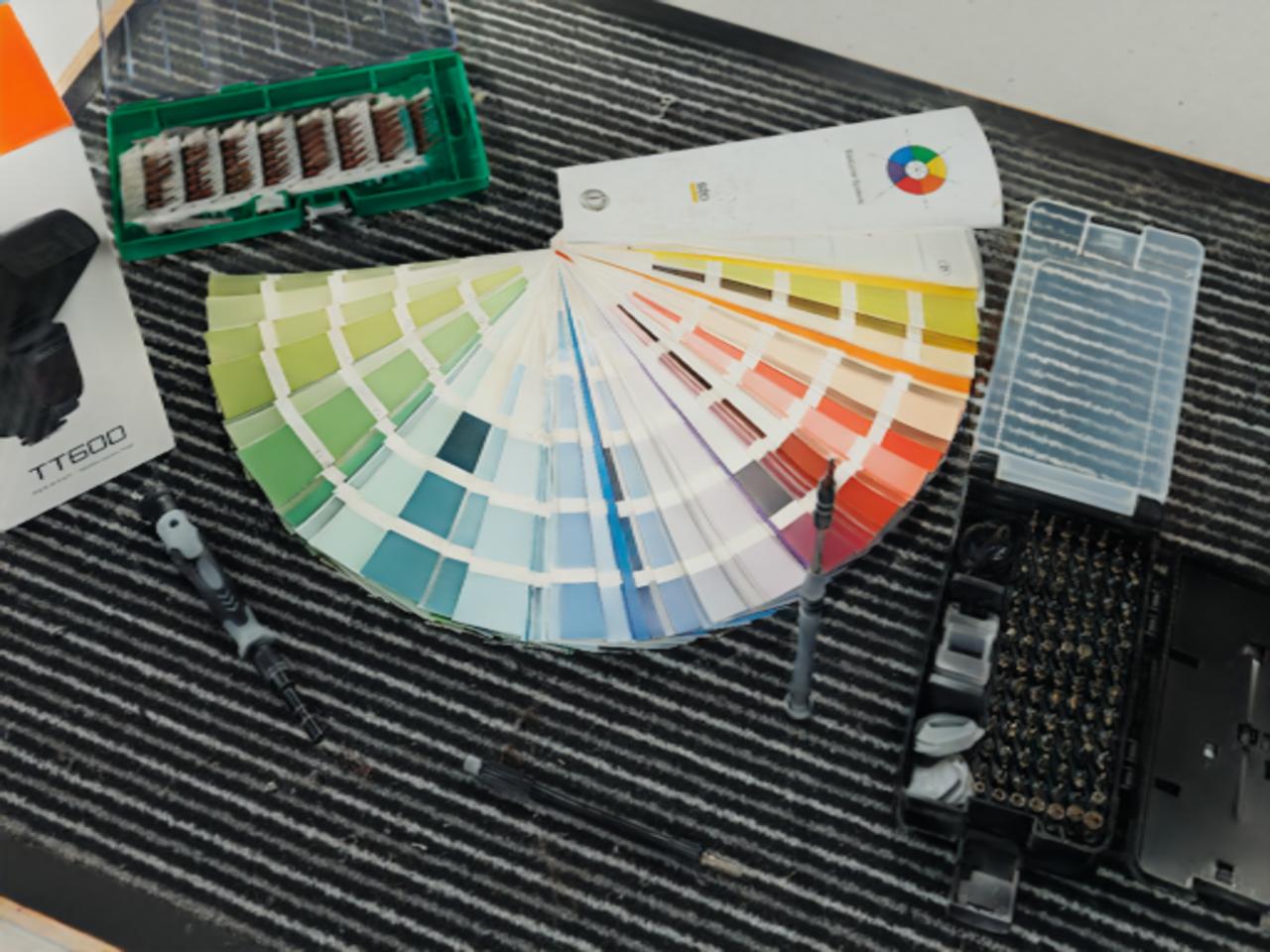} & \includegraphics[width=\widthcomp\linewidth]{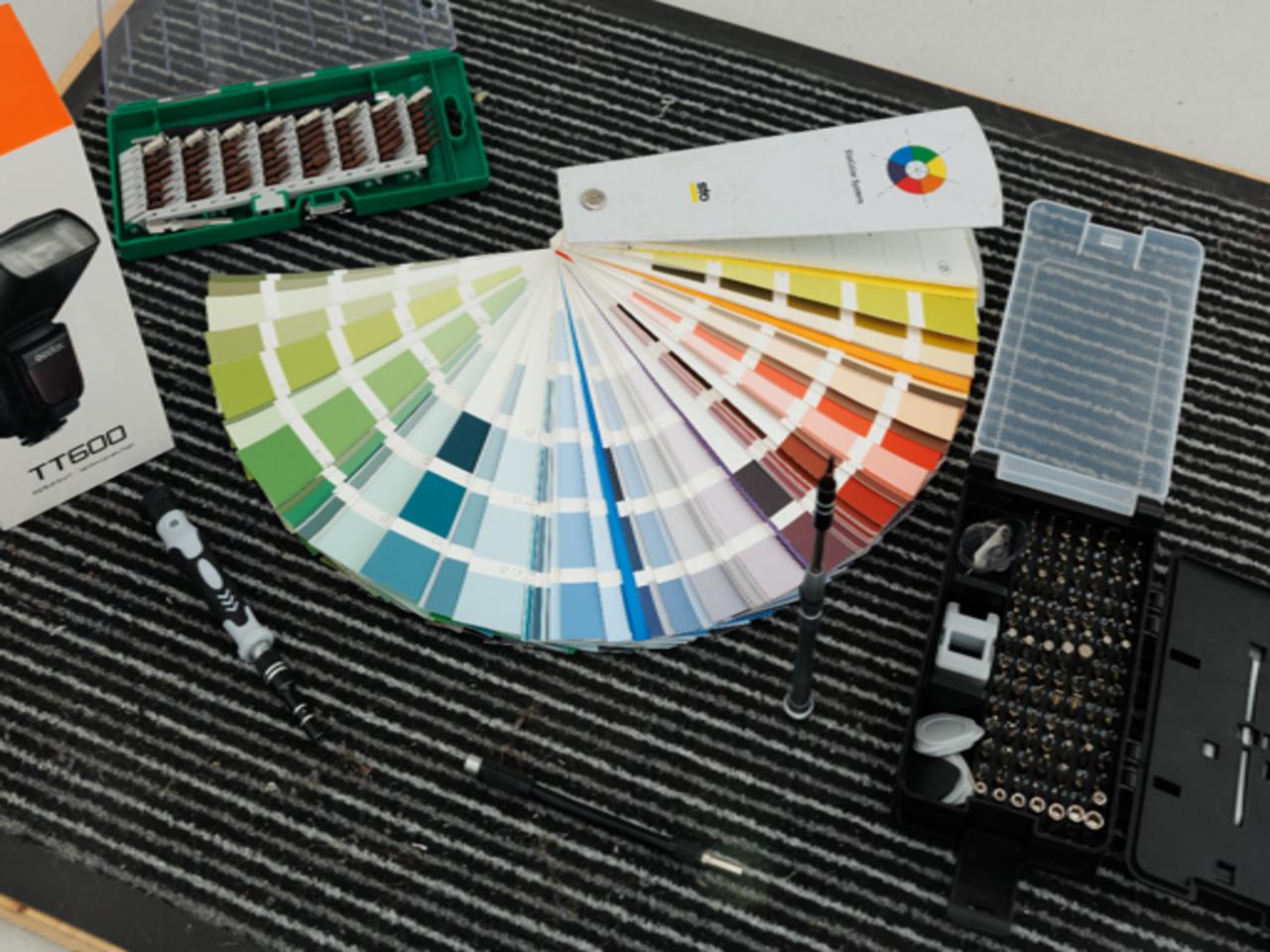}
    \end{tabularx}
    \vspace{-4mm}
    \caption{\textbf{Qualitative comparison }on challenging samples from the Ambient6K testing set. Our IFBlend achieves better light normalization and outperforms its counterparts.}
    \label{fig:qualitative-asdn}
    \vspace{-3mm}
\end{figure*}

\noindent \textbf{Qualitative Comparison:}  In \cref{fig:qualitative-asdn}, we present a qualitative comparison against established restoration methods on various challenging samples from our Ambient6K dataset. It can be seen that these methods struggle to accurately restore information across all local regions, often leaving some areas either under-normalized or over-normalized. This limitation is primarily attributed to the network design. As depicted in \cref{fig:teaser}, conventional restoration methods are optimized for global restoration tasks like deraining and dehazing, where the degradation is overall uniformly distributed across the entire image. However, in ALN, while global restoration is necessary, the degradation exhibits significant and unavoidable local characteristics. This complexity results in varying restoration levels across different regions, making the ALN task challenging. This challenging nature can be further supported by the lower PSNR values in \cref{tab:quanti-alnd-test}. In contrast, our IFBlend model employs a shrinkage image-frequency fusion strategy, allowing for a more nuanced understanding of different regions. Consequently, it selectively normalizes the lighting on a region-by-region basis, yielding output closer to ground truth.

\subsection{Comparison on Conventional Shadow Removal Benchmarks}

\noindent \textbf{Metrics:} For the comparison on the shadow removal established ISTD family benchmarks \cite{ISTDwang2018STCGAN, Le_2019_ICCV}, as it became a standard in the public literature, we report the RMSE computed in the Lab image space, for the global image and the segments defined by the binary shadow mask. Since the only publicly available set of masks for the SRD dataset is generated by DHAN~\cite{cun2019ghostfree}, we only report statistics regarding the global image, in terms of fidelity and perceptual properties observed in the RGB image space.

\noindent \textbf{Compared methods:} As the shadow mask is consistently provided in shadow removal benchmarks, we first compare against the top-performing methods that rely on masks, such as STCGAN~\cite{ISTDwang2018STCGAN}, SP-M Net \cite{Le_2019_ICCV}, DHAN \cite{cun2019ghostfree}, PULSr \cite{vasluianu2021shadow}, DNSR \cite{vasluianu2023ntire}, AEF \cite{fu2021auto}, ShadowFormer \cite{guo2023shadowformer}, and UNSR \cite{zhu2022efficient}. We also include two mask-free approaches, DCShadowNet \cite{jin2021dc} and DeS3 \cite{jin2023des3}, which solely rely on image inputs without any mask prior. Additionally, recognizing shadow removal as a specific application of image restoration, we retrain three mainstream restoration methods: SwinIR \cite{liang2021swinir}, Uformer \cite{wang2022uformer}, and Restormer \cite{zamir2022restormer}.

\begin{table}[t]
\centering
%\tiny
\setlength{\tabcolsep}{2pt}
\caption{Quantitative comparison on the ISTD and ISTD+ datasets \cite{ISTDwang2018STCGAN, Le_2019_ICCV}, in terms of Lab space RMSE. \textbf{Best results are in bold} for mask-based and mask-free methods.}
\vspace{-3mm}
\resizebox{\linewidth}{!}
{
\begin{tabular}{l|cc|ccc|ccc}
\toprule
Method & Evaluation     & Mask  & \multicolumn{3}{c|}{ISTD \cite{ISTDwang2018STCGAN}}                     & \multicolumn{3}{c}{ISTD+\cite{ISTDwang2018STCGAN, Le_2019_ICCV}}       \\
name                                    & Resolution     & Free   & \text{Shadow} & \text{Shadow free} & \text{Total} & \text{Shadow} & \text{Shadow free} & \text{Total} \\ \midrule
unprocessed                             & 640$\times$480 & -      & 15.07                & 3.86                      & 6.80         & 17.53                & 1.82                      & 7.15         \\
ST-CGAN\cite{ISTDwang2018STCGAN}         & 256$\times$256 & \xmark & 4.83                 & 3.44                      & 4.05         & n/a                  & n/a                       & n/a          \\
SP-M Net \cite{Le_2019_ICCV}            & 512$\times$512 & \xmark & n/a                  & n/a                       & n/a          & 4.79                 & 4.27                      & 4.37         \\
DHAN \cite{cun2019ghostfree}            & 640$\times$480 & \xmark & 4.65                 & 3.13                      & 3.43         & 4.04                 & 2.97                      & 3.19         \\
PULSr \cite{vasluianu2021shadow}        & 512$\times$512 & \xmark & 4.48                 & 3.03                      & 3.33         & 4.12                 & 2.39                      & 2.82         \\
DNSR \cite{Vasluianu_2023_WSRD}         & 640$\times$480 & \xmark & 4.39                 & 2.47                      & 2.84         & 3.92                 & 1.80                      & 2.24         \\
AEF \cite{fu2021auto}                   & 256$\times$256 & \xmark & 3.75                 & 2.79                      & 3.10          & 3.23                 & 2.05                      & 2.31         \\
ShadowFormer \cite{guo2023shadowformer} & 640$\times$480 & \xmark & \textbf{3.25}        & \textbf{2.38}            & \textbf{2.43} & \textbf{2.93}               & \textbf{1.66}                   & \textbf{1.93}         \\ \midrule
DCShadowNet \cite{jin2021dc}           & 640$\times$480 & \cmark & 5.97                 & 3.62                      & 4.06             & 10.30                 & 3.50                       & 4.60          \\
DeS3  \cite{jin2023des3}                & 640$\times$480 & \cmark & n/a                    & n/a                        & n/a             & 4.90                  & 2.70                       & 2.80          \\
SwinIR \cite{liang2021swinir}           & 640$\times$480 & \cmark & 4.84                 & 3.44                     &  3.70         & 4.54                     & 2.44               & 3.00             \\
Uformer \cite{wang2022uformer}          & 512$\times$512 & \cmark & 4.83                 & 3.47                      & 3.72          & 4.50                     & 2.71               & 3.17             \\
Restormer \cite{zamir2022restormer}     & 640$\times$480 & \cmark & 5.30                 & 3.44                      & 3.73          & 4.17                     & 2.79               & 3.13             \\ %\midrule
IFBlend (\emph{ours})                   & 640$\times$480 & \cmark & \textbf{4.02}                 & \textbf{3.38}                      & \textbf{3.52}         & \textbf{3.18}                 & \textbf{2.14}                      & \textbf{2.40}         \\ \bottomrule
\end{tabular}
}
\label{tab:quanti_istd-test}
\vspace{-2mm}
\end{table} 

\noindent \textbf{Quantitative Comparison:}
In \cref{tab:quanti_istd-test}, we conduct a comprehensive evaluation of our proposed IFBlend model against leading solutions in the shadow removal field using the benchmark ISTD and ISTD+ datasets \cite{ISTDwang2018STCGAN, Le_2019_ICCV}. Our analysis reveals that models leveraging mask priors, such as PULSr \cite{vasluianu2021shadow}, DNSR \cite{Vasluianu_2023_WSRD}, AEF \cite{fu2021auto}, and ShadowFormer \cite{guo2023shadowformer}, outperform mask-free models like DCShadowNet \cite{jin2021dc} and DeS3 \cite{jin2023des3}. This advantage stems from the localization information provided by masks, allowing mask-prior models to focus on shadow-affected segments effectively (see \cref{fig:teaser}). Nevertheless, our IFBlend model, without mask inputs or guidance, remains competitive in shadow removal compared to mask-based works. While comparing with mask-free methods employing complex adversarial learning \cite{jin2021dc} or iterative diffusion-based inference \cite{jin2023des3}, our IFBlend shows increased robustness and significantly better performance. Furthermore, IFBlend consistently outperforms transformer-based solutions \cite{liang2021swinir, wang2022uformer, zamir2022restormer}, validating our generalization capability across ALN and shadow removal.

In \cref{tab:quanti-srd-test}, we present a comparison with the SOTA methods on the SRD benchmark \cite{SRDDESHADOW}. Without leveraging mask localization information, IFBlend shows competitive performance in terms of reconstruction fidelity, narrowing the gap with ShadowFormer \cite{guo2023shadowformer} while achieving top-notch perceptual quality.

\begin{table}[t]
\centering
%\tiny
\setlength{\tabcolsep}{3pt}
%\vspace{-3mm}
\caption{Quantitative Comparison on the SRD dataset~\cite{SRDDESHADOW}.}
\vspace{-3mm}
\begin{tabular}{@{}Xlcccc@{}}
\toprule
Method  & Mask Free &\text{PSNR}$\uparrow$ & \text{SSIM}$\uparrow$ & \text{LPIPS}$\downarrow$ \\ \midrule
AEF \cite{fu2021auto}                         &\xmark & 28.40                 & 0.893                 & 0.182                    \\
DHAN \cite{cun2019ghostfree}                  &\xmark & 30.51                 & 0.949                 & 0.165                         \\
UNSR \cite{zhu2022efficient}                  &\xmark & 31.72                 & \textbf{0.955}                 & 0.149                    \\
ShadowFormer \cite{guo2023shadowformer}       &\xmark & \textbf{32.38}                 & \textbf{0.955}                 & n/a                  \\ \midrule
DeS3 \cite{jin2023des3}                       & \cmark  & 30.38               & 0.874 & 0.077 \\
DCShadowNet \cite{jin2021dc}                 & \cmark  & 30.85               & 0.913 & \textbf{0.066}  \\
IFBlend (\emph{ours})                             &\cmark & \textbf{31.95}                 & \textbf{0.950}                 & 0.072                    \\ 
\bottomrule
\end{tabular}
\label{tab:quanti-srd-test}
\vspace{-3mm}
\end{table}

\begin{figure}[t!]
    \centering
    \renewcommand{\arraystretch}{0.6}
    \setlength{\tabcolsep}{0.5pt}
    \def\widthistd{0.14}
    %\resizebox{\textwidth}{!}{
    \begin{tabularx}{\textwidth}{c|ccc|cc|c}
         \renewcommand{\arraystretch}{1.0}
         \tiny  \multirow{2}{*}{Input} & \tiny \multirow{2}{*}{Mask} & \tiny AEF & \tiny ShadowFormer   & \tiny DCShadow &  \tiny \multirow{2}{*}{\emph{Ours}} & \tiny \multirow{2}{*}{Ground Truth}\\
         & & \tiny\cite{fu2021auto} & \tiny\cite{guo2023shadowformer} &\tiny\cite{jin2021dc} &  & \\
         % \rotatebox[origin=c]{90}{Shadowformer}
         \includegraphics[width=\widthistd\linewidth]{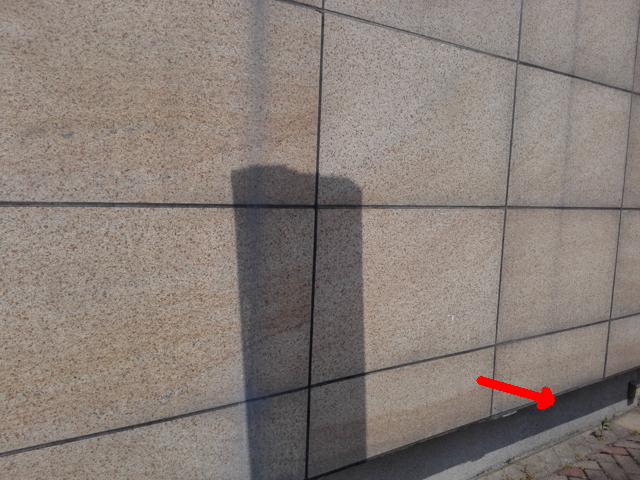}
         &  \includegraphics[width=\widthistd\linewidth]{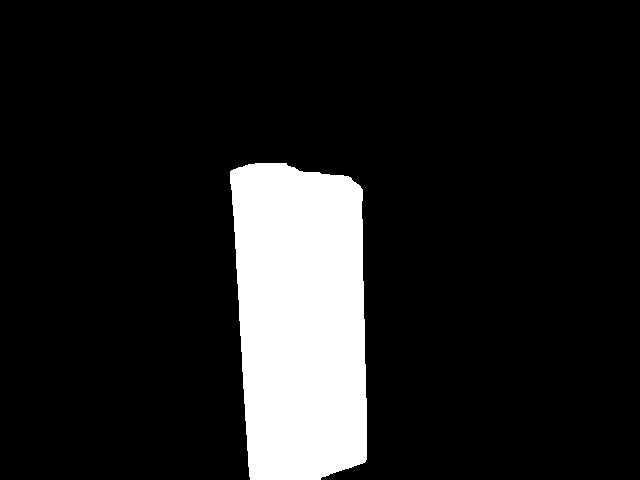}
         &  \includegraphics[width=\widthistd\linewidth]{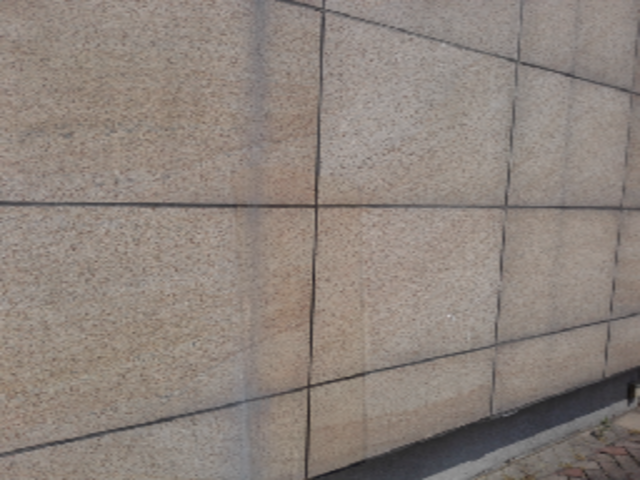}
         &  \includegraphics[width=\widthistd\linewidth]{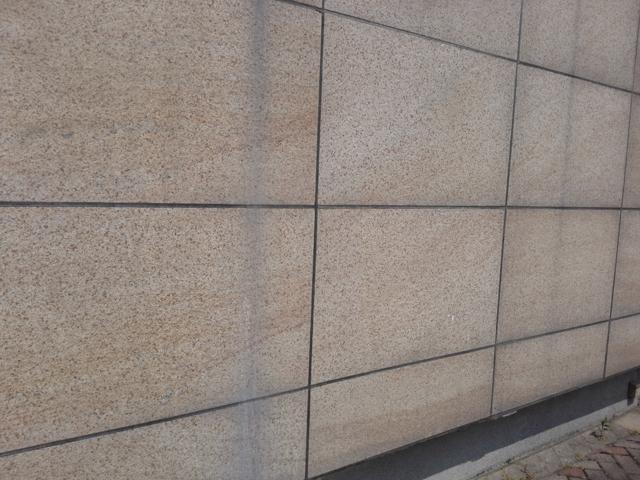}
         &  \includegraphics[width=\widthistd\linewidth]{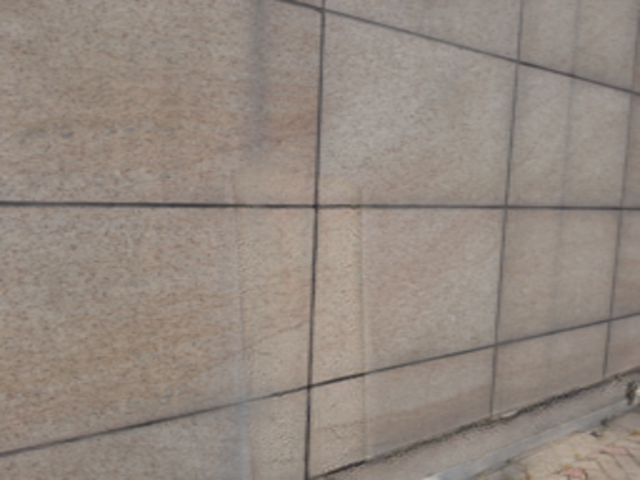}
         &  \includegraphics[width=\widthistd\linewidth]{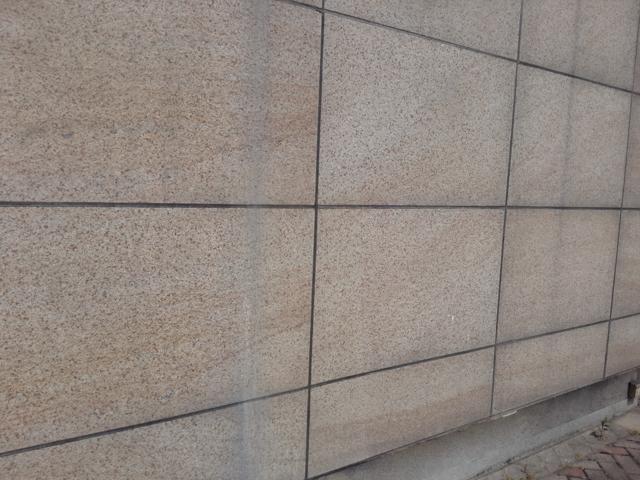}
         &  \includegraphics[width=\widthistd\linewidth]{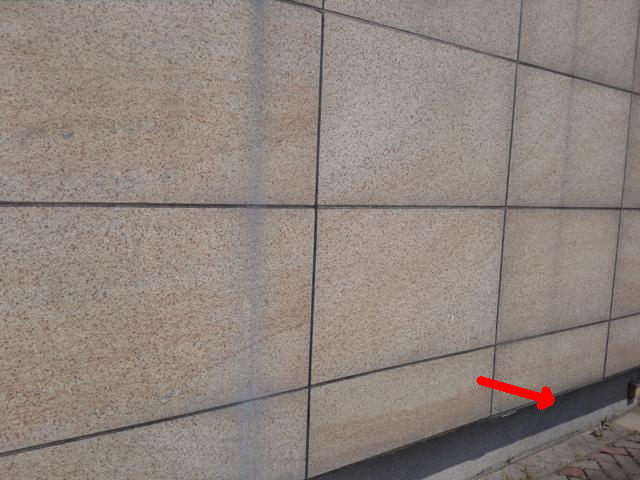} \\
         %\rotatebox[origin=c]{90}{CLN \emph{(ours)}}
         \includegraphics[width=\widthistd\linewidth]{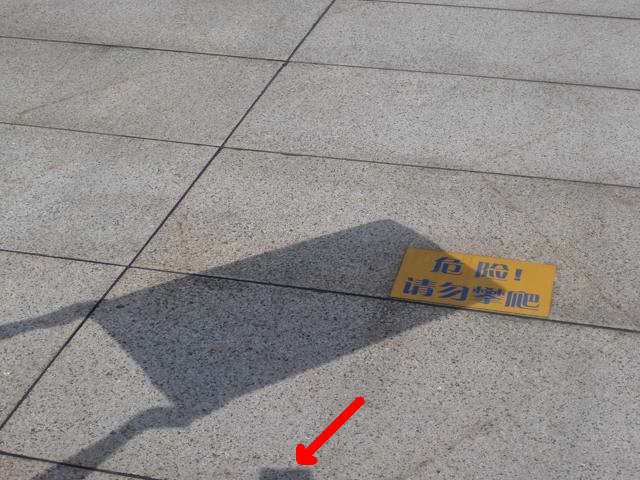}
         &  \includegraphics[width=\widthistd\linewidth]{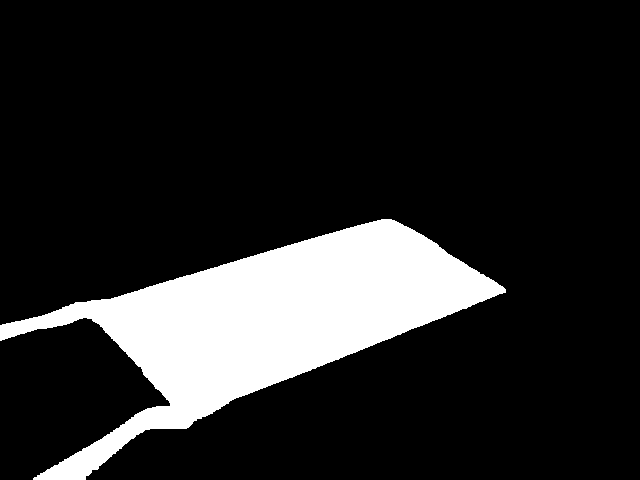}
         &  \includegraphics[width=\widthistd\linewidth]{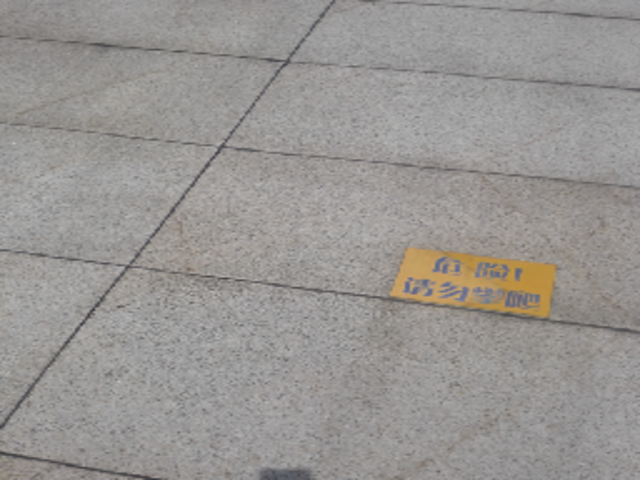}
         &  \includegraphics[width=\widthistd\linewidth]{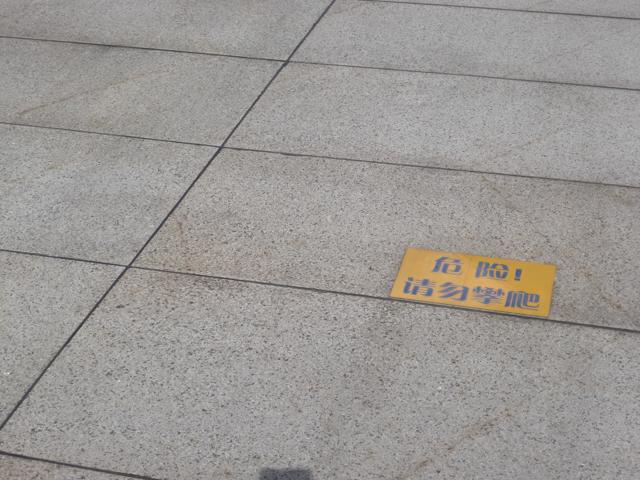}
         &  \includegraphics[width=\widthistd\linewidth]{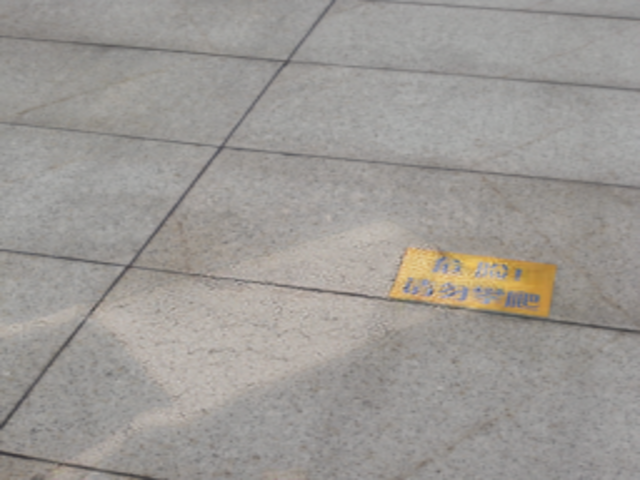}
         &  \includegraphics[width=\widthistd\linewidth]{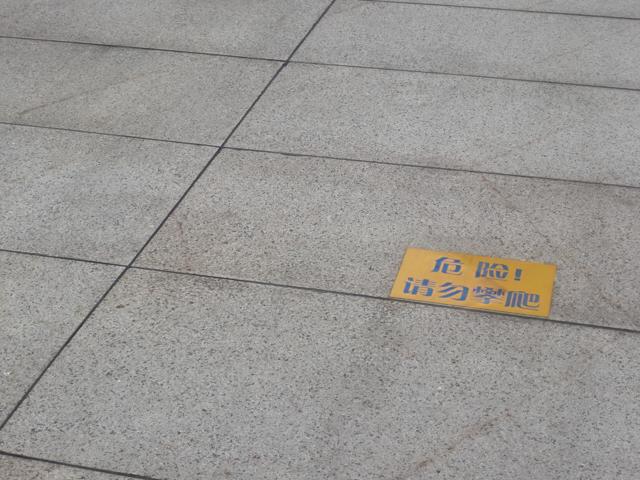}
         &  \includegraphics[width=\widthistd\linewidth]{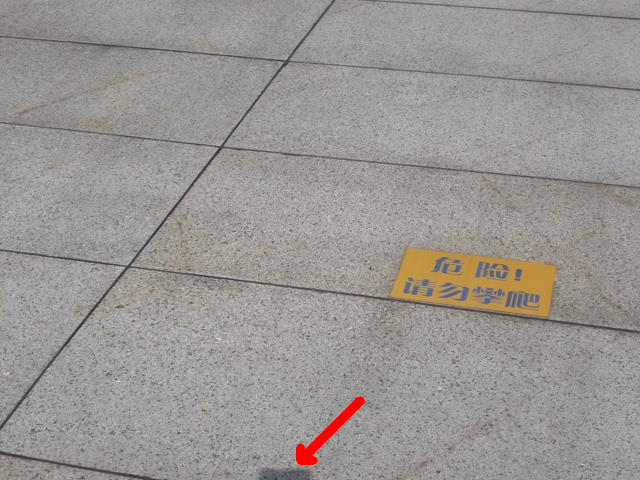} \\
         \includegraphics[width=\widthistd\linewidth]{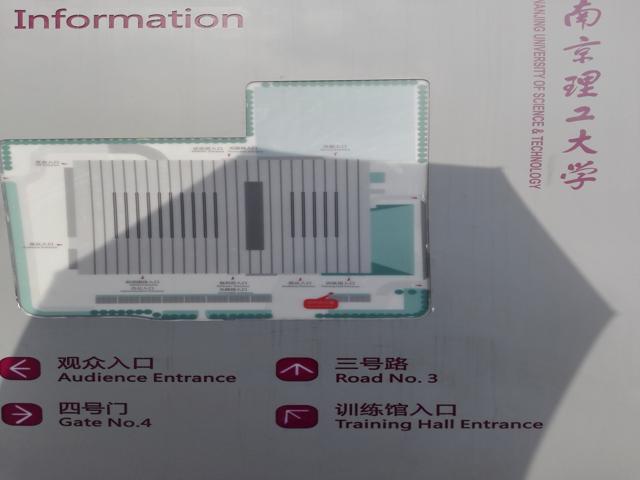}
         &  \includegraphics[width=\widthistd\linewidth]{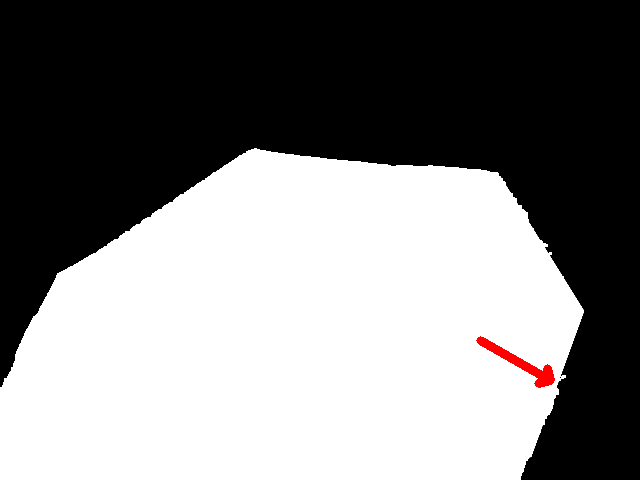}
         &  \includegraphics[width=\widthistd\linewidth]{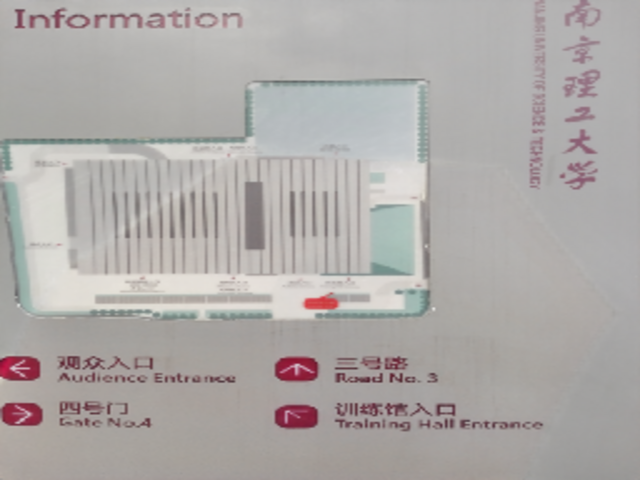}
         &  \includegraphics[width=\widthistd\linewidth]{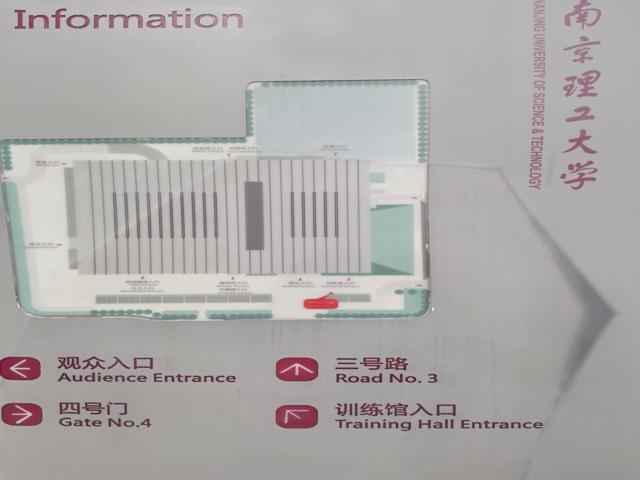}
         &  \includegraphics[width=\widthistd\linewidth]{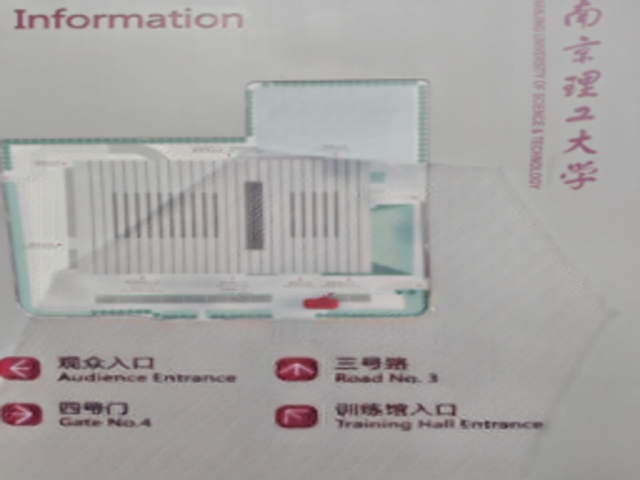}
         &  \includegraphics[width=\widthistd\linewidth]{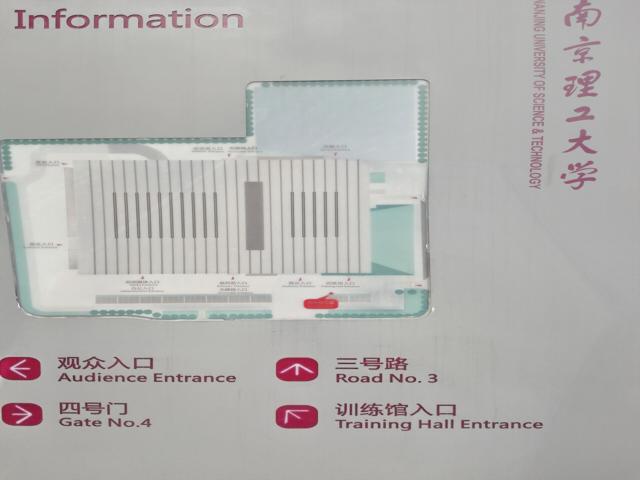}
         &  \includegraphics[width=\widthistd\linewidth]{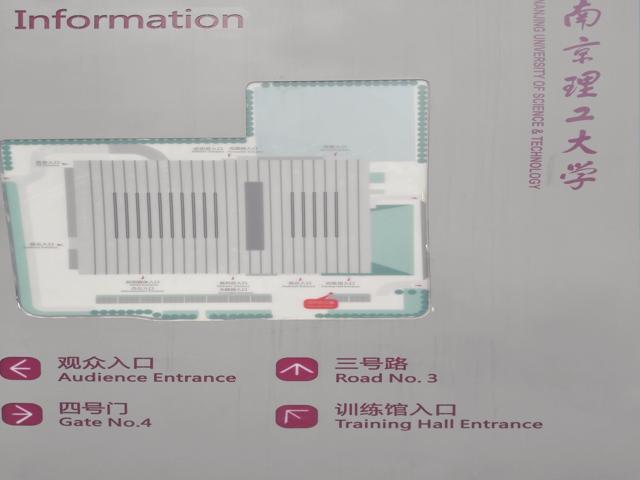} \\
         \includegraphics[width=\widthistd\linewidth]{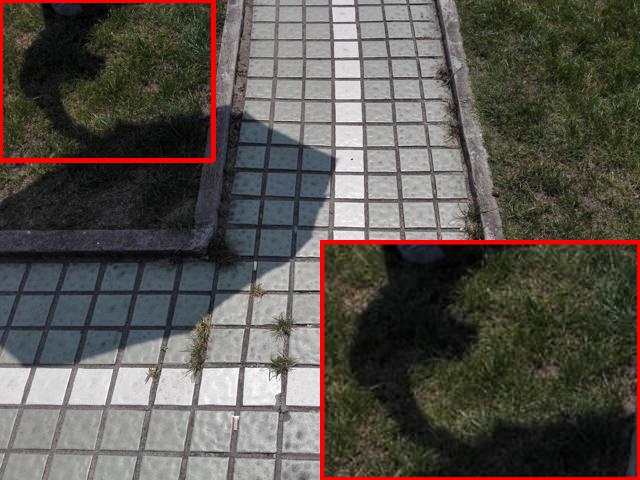}
         &  \includegraphics[width=\widthistd\linewidth]{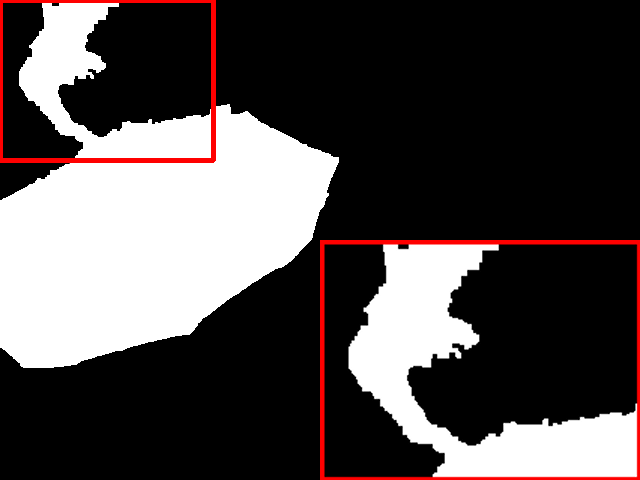}
         &  \includegraphics[width=\widthistd\linewidth]{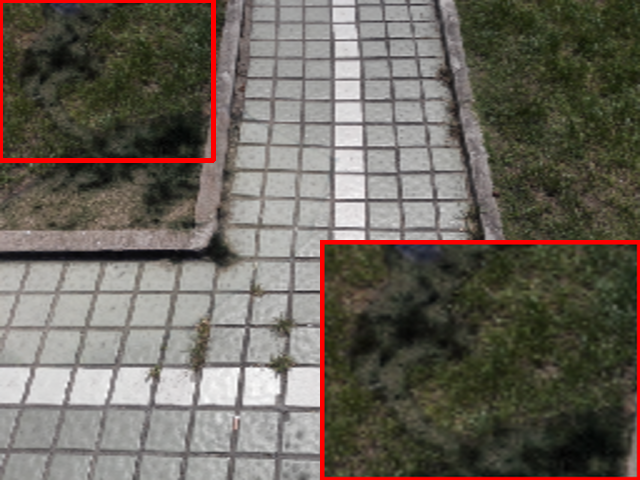}
         &  \includegraphics[width=\widthistd\linewidth]{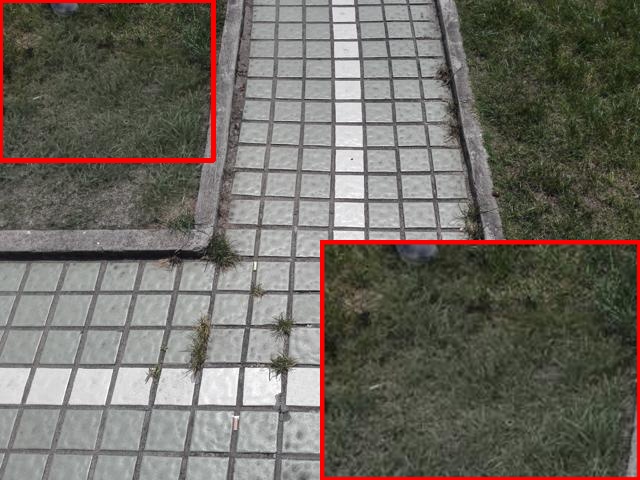}
         &  \includegraphics[width=\widthistd\linewidth]{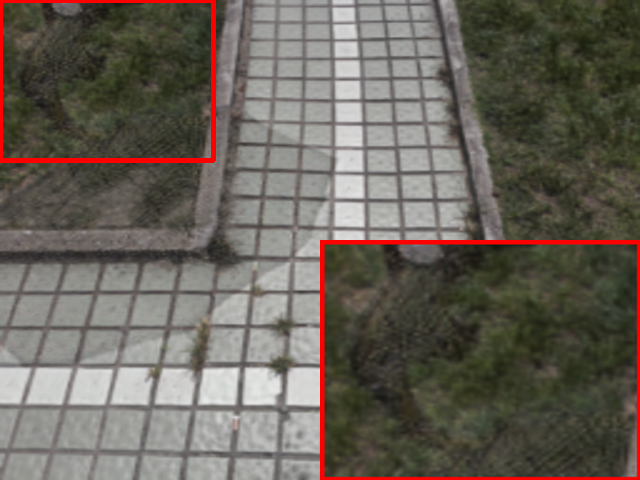}
         &  \includegraphics[width=\widthistd\linewidth]{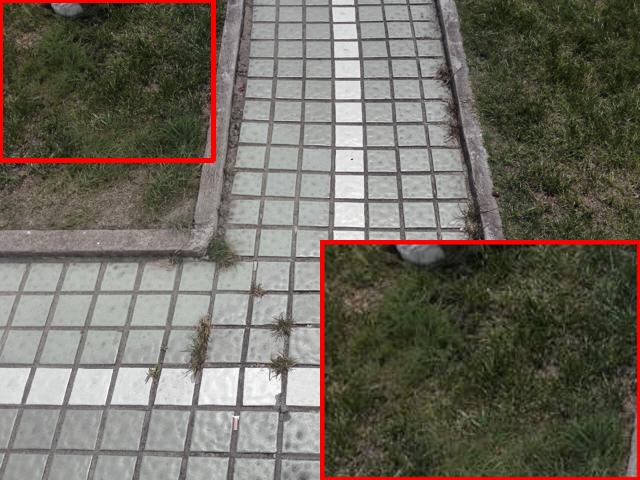}
         &  \includegraphics[width=\widthistd\linewidth]{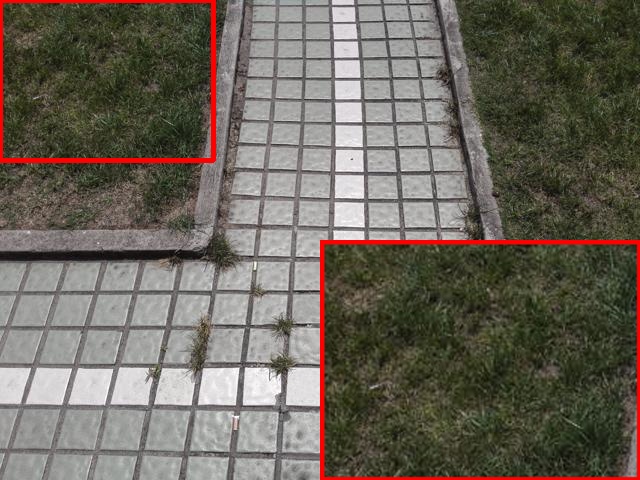} \\
         & \multicolumn{3}{c|}{\tiny w/ mask prior} &  \multicolumn{2}{c}{\tiny w/o mask prior} & \\
    \end{tabularx}
    \vspace{-3mm}
    \caption{Qualitative comparison on the ISTD+ \cite{ISTDwang2018STCGAN, Le_2019_ICCV} dataset.}
    \label{fig:istdpqual}
    \vspace{-6mm}
\end{figure}

\noindent \textbf{Remarks on the Data Quality:} While we report comparisons on shadow removal benchmarks, these datasets have inherent limitations. ISTD and ISTD+ are known to contain a high number of noisy samples (\eg incomplete masks, semantic inconsistencies), as pointed by \cite{vasluianu2021shadow}, which can be visualized in \cref{fig:istdpqual}. While a color correction method was proposed in \cite{Le_2019_ICCV} for ISTD to form ISTD+, the semantic misalignment, and the incomplete mask remain an issue. On the other hand, SRD \cite{SRDDESHADOW} suffers from significant pixel misalignment.

\noindent \textbf{Qualitative Comparison:} \cref{fig:istdpqual} presents visual comparisons on the ISTD+ dataset \cite{ISTDwang2018STCGAN, Le_2019_ICCV}. In the first three rows, models utilizing mask priors demonstrate over-reliance on localization information, particularly evident where shadow masks fail to cover all affected areas, highlighted by the \textcolor{red}{$\searrow$} arrow. For instance, the inclined shadow on the bottom right of the first-row input and the less noticeable shadow on the middle bottom of the second-row input remains unresolved. Similarly, inaccuracies in the shadow mask are reflected in the artifacts produced by ShadowFormer \cite{guo2023shadowformer}, as seen in the uncovered right extremity of its prediction, as shown in the third row. 

This observation is further backed by the last row, in which mask-based solutions tend to produce artifacts over contents accidentally appearing in the input image, which seems to be the shoe of the photographer on the top left corner (better visualized when zooming in). However, mask-free solutions like DCShadowNet \cite{jin2021dc} and our IFBlend accurately remove shadows in noisy areas, rendering a more faithful representation of the content. Compared to DCShadowNet \cite{jin2021dc}, IFBlend exhibits significantly fewer ghosting artifacts, with restored frames characterized by correct colors and improved contrast.  

\begin{figure}[t]
    \centering
    \renewcommand{\arraystretch}{0.6}
    \setlength{\tabcolsep}{0.8pt}
    \def\widthistd{0.16}
    \begin{tabularx}{\textwidth}{cccccc}
         \renewcommand{\arraystretch}{1.0}
         \scriptsize Input & \scriptsize  UNSR \cite{zhu2022efficient} & \scriptsize  DCShadow \cite{jin2021dc} &  \scriptsize  AEF \cite{fu2021auto} & \scriptsize  \emph{Ours} & \scriptsize  Ground Truth\\
         \includegraphics[width=\widthistd\linewidth]{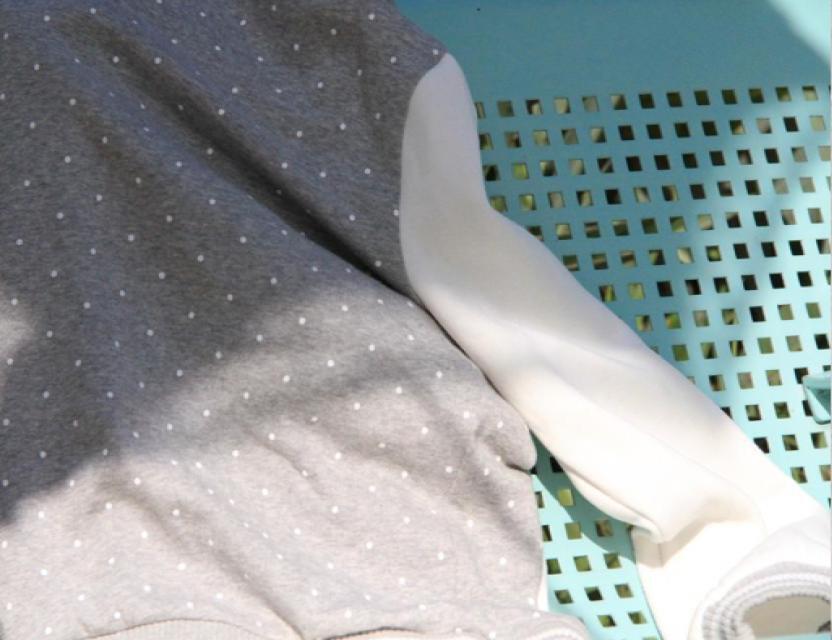}
         &  \includegraphics[width=\widthistd\linewidth]{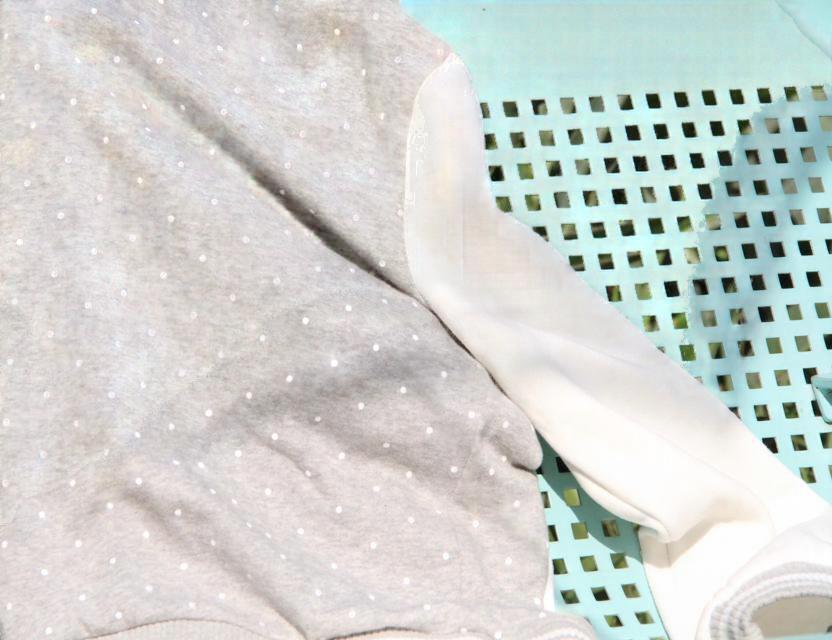}
         &  \includegraphics[width=\widthistd\linewidth]{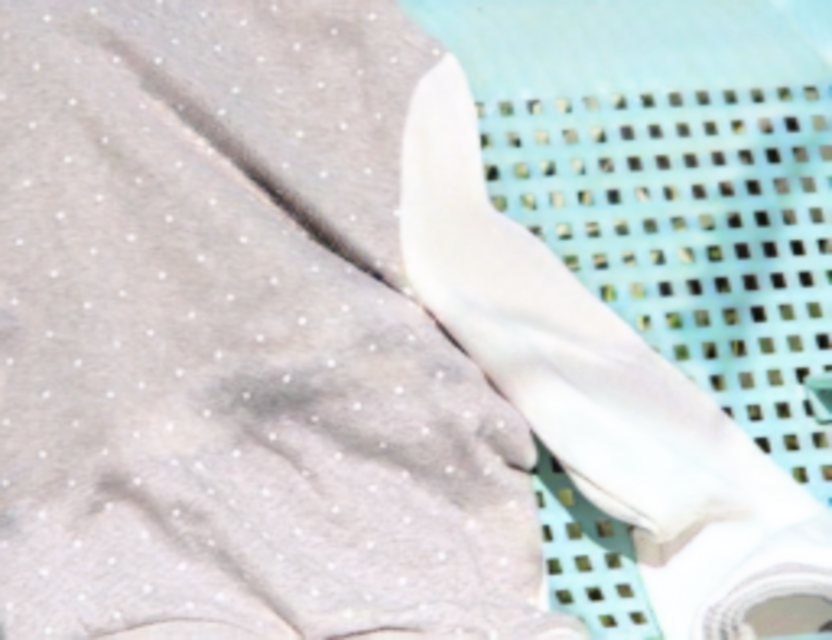}
         &  \includegraphics[width=\widthistd\linewidth]{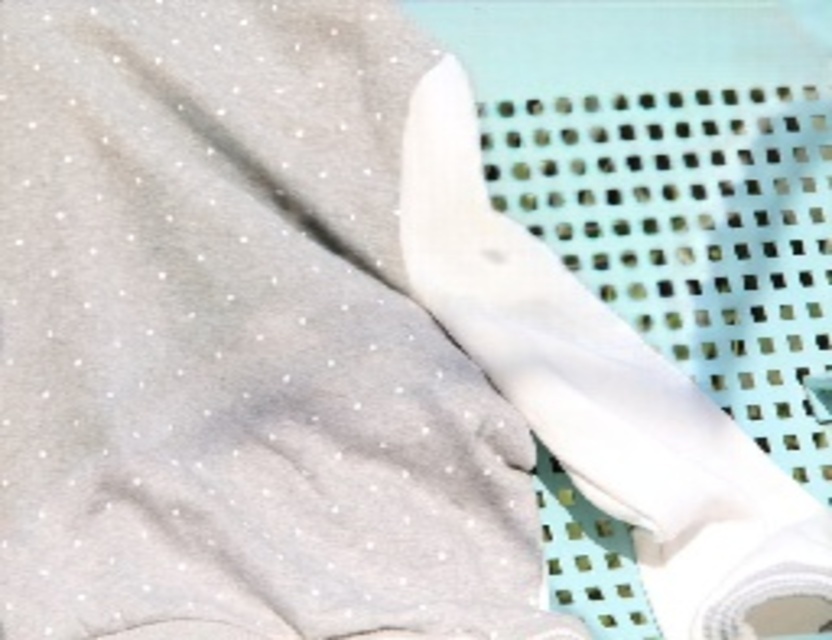}
         &  \includegraphics[width=\widthistd\linewidth]{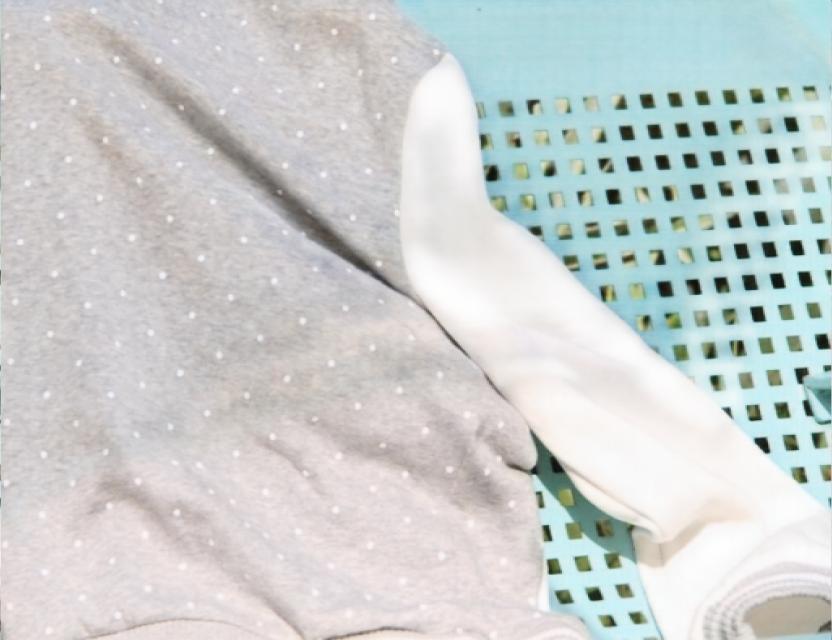}
         &  \includegraphics[width=\widthistd\linewidth]{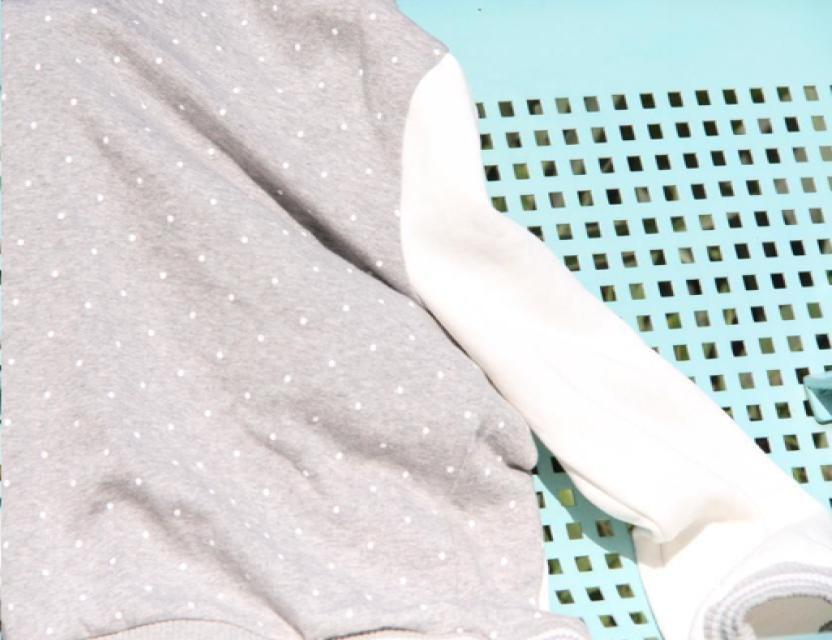} \\
         \includegraphics[width=\widthistd\linewidth]{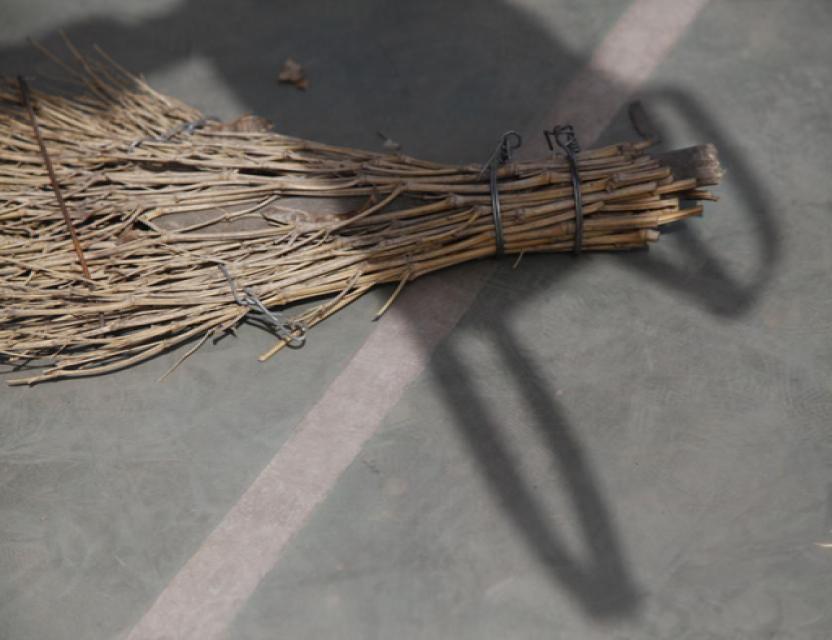}
         &  \includegraphics[width=\widthistd\linewidth]{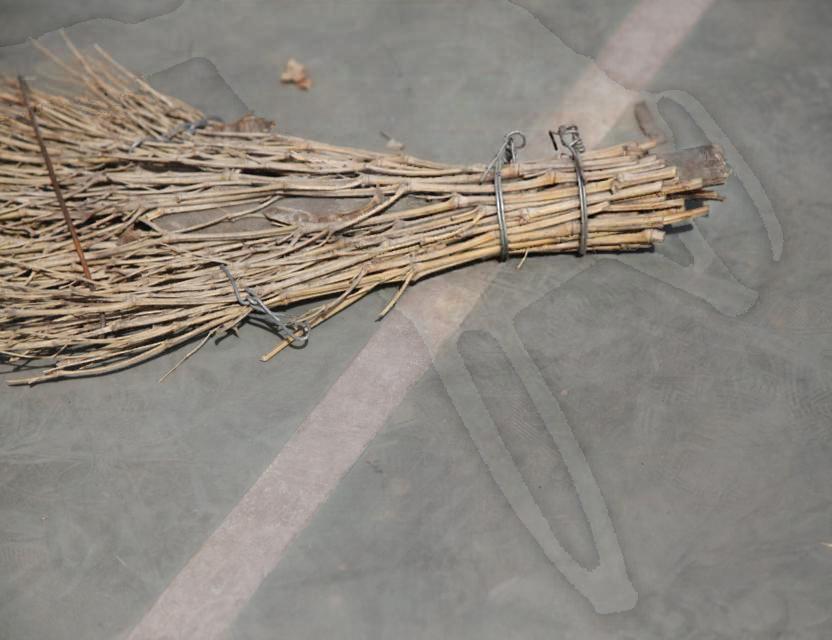}
         &  \includegraphics[width=\widthistd\linewidth]{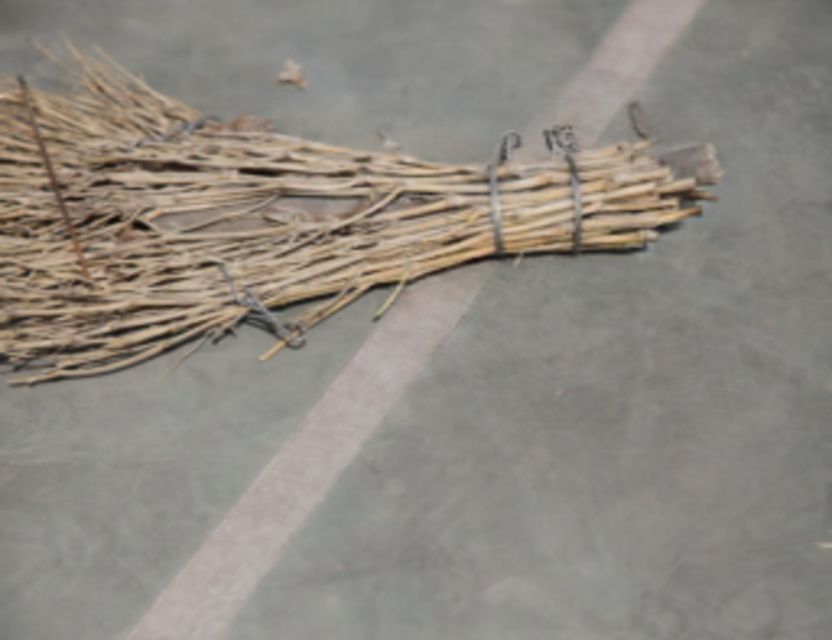}
         &  \includegraphics[width=\widthistd\linewidth]{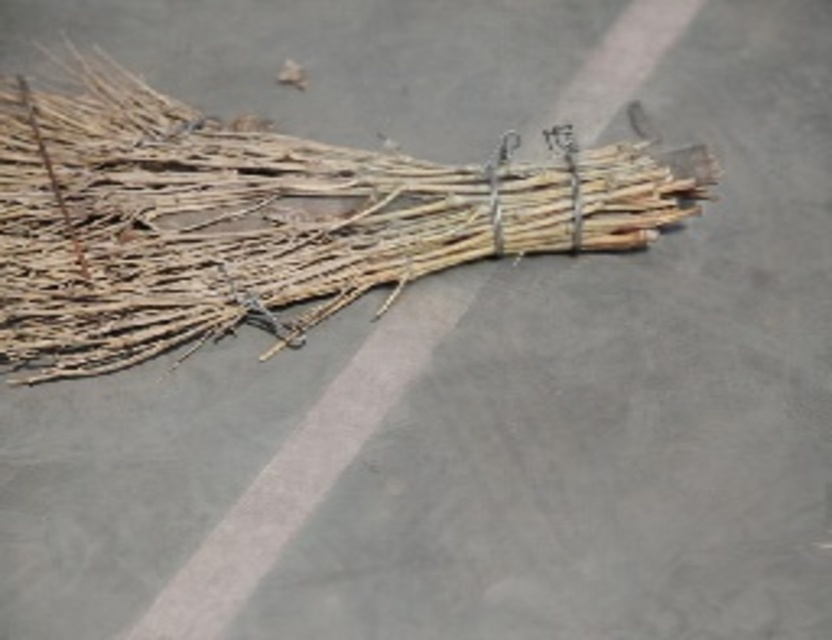}
         &  \includegraphics[width=\widthistd\linewidth]{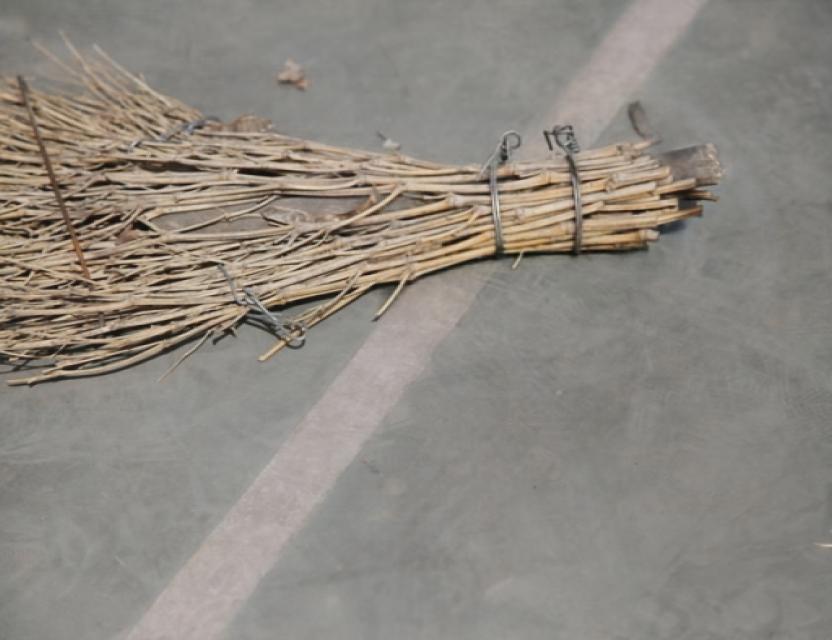}
         &  \includegraphics[width=\widthistd\linewidth]{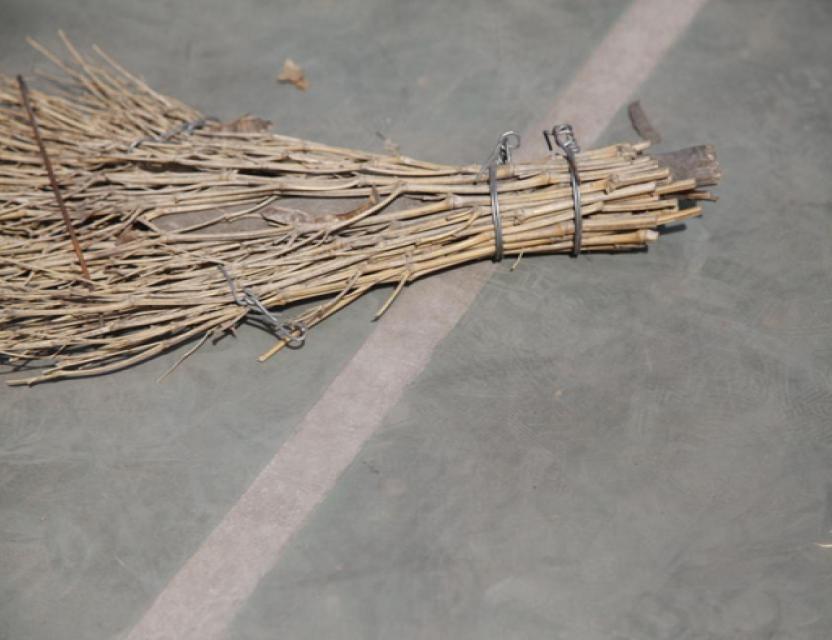} \\
         \includegraphics[width=\widthistd\linewidth]{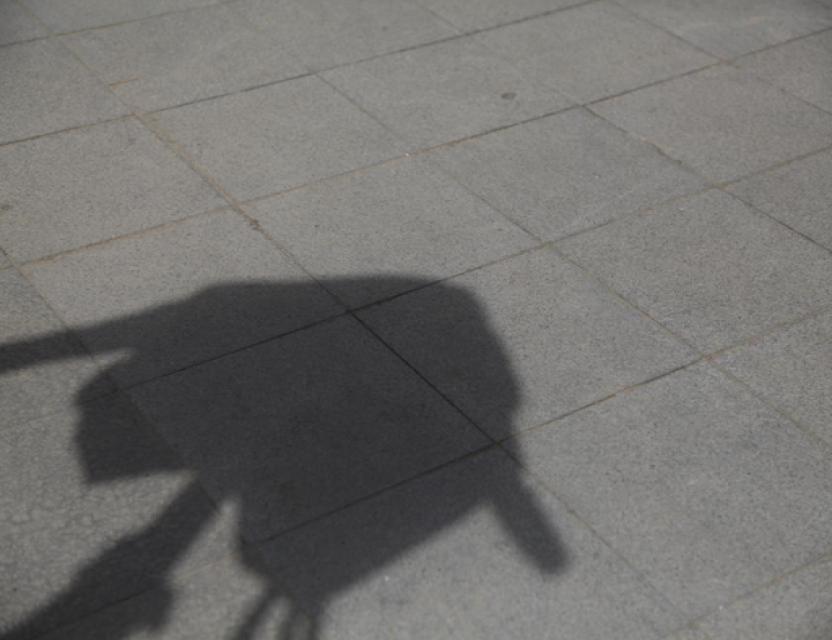}
         &  \includegraphics[width=\widthistd\linewidth]{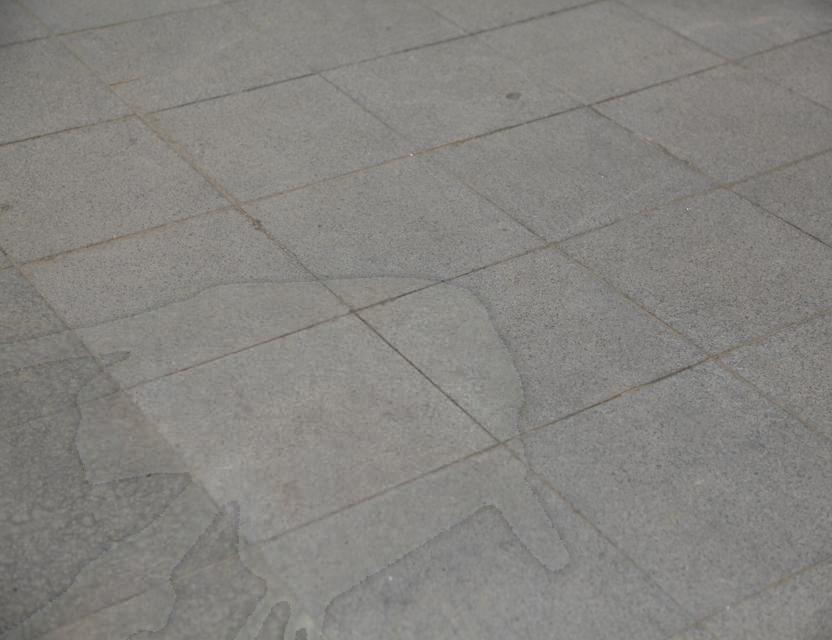}
         &  \includegraphics[width=\widthistd\linewidth]{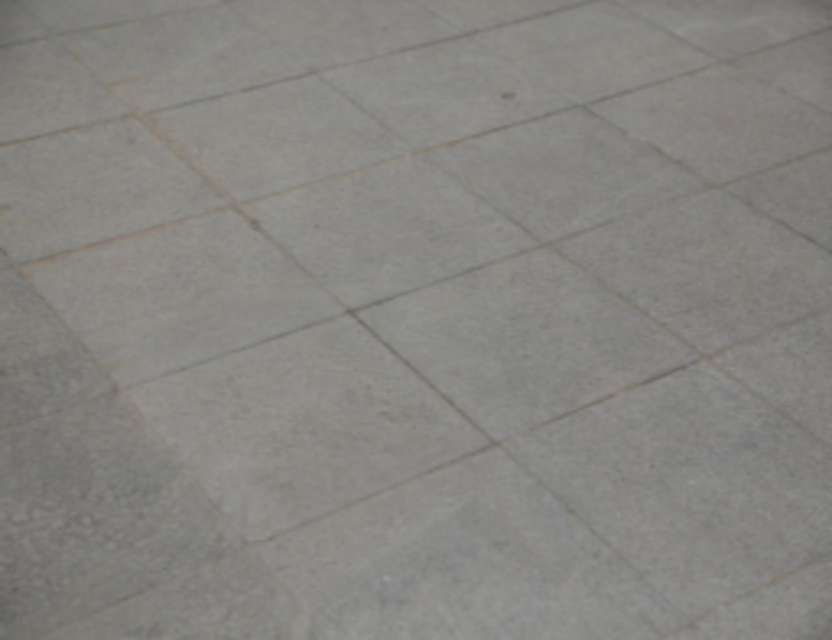}
         &  \includegraphics[width=\widthistd\linewidth]{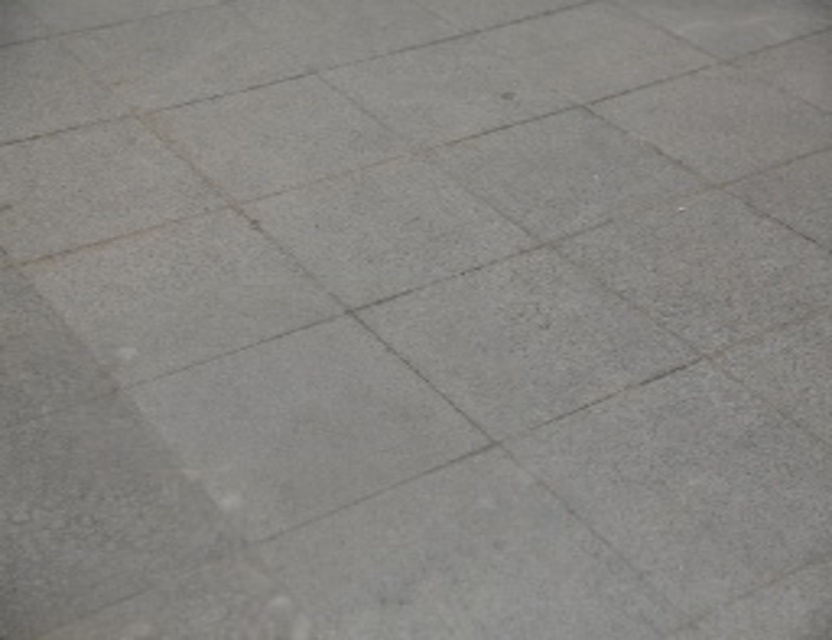}
         &  \includegraphics[width=\widthistd\linewidth]{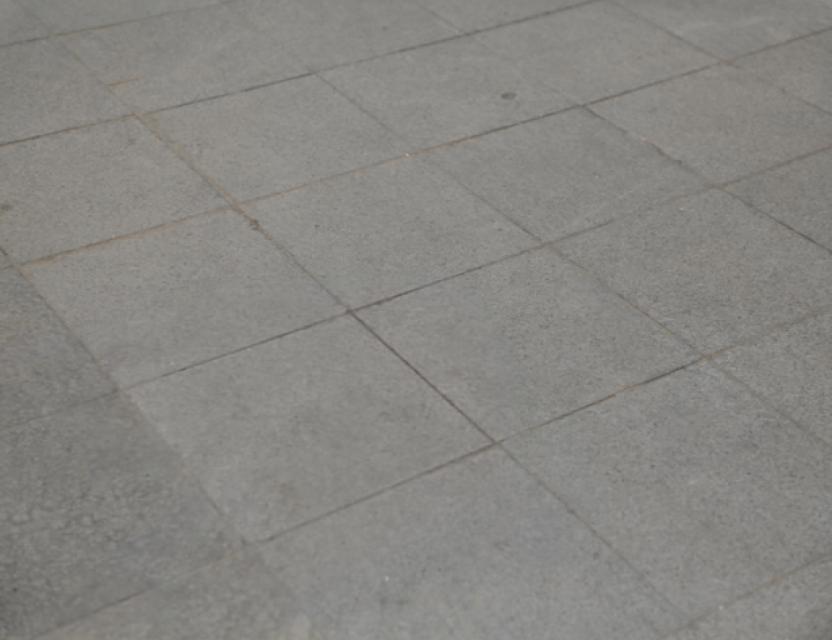}
         &  \includegraphics[width=\widthistd\linewidth]{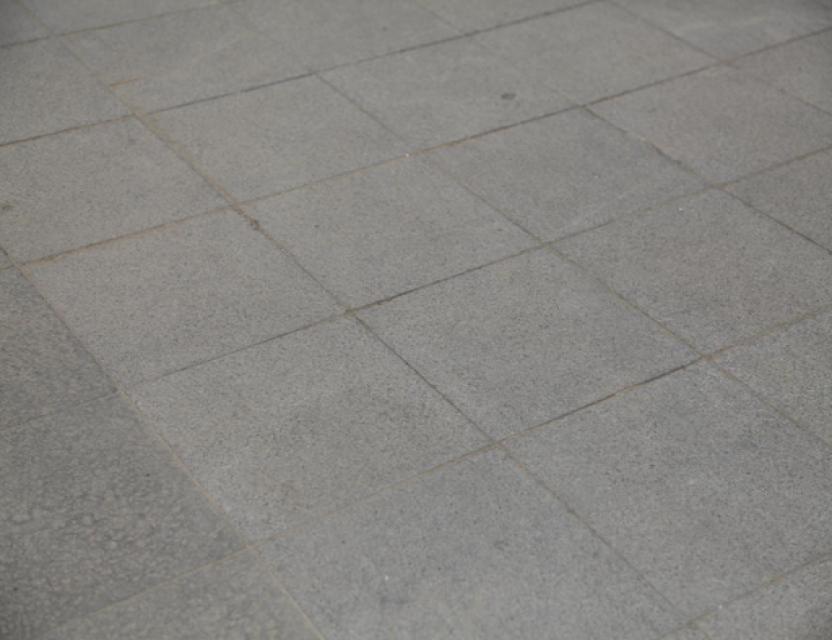} \\
         \includegraphics[width=\widthistd\linewidth]{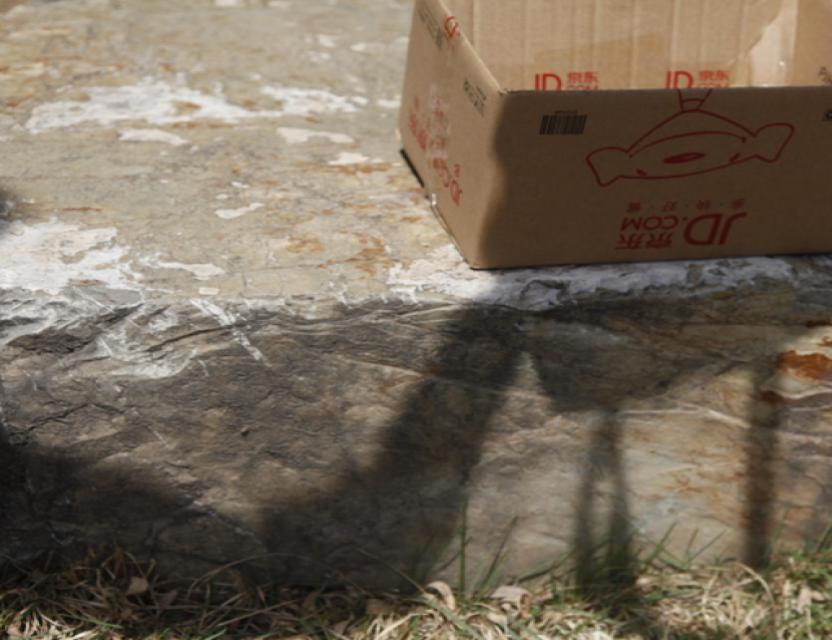}
         &  \includegraphics[width=\widthistd\linewidth]{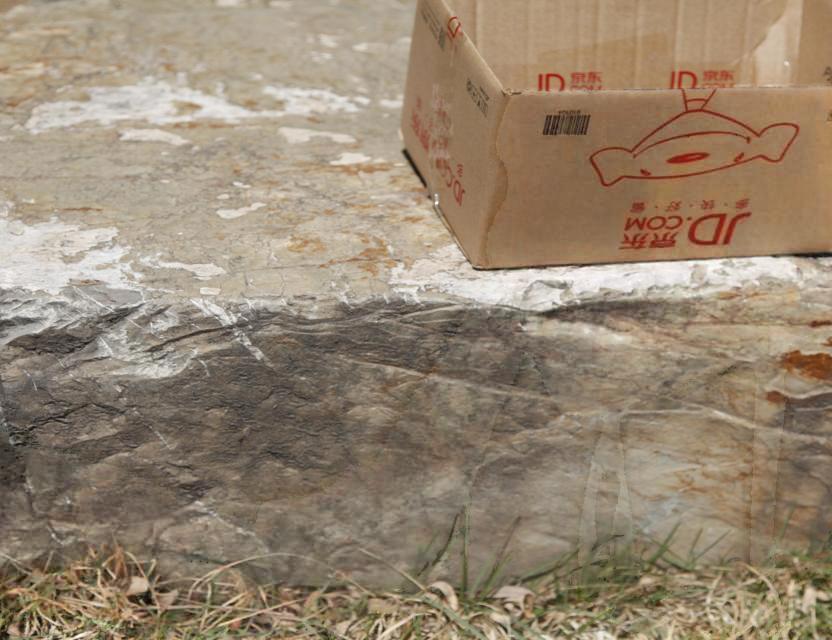}
         &  \includegraphics[width=\widthistd\linewidth]{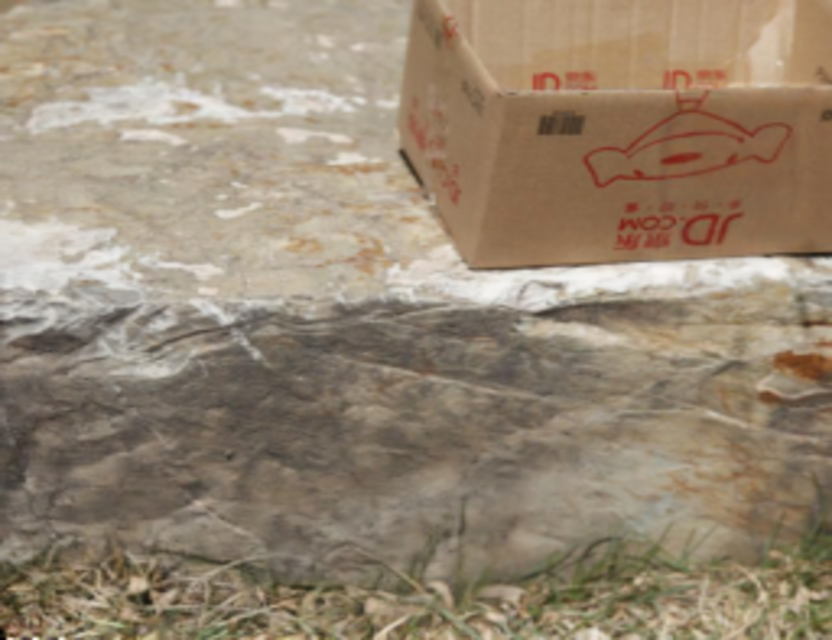}
         &  \includegraphics[width=\widthistd\linewidth]{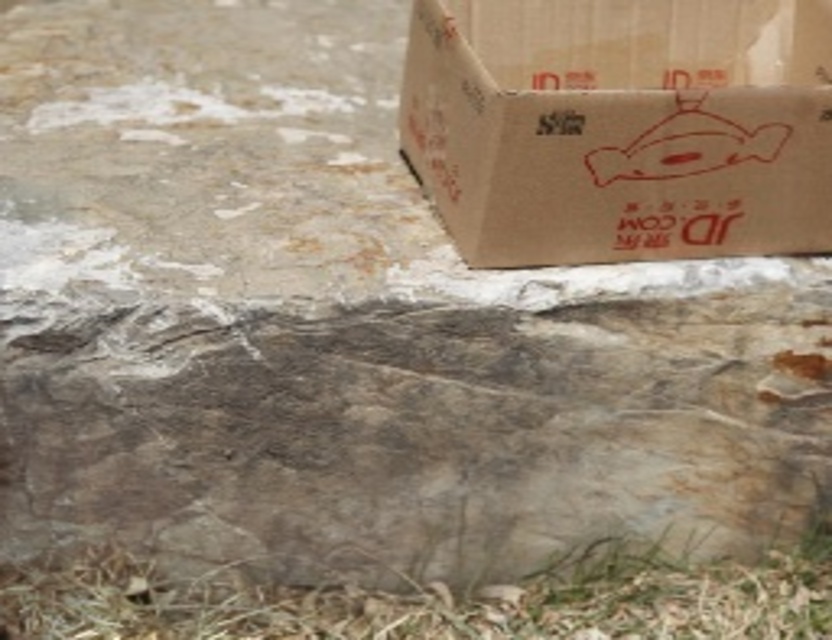}
         &  \includegraphics[width=\widthistd\linewidth]{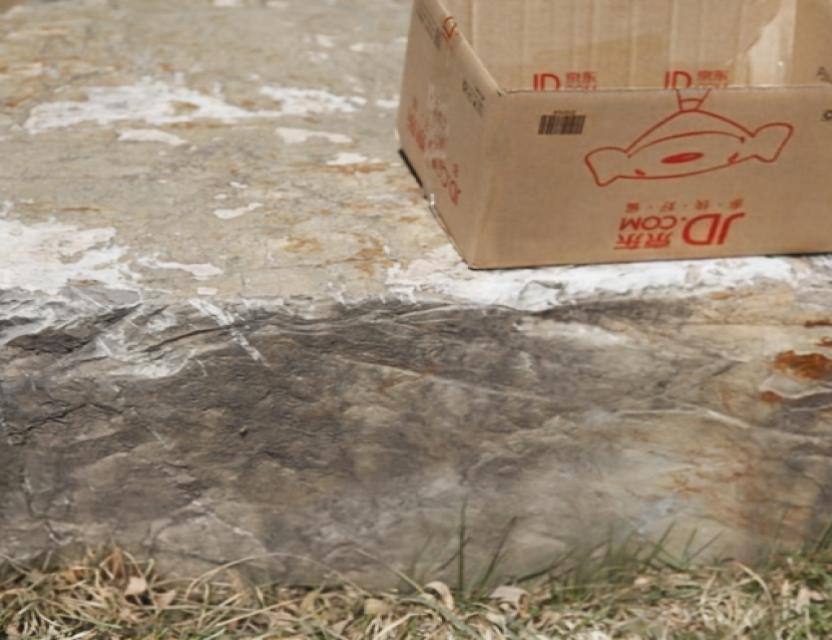}
         &  \includegraphics[width=\widthistd\linewidth]{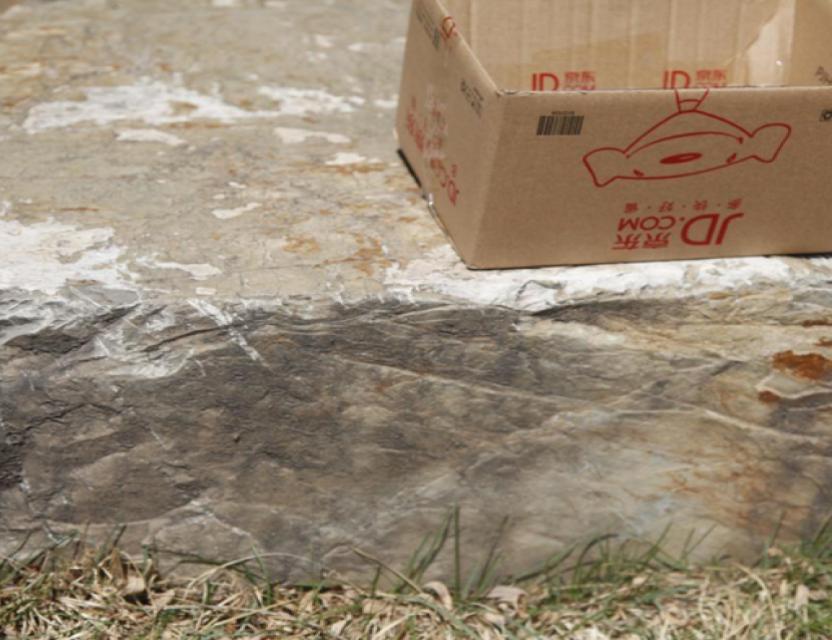} \\
    \end{tabularx}
        \vspace{-3mm}
    \caption{Visual comparison on the SRD~\cite{SRDDESHADOW} dataset.}
    \label{fig:srdqual}
\end{figure}

In \cref{fig:srdqual}, we present a qualitative comparison on the SRD dataset \cite{SRDDESHADOW}. Our model demonstrates superior performance, particularly in local regions where other SOTA methods falter, producing outputs closer to the ground truth.

\begin{table}[t]
\centering
%\tiny
\setlength{\tabcolsep}{3pt}
\caption{Ablation study conducted on Ambient6K dataset.}
\vspace{-3mm}
%\resizebox{\linewidth}{!}
{%
\begin{tabular}{lcccc}
\toprule
Variants                                & MACs (G.) & \text{PSNR}$\uparrow$ & \text{SSIM}$\uparrow$ & \text{LPIPS}$\downarrow$ \\ \midrule

\emph{no DWT feats. }           & 25.52     & 19.624                & 0.753                 & 0.147                         \\
\emph{no RGB splitting}          & 26.16    &  20.430               &  0.799                & 0.130                         \\ 
\emph{no SSIM loss}            & 26.01     & \textbf{20.752}        & 0.792                 & 0.131                \\ \midrule
IFBlend (\emph{ours})                     & 26.01     & 20.714                & \textbf{0.810}                 & \textbf{0.122}       \\ 
\bottomrule
\end{tabular}%
}
\label{tab:ablation-alnd-test}
\vspace{-3mm}
\end{table}

\subsection{Ablation Studies}  
We conducted ablation studies on key components of our model using the Ambient6K dataset. The results in \cref{tab:ablation-alnd-test} validate the inclusion of DWT Haar features, RGB information splitting, and the SSIM loss term, proving our effectiveness. Further studies are available in the supplementary material.

\section{Conclusion}

We provide a comprehensive study on Ambient Lighting Normalization (ALN), investigating shadow interactions. To address the lack of an appropriate and high-quality large-scale dataset, we present the Ambient6K, which contains samples obtained from complex and realistic lighting conditions. A number of existing mainstream models are selected and evaluated on Ambient6K, serving for benchmarking proposals. In addition, we propose a strong baseline for ALN named IFBlend, which is equipped with novel image-frequency shrinkage fusion designs, combining features of different bands. IFBlend achieves SOTA results on Ambient6K and competitive performance on conventional shadow removal benchmarks, without requiring extra mask priors. In summary, our dataset and method offer a valuable platform for future research in this domain.

% ---- Bibliography ----
%
% BibTeX users should specify bibliography style 'splncs04'.
% References will then be sorted and formatted in the correct style.
%
\bibliographystyle{splncs04}
\bibliography{references}

\begin{thebibliography}{10}
\providecommand{\url}[1]{\texttt{#1}}
\providecommand{\urlprefix}{URL }
\providecommand{\doi}[1]{https://doi.org/#1}

\bibitem{ancuti2018haze}
Ancuti, C.O., Ancuti, C., Timofte, R., De~Vleeschouwer, C.: O-haze: a dehazing benchmark with real hazy and haze-free outdoor images. In: Proceedings of the IEEE conference on computer vision and pattern recognition workshops. pp. 754--762 (2018)

\bibitem{ancuti2020ntire}
Ancuti, C.O., Ancuti, C., Vasluianu, F.A., Timofte, R.: Ntire 2020 challenge on nonhomogeneous dehazing. In: Proceedings of the IEEE/CVF Conference on Computer Vision and Pattern Recognition Workshops. pp. 490--491 (2020)

\bibitem{buades2011non}
Buades, A., Coll, B., Morel, J.M.: Non-local means denoising. Image Processing On Line  \textbf{1},  208--212 (2011)

\bibitem{Cai_2019_ICCV}
Cai, J., Zeng, H., Yong, H., Cao, Z., Zhang, L.: Toward real-world single image super-resolution: A new benchmark and a new model. In: Proceedings of the IEEE/CVF International Conference on Computer Vision (ICCV) (October 2019)

\bibitem{chen2022simple}
Chen, L., Chu, X., Zhang, X., Sun, J.: Simple baselines for image restoration. arXiv preprint arXiv:2204.04676  (2022)

\bibitem{Chen_2021_CVPR}
Chen, L., Lu, X., Zhang, J., Chu, X., Chen, C.: Hinet: Half instance normalization network for image restoration. In: Proceedings of the IEEE/CVF Conference on Computer Vision and Pattern Recognition (CVPR) Workshops. pp. 182--192 (June 2021)

\bibitem{chen2021hdrunet}
Chen, X., Liu, Y., Zhang, Z., Qiao, Y., Dong, C.: Hdrunet: Single image hdr reconstruction with denoising and dequantization. In: Proceedings of the IEEE/CVF Conference on Computer Vision and Pattern Recognition (CVPR) Workshops. pp. 354--363 (June 2021)

\bibitem{cui2022selective}
Cui, Y., Tao, Y., Bing, Z., Ren, W., Gao, X., Cao, X., Huang, K., Knoll, A.: Selective frequency network for image restoration. In: The Eleventh International Conference on Learning Representations (2022)

\bibitem{cun2019ghostfree}
Cun, X., Pun, C.M., Shi, C.: Towards ghost-free shadow removal via dual hierarchical aggregation network and shadow matting gan (2019)

\bibitem{delbracio2023inversion}
Delbracio, M., Milanfar, P.: Inversion by direct iteration: An alternative to denoising diffusion for image restoration. arXiv preprint arXiv:2303.11435  (2023)

\bibitem{Finlayson_entropyminimization}
Finlayson, G.D., Drew, M.S., Lu, C.: Entropy minimization for shadow removal. International Journal of Computer Vision  \textbf{85}(1),  35--57 (2009)

\bibitem{finlayson2002removing}
Finlayson, G.D., Hordley, S.D., Drew, M.S.: Removing shadows from images. In: European conference on computer vision. pp. 823--836. Springer (2002)

\bibitem{fu2021auto}
Fu, L., Zhou, C., Guo, Q., Juefei-Xu, F., Yu, H., Feng, W., Liu, Y., Wang, S.: Auto-exposure fusion for single-image shadow removal. In: Proceedings of the IEEE/CVF conference on computer vision and pattern recognition. pp. 10571--10580 (2021)

\bibitem{fu2021dw}
Fu, M., Liu, H., Yu, Y., Chen, J., Wang, K.: Dw-gan: A discrete wavelet transform gan for nonhomogeneous dehazing. In: Proceedings of the IEEE/CVF Conference on Computer Vision and Pattern Recognition. pp. 203--212 (2021)

\bibitem{garcia2018survey}
Garcia-Garcia, A., Orts-Escolano, S., Oprea, S., Villena-Martinez, V., Martinez-Gonzalez, P., Garcia-Rodriguez, J.: A survey on deep learning techniques for image and video semantic segmentation. Applied Soft Computing  \textbf{70},  41--65 (2018)

\bibitem{gong2014interactive}
Gong, H., Cosker, D.: Interactive shadow removal and ground truth for variable scene categories. In: Proceedings of the British Machine Vision Conference (2014)

\bibitem{guo2023shadowformer}
Guo, L., Huang, S., Liu, D., Cheng, H., Wen, B.: Shadowformer: Global context helps image shadow removal. arXiv preprint arXiv:2302.01650  (2023)

\bibitem{guo2023shadowdiffusion}
Guo, L., Wang, C., Yang, W., Huang, S., Wang, Y., Pfister, H., Wen, B.: Shadowdiffusion: When degradation prior meets diffusion model for shadow removal. In: Proceedings of the IEEE/CVF Conference on Computer Vision and Pattern Recognition. pp. 14049--14058 (2023)

\bibitem{guo2013paired}
Guo, R., Dai, Q., Hoiem, D.: Paired regions for shadow detection and removal. IEEE Transactions on Pattern Analysis and Machine Intelligence  \textbf{35}(12),  2956--2967 (11 2013). \doi{10.1109/TPAMI.2012.214}

\bibitem{he2017mask}
He, K., Gkioxari, G., Doll{\'a}r, P., Girshick, R.: Mask r-cnn. In: Proceedings of the IEEE international conference on computer vision. pp. 2961--2969 (2017)

\bibitem{heusch2005lighting}
Heusch, G., Cardinaux, F., Marcel, S.: Lighting normalization algorithms for face verification. Tech. rep., IDIAP (2005)

\bibitem{ho2020denoising}
Ho, J., Jain, A., Abbeel, P.: Denoising diffusion probabilistic models. Advances in neural information processing systems  \textbf{33},  6840--6851 (2020)

\bibitem{hu2019direction}
Hu, X., Fu, C.W., Zhu, L., Qin, J., Heng, P.A.: Direction-aware spatial context features for shadow detection and removal. IEEE Transactions on Pattern Analysis and Machine Intelligence  (2019), to appear

\bibitem{USRhu2019mask}
Hu, X., Jiang, Y., Fu, C.W., Heng, P.A.: {Mask-ShadowGAN}: Learning to remove shadows from unpaired data. In: ICCV (2019)

\bibitem{jin2021dc}
Jin, Y., Sharma, A., Tan, R.T.: Dc-shadownet: Single-image hard and soft shadow removal using unsupervised domain-classifier guided network. In: Proceedings of the IEEE/CVF International Conference on Computer Vision. pp. 5027--5036 (2021)

\bibitem{jin2023des3}
Jin, Y., Yang, W., Ye, W., Yuan, Y., Tan, R.T.: Des3: Adaptive attention-driven self and soft shadow removal using vit similarity (2023)

\bibitem{kingma2014adam}
Kingma, D.P., Ba, J.: Adam: A method for stochastic optimization. arXiv preprint arXiv:1412.6980  (2014)

\bibitem{kristan2015visual}
Kristan, M., Matas, J., Leonardis, A., Felsberg, M., Cehovin, L., Fernandez, G., Vojir, T., Hager, G., Nebehay, G., Pflugfelder, R.: The visual object tracking vot2015 challenge results. In: Proceedings of the IEEE international conference on computer vision workshops. pp. 1--23 (2015)

\bibitem{lahiri2021lipsync3d}
Lahiri, A., Kwatra, V., Frueh, C., Lewis, J., Bregler, C.: Lipsync3d: Data-efficient learning of personalized 3d talking faces from video using pose and lighting normalization. In: Proceedings of the IEEE/CVF conference on computer vision and pattern recognition. pp. 2755--2764 (2021)

\bibitem{Le_2019_ICCV}
Le, H., Samaras, D.: Shadow removal via shadow image decomposition. In: The IEEE International Conference on Computer Vision (ICCV) (October 2019)

\bibitem{Le_2020_ECCV}
Le, H., Samaras, D.: From shadow segmentation to shadow removal. In: The IEEE European Conference on Computer Vision (ECCV) (August 2020)

\bibitem{lee2005practical}
Lee, K.C., Moghaddam, B.: A practical face relighting method for directional lighting normalization. In: International Workshop on Analysis and Modeling of Faces and Gestures. pp. 155--169. Springer (2005)

\bibitem{li2022srdiff}
Li, H., Yang, Y., Chang, M., Chen, S., Feng, H., Xu, Z., Li, Q., Chen, Y.: Srdiff: Single image super-resolution with diffusion probabilistic models. Neurocomputing  \textbf{479},  47--59 (2022)

\bibitem{Li_2023_ICCV}
Li, Z., Chen, X., Pun, C.M., Cun, X.: High-resolution document shadow removal via a large-scale real-world dataset and a frequency-aware shadow erasing net. In: Proceedings of the IEEE/CVF International Conference on Computer Vision (ICCV). pp. 12449--12458 (October 2023)

\bibitem{liang2022vrt}
Liang, J., Cao, J., Fan, Y., Zhang, K., Ranjan, R., Li, Y., Timofte, R., Van~Gool, L.: Vrt: A video restoration transformer. arXiv preprint arXiv:2201.12288  (2022)

\bibitem{liang2021swinir}
Liang, J., Cao, J., Sun, G., Zhang, K., Van~Gool, L., Timofte, R.: Swinir: Image restoration using swin transformer. In: Proceedings of the IEEE/CVF international conference on computer vision. pp. 1833--1844 (2021)

\bibitem{liu2021Swin}
Liu, Z., Lin, Y., Cao, Y., Hu, H., Wei, Y., Zhang, Z., Lin, S., Guo, B.: Swin transformer: Hierarchical vision transformer using shifted windows. In: Proceedings of the IEEE/CVF International Conference on Computer Vision (ICCV) (2021)

\bibitem{liu2022convnet}
Liu, Z., Mao, H., Wu, C.Y., Feichtenhofer, C., Darrell, T., Xie, S.: A convnet for the 2020s. Proceedings of the IEEE/CVF Conference on Computer Vision and Pattern Recognition (CVPR)  (2022)

\bibitem{lugmayr2020srflow}
Lugmayr, A., Danelljan, M., Van~Gool, L., Timofte, R.: Srflow: Learning the super-resolution space with normalizing flow. In: Computer Vision--ECCV 2020: 16th European Conference, Glasgow, UK, August 23--28, 2020, Proceedings, Part V 16. pp. 715--732. Springer (2020)

\bibitem{luo2023image}
Luo, Z., Gustafsson, F.K., Zhao, Z., Sj{\"o}lund, J., Sch{\"o}n, T.B.: Image restoration with mean-reverting stochastic differential equations. arXiv preprint arXiv:2301.11699  (2023)

\bibitem{luo2023refusion}
Luo, Z., Gustafsson, F.K., Zhao, Z., Sj{\"o}lund, J., Sch{\"o}n, T.B.: Refusion: Enabling large-size realistic image restoration with latent-space diffusion models. In: Proceedings of the IEEE/CVF Conference on Computer Vision and Pattern Recognition Workshops. pp. 1680--1691 (2023)

\bibitem{mehri2021mprnet}
Mehri, A., Ardakani, P.B., Sappa, A.D.: Mprnet: Multi-path residual network for lightweight image super resolution. In: Proceedings of the IEEE/CVF Winter Conference on Applications of Computer Vision. pp. 2704--2713 (2021)

\bibitem{SRDDESHADOW}
{Qu}, L., {Tian}, J., {He}, S., {Tang}, Y., {Lau}, R.W.H.: Deshadownet: A multi-context embedding deep network for shadow removal. In: 2017 IEEE Conference on Computer Vision and Pattern Recognition (CVPR). pp. 2308--2316 (July 2017). \doi{10.1109/CVPR.2017.248}

\bibitem{rim2022realistic}
Rim, J., Kim, G., Kim, J., Lee, J., Lee, S., Cho, S.: Realistic blur synthesis for learning image deblurring. In: European conference on computer vision. pp. 487--503. Springer (2022)

\bibitem{saharia2022image}
Saharia, C., Ho, J., Chan, W., Salimans, T., Fleet, D.J., Norouzi, M.: Image super-resolution via iterative refinement. IEEE Transactions on Pattern Analysis and Machine Intelligence  \textbf{45}(4),  4713--4726 (2022)

\bibitem{shelhamer2016fully}
Shelhamer, E., Long, J., Darrell, T.: Fully convolutional networks for semantic segmentation. IEEE Transactions on Pattern Analysis and Machine Intelligence  \textbf{39}(4),  640--651 (2016)

\bibitem{9917526}
Sheng, Z., Liu, X., Cao, S.Y., Shen, H.L., Zhang, H.: Frequency-domain deep guided image denoising. IEEE Transactions on Multimedia  \textbf{25},  6767--6781 (2023). \doi{10.1109/TMM.2022.3214375}

\bibitem{Shor:2008:TSM}
Shor, Y., Lischinski, D.: The shadow meets the mask: Pyramid-based shadow removal. Computer Graphics Forum  \textbf{27}(2),  577--586 (Apr 2008)

\bibitem{stamminger2002perspective}
Stamminger, M., Drettakis, G.: Perspective shadow maps. In: Proceedings of the 29th annual conference on Computer graphics and interactive techniques. pp. 557--562 (2002)

\bibitem{takano2007rotation}
Takano, H., Kobayashi, H., Nakamura, K.: Rotation invariant iris recognition method adaptive to ambient lighting variation. IEICE transactions on information and systems  \textbf{90}(6),  955--962 (2007)

\bibitem{tel2023alignment}
Tel, S., Wu, Z., Zhang, Y., Heyrman, B., Demonceaux, C., Timofte, R., Ginhac, D.: Alignment-free hdr deghosting with semantics consistent transformer. arXiv preprint arXiv:2305.18135  (2023)

\bibitem{10.1007/978-3-319-16817-3_8}
Timofte, R., De Smet, V., Van Gool, L.: A+: Adjusted anchored neighborhood regression for fast super-resolution. In: Cremers, D., Reid, I., Saito, H., Yang, M.H. (eds.) Computer Vision -- ACCV 2014. pp. 111--126. Springer International Publishing, Cham (2015)

\bibitem{vasluianu2021shadow}
Vasluianu, F.A., Romero, A., Van~Gool, L., Timofte, R.: Shadow removal with paired and unpaired learning. In: Proceedings of the IEEE/CVF Conference on Computer Vision and Pattern Recognition. pp. 826--835 (2021)

\bibitem{vasluianu2023ntire}
Vasluianu, F.A., Seizinger, T., Timofte, R.: Ntire 2023 image shadow removal challenge report. In: New Trends in Image Restoration (NTIRE 2023) Workshop. (2023)

\bibitem{Vasluianu_2023_WSRD}
Vasluianu, F.A., Seizinger, T., Timofte, R.: Wsrd: A novel benchmark for high resolution image shadow removal. In: Proceedings of the IEEE/CVF Conference on Computer Vision and Pattern Recognition (CVPR) Workshops. pp. 1826--1835 (June 2023)

\bibitem{vaswani2017attention}
Vaswani, A., Shazeer, N., Parmar, N., Uszkoreit, J., Jones, L., Gomez, A.N., Kaiser, {\L}., Polosukhin, I.: Attention is all you need. Advances in neural information processing systems  \textbf{30} (2017)

\bibitem{7893803}
{Vicente}, T.F.Y., {Hoai}, M., {Samaras}, D.: Leave-one-out kernel optimization for shadow detection and removal. IEEE Transactions on Pattern Analysis and Machine Intelligence  \textbf{40}(3),  682--695 (March 2018). \doi{10.1109/TPAMI.2017.2691703}

\bibitem{10.1007/978-3-319-46466-4_49}
Vicente, T.F.Y., Hou, L., Yu, C.P., Hoai, M., Samaras, D.: Large-scale training of shadow detectors with noisily-annotated shadow examples. In: Leibe, B., Matas, J., Sebe, N., Welling, M. (eds.) Computer Vision -- ECCV 2016. pp. 816--832. Springer International Publishing, Cham (2016)

\bibitem{ISTDwang2018STCGAN}
Wang, J., Li, X., Yang, J.: Stacked conditional generative adversarial networks for jointly learning shadow detection and shadow removal. In: Proceedings of the IEEE Conference on Computer Vision and Pattern Recognition. pp. 1788--1797 (2018)

\bibitem{wang2023frequency}
Wang, J., Wu, S., Xu, K., Yuan, Z.: Frequency compensated diffusion model for real-scene dehazing. arXiv preprint arXiv:2308.10510  (2023)

\bibitem{wang2022uformer}
Wang, Z., Cun, X., Bao, J., Zhou, W., Liu, J., Li, H.: Uformer: A general u-shaped transformer for image restoration. In: Proceedings of the IEEE/CVF conference on computer vision and pattern recognition. pp. 17683--17693 (2022)

\bibitem{1284395}
Wang, Z., Bovik, A., Sheikh, H., Simoncelli, E.: Image quality assessment: from error visibility to structural similarity. IEEE Transactions on Image Processing  \textbf{13}(4),  600--612 (2004). \doi{10.1109/TIP.2003.819861}

\bibitem{10.1145/1243980.1243982}
Wu, T.P., Tang, C.K., Brown, M.S., Shum, H.Y.: Natural shadow matting. ACM Trans. Graph.  \textbf{26}(2),  8–es (Jun 2007). \doi{10.1145/1243980.1243982}, \url{https://doi.org/10.1145/1243980.1243982}

\bibitem{5658185}
Xie, X., Zheng, W.S., Lai, J., Yuen, P.C., Suen, C.Y.: Normalization of face illumination based on large-and small-scale features. IEEE Transactions on Image Processing  \textbf{20}(7),  1807--1821 (2011). \doi{10.1109/TIP.2010.2097270}

\bibitem{xie2006efficient}
Xie, X., Lam, K.M.: An efficient illumination normalization method for face recognition. Pattern Recognition Letters  \textbf{27}(6),  609--617 (2006)

\bibitem{xu2022fmnet}
Xu, G., Hou, Q., Zhang, L., Cheng, M.M.: Fmnet: Frequency-aware modulation network for sdr-to-hdr translation. In: Proceedings of the 30th ACM International Conference on Multimedia. pp. 6425--6435 (2022)

\bibitem{6241432}
Yang, Q., Tan, K.H., Ahuja, N.: Shadow removal using bilateral filtering. IEEE Transactions on Image Processing  \textbf{21}(10),  4361--4368 (2012). \doi{10.1109/TIP.2012.2208976}

\bibitem{zamir2022restormer}
Zamir, S.W., Arora, A., Khan, S., Hayat, M., Khan, F.S., Yang, M.H.: Restormer: Efficient transformer for high-resolution image restoration. In: Proceedings of the IEEE/CVF conference on computer vision and pattern recognition. pp. 5728--5739 (2022)

\bibitem{zamir2020cycleisp}
Zamir, S.W., Arora, A., Khan, S., Hayat, M., Khan, F.S., Yang, M.H., Shao, L.: Cycleisp: Real image restoration via improved data synthesis. In: Proceedings of the IEEE/CVF conference on computer vision and pattern recognition. pp. 2696--2705 (2020)

\bibitem{10208804}
Zhang, D., Ouyang, J., Liu, G., Wang, X., Kong, X., Jin, Z.: Ff-former: Swin fourier transformer for nighttime flare removal. In: 2023 IEEE/CVF Conference on Computer Vision and Pattern Recognition Workshops (CVPRW). pp. 2824--2832 (2023). \doi{10.1109/CVPRW59228.2023.00283}

\bibitem{zhang2021designing}
Zhang, K., Liang, J., Van~Gool, L., Timofte, R.: Designing a practical degradation model for deep blind image super-resolution. In: Proceedings of the IEEE/CVF International Conference on Computer Vision. pp. 4791--4800 (2021)

\bibitem{7839189}
Zhang, K., Zuo, W., Chen, Y., Meng, D., Zhang, L.: Beyond a gaussian denoiser: Residual learning of deep cnn for image denoising. IEEE Transactions on Image Processing  \textbf{26}(7),  3142--3155 (2017). \doi{10.1109/TIP.2017.2662206}

\bibitem{zhang2018perceptual}
Zhang, R., Isola, P., Efros, A.A., Shechtman, E., Wang, O.: The unreasonable effectiveness of deep features as a perceptual metric. In: CVPR (2018)

\bibitem{zhang2022spa}
Zhang, X.F., Gu, C.C., Zhu, S.Y.: Spa-former: Transformer image shadow detection and removal via spatial attention. arXiv e-prints pp. arXiv--2206 (2022)

\bibitem{zhou2023breaking}
Zhou, H., Dong, W., Liu, Y., Chen, J.: Breaking through the haze: An advanced non-homogeneous dehazing method based on fast fourier convolution and convnext. In: Proceedings of the IEEE/CVF Conference on Computer Vision and Pattern Recognition. pp. 1894--1903 (2023)

\bibitem{zhu2022efficient}
Zhu, Y., Xiao, Z., Fang, Y., Fu, X., Xiong, Z., Zha, Z.J.: Efficient model-driven network for shadow removal  (2022)

\end{thebibliography}
\end{document}